\documentclass[preprint,12pt]{elsarticle}

\usepackage[letterpaper,top=2cm,bottom=2cm,left=2cm,right=2cm,marginparwidth=1.75cm]{geometry}
\usepackage{amssymb}
\usepackage{amsmath}
\usepackage{comment}
\usepackage{soul}

\usepackage{graphicx}
\usepackage{microtype}
\usepackage{subcaption}
\usepackage{todonotes}
\usepackage{float}
\usepackage{color,hyperref}
\usepackage{multirow} 
\usepackage{rotating}
\usepackage{pdflscape}
\usepackage{hyperref}

\journal{Computer Methods in Applied Mechanics and Engineering}

\begin{document}

\begin{frontmatter}



\title{\textbf{BubbleOKAN: A Physics-Informed Interpretable Neural Operator for High-Frequency Bubble Dynamics}}

\author{ Yunhao Zhang$^{1,\dagger}$, Sidharth S. Menon$^{2,\dagger}$,  Lin Cheng$^{3}$, Aswin Gnanaskandan$^{1}$, Ameya D. Jagtap$^{2, *}$}
\cortext[mycorrespondingauthor]{Corresponding Author: 
  ajagtap@wpi.edu (Ameya D. Jagtap). \\ $\dagger$ First Co-authors (Equal Contributions) \\ \\ Published in Computer Methods in Applied Mechanics and Engineering
}

\address{$^1$ Department of Mechanical and Materials Engineering, Worcester Polytechnic Institute, 100 Institute Road, Worcester, 01609, MA, USA.}
\address{$^2$ Department of Aerospace Engineering, Worcester Polytechnic Institute, 100 Institute Road, Worcester, 01609, MA, USA.}
\address{$^3$ Department of Mechanical Engineering, University of Maryland, 841 Campus Dr, College Park, 20742, MD, USA.}

\begin{abstract}
{In this work, we employ physics-informed neural operators to map pressure profiles from an input function space to the corresponding bubble radius responses. Our approach employs a two-step DeepONet architecture. To address the intrinsic spectral bias of deep learning models, our model incorporates the Rowdy adaptive activation function, enhancing the representation of high-frequency features. Moreover, we introduce the Kolmogorov-Arnold network (KAN) based \textit{two-step DeepOKAN} model, which enhances interpretability (often lacking in conventional multilayer perceptron architectures) while efficiently capturing high-frequency bubble dynamics without explicit utilization of activation functions in any form. We particularly investigate the use of spline basis functions in combination with radial basis functions (RBF) within our architecture, as they demonstrate superior performance in constructing a universal basis for approximating high-frequency bubble dynamics compared to alternative formulations. Furthermore, we emphasize on the performance bottleneck of RBF while learning the high frequency bubble dynamics and showcase the advantage of using spline basis function for the trunk network in overcoming this inherent spectral bias. The model is systematically evaluated across three representative scenarios: (1) bubble dynamics governed by the Rayleigh–Plesset equation with a single initial radius, (2) bubble dynamics governed by the Keller–Miksis equation with a single initial radius, and (3) Keller–Miksis dynamics with multiple initial radii. We also compare our results with state-of-the-art neural operators, including Fourier Neural Operators, Wavelet Neural Operators, OFormer, and Convolutional Neural Operators. Our findings demonstrate that the two-step DeepOKAN accurately captures both low- and high-frequency behaviors, and offers a promising alternative to conventional numerical solvers. The two-step DeepOKAN code is available at \href{https://github.com/ParamIntelligence/Two-Step-DeepOKAN}{https://github.com/ParamIntelligence/Two-Step-DeepOKAN}}
\end{abstract}


\end{frontmatter}


\section{Introduction}
\label{sec1}
Microscopic gas bubbles are integral to a wide range of scientific and engineering applications, including medical techniques like ultrasound-mediated drug delivery \cite{pua2009ultrasound}, thermal ablation \cite{gnanaskandan2019modeling}, and industrial operations such as cavitation-based jet cleaning \cite{cako2022cavitation}. On the other hand, cavitating bubbles also have harmful effects such as erosion and noise generation \cite{gnanaskandan2016large}, thus making an in-depth understanding of bubbles imperative to both harnessing their benefits and mitigating their harmful effects. Understanding bubble dynamics, especially at high driving frequencies, is fundamental to a range of modern technologies. At high frequencies, bubbles experience rapid oscillations and exhibit shape instabilities that can generate intense local pressures, shear stresses, and microstreaming flows. These phenomena govern energy localization, transport, and mixing at microscales. Applications of such high-frequency bubble oscillations include ultrasound contrast imaging (where microbubbles enhance acoustic scattering), targeted drug and gene delivery (where oscillations transiently permeabilize cell membranes), and high-intensity focused ultrasound therapy (where controlled cavitation can ablate tissue). Beyond medicine, high-frequency bubble dynamics are leveraged in sonochemistry to accelerate chemical reactions, and in microfluidics for pumping, mixing, and cleaning at microscales. The dynamic behavior of these oscillating bubbles is traditionally described by models such as Rayleigh-Plesset (R-P) equation \cite{rayleigh1917viii,franc2007rayleigh,brennen2014cavitation}, Keller-Miksis (K-M) equation \cite{keller1980bubble}, Keller-Herring equation\cite{prosperetti1986bubble}, and Gilmore equation \cite{gilmore1952growth}. These models offer essential theoretical understanding of single-bubble dynamics. However, their extension to systems involving multiple interacting bubbles introduces significant computational challenges. The resulting complexity and high computational cost often hinder their feasibility in real-time analysis and large-scale simulations.
\\
The rapid advancement of machine learning has opened new avenues for developing surrogate models that can augment or even replace traditional numerical solvers for complex physical systems; a research direction now widely recognized as scientific machine learning (SciML). Among emerging approaches, neural operator-based frameworks have garnered significant attention for their ability to learn mappings between infinite-dimensional function spaces. A leading example is the Deep Operator Network (DeepONet), introduced by Lu \textit{et al.}~\cite{Lu2021}, which is theoretically grounded in the universal approximation theorem for operators. The DeepONet architecture comprises of two subnetworks: a branch network that encodes the input function and a trunk network that represents the output domain. Their outputs are combined to predict the target function, enabling the model to learn nonlinear operator mappings effectively. Since its introduction, DeepONet has been widely adopted and extended across a range of scientific disciplines. For example, Mao \textit{et al.}~\cite{Mao2021} developed DeepM\&Mnet to model chemical dissociation reactions, while Jin \textit{et al.}~\cite{Jin2022} proposed MIOnet for multi-input operator regression, applying it to coupled diffusion–reaction systems. Peyvan \textit{et al.}~\cite{Peyvan2024} designed a two-step training methodology to enhance DeepONet’s performance on the Riemann problem, and Goswami \textit{et al.}~\cite{goswami2024learning} introduced a Partition-of-Unity (PoU) DeepONet framework for learning operators in chemical kinetics and turbulent flows. Moreover, Osorio \textit{et al.}~\cite{osorio2022forecasting}  explored the use of DeepONet for forecasting dynamics in solar-thermal energy systems. Collectively, these developments highlight DeepONet’s potential as a general-purpose tool for solving physics-informed problems through operator learning over functional spaces.
\\
Another prominent SciFM methodology for solving physics-governed problems is the Physics-Informed Neural Networks (PINNs). The central idea behind PINNs is the incorporation of physical laws, typically expressed as partial differential equations, directly into the loss function of neural networks, thereby guiding the training process through the underlying governing equations \cite{Raissi2019}. Since their introduction by Raissi \textit{et al.}~\cite{Raissi2019}, PINNs have been widely adopted in diverse scientific and engineering domains. By embedding physical constraints, PINNs are capable of generating highly accurate solutions while substantially reducing dependence on large labeled datasets, distinguishing them from purely data-driven models \cite{Raissi2019,Karniadakis2021}. To address limitations related to scalability and computational efficiency, several enhanced variants of the original PINNs framework have been developed. Notable examples include Conservative PINNs \cite{jagtap2020conservative} and eXtended PINNs (XPINNs) \cite{jagtap2020extended}, which leverage conservation principles and domain decomposition strategies to improve performance; see also \cite{penwarden2023unified,hu2023augmented} for unified temporal decomposition and adaptive domain decomposition, respectively. These innovations have enabled the application of PINNs to increasingly complex and high-dimensional problems. For example, Shukla \textit{et al.}~~\cite{shukla2021physics} employed PINNs for the identification of material properties, while Jagtap \textit{et al.}~~\cite{jagtap2022deep} addressed inverse nonlinear water wave problems. Abbasi \textit{et al.}~\cite{abbasi2025history} developed PINN-based workflows for porous media; see also their survey on PINNs applied to shock waves \cite{abbasi2025challenges}. Zhu \textit{et al.}~~\cite{Zhu2021} constructed a surrogate PINN model to simulate melt pool dynamics in metal additive manufacturing, and Mao \textit{et al.}~~\cite{Mao2020} applied PINNs to the analysis of high-speed compressible fluid flows; see also Jagtap \textit{et al.}~~\cite{jagtap2022physics} for solving inverse problems in high-speed flows. Cheng \textit{et al.}~~\cite{Cheng2022} integrated PINNs into a representative volume element (RVE) framework to capture structure–property relationships in heterogeneous materials. Furthermore, Zhu \textit{et al.}~~\cite{Zhu2019} employed PINNs for surrogate modeling aimed at uncertainty quantification. PINNs have also been employed for the solution of high-dimensional partial differential equations~\cite{hu2024tackling,menon2025anant}. The foundational mathematical theory underlying PINNs and their extended variants (XPINNs) has been developed by several authors~\cite{de2024error,hu2021extended,shin2020convergence}. A comprehensive survey of the approximation theory for physics-informed machine learning is provided by Mhaskar et al.~\cite{mhaskar2025approximation}.
\\
The application of machine learning to bubble dynamics remains relatively underexplored. Lin \textit{et al.}~\cite{Lin2021_1,Lin2021_2} employed DeepONet to investigate multi-scale bubble behavior. However, their study addresses only a limited subset of the broader problem and exhibits several notable limitations. First, the pressure profiles in their simulations are generated using Gaussian random fields, which do not accurately reflect the sinusoidal nature of pressure variations encountered in various practical  applications. Second, their dataset is constrained to low amplitude pressure fluctuations on the order of $10^3$~Pa, whereas realistic scenarios often involve amplitudes up to $10^6$~Pa. Moreover, a critical limitation lies in their exclusive focus on low-frequency excitations. Bubble dynamics can vary significantly under different frequency regimes, and the absence of high-frequency data impairs the model’s generalizability. These limitations emphasize the necessity for a more comprehensive machine learning framework for modeling bubble dynamics. In this work, we overcome these limitations. The main contributions of this work are summarized as follows:
\begin{itemize}
    \item We employ a physics-informed deep neural operator designed for simulating complex bubble dynamics. Specifically, we use a two-step DeepONet architecture with Rowdy activation functions \cite{Jagtap2022}. A systematic \textit{ablation study} is conducted to quantify the influence of Rowdy activations on predictive performance.
    
    \item Furthermore, we propose an interpretable variant based on Kolmogorov–Arnold Networks, named \textit{two-step DeepOKAN}. The proposed architectures establish a robust and generalizable computational framework that integrates physical laws with deep operator learning to enhance predictive accuracy in modeling multiphase flow phenomena.
    
    \item We address the challenging regime of bubble dynamics characterized by high frequencies and large amplitudes, thereby extending the applicability of physics-informed operator networks to highly nonlinear fluid systems. In particular, we solve two different types of bubble dynamics equations, namely, the Rayleigh--Plesset and Keller--Miksis equations, over a large range of amplitudes (\( [1,10] \times 10^5 \) Pa) and frequencies (\( [200,2000] \times 10^6 \) KHz), representing a wide range of physically realistic scenarios.
    
    \item We provide a comprehensive benchmark for assessing the performance of physics-informed operator networks in reproducing bubble behavior under diverse flow and boundary conditions, and we compare the proposed method against state-of-the-art neural operators, including Fourier Neural Operators (FNO), Wavelet Neural Operators (WNO), OFormer, and Convolutional Neural Operators (CNO).
\end{itemize}

The remainder of this paper is organized as follows. Section~2 presents the governing R-P and K-M equations for bubble dynamics. Section~3 highlights the limitations of the original DeepONet architecture in learning R-P equation-based bubble dynamics. In Section~4, we discuss various operator learning architectures, including physics-informed DeepONet, two-step DeepONet, and the proposed two-step DeepOKAN. Section~5 presents the results, ranging from single-step initial radius cases based on the R-P and K-M equations to multiple initial radius scenarios using the K-M equation with two-step DeepONet and two-step DeepOKAN. We also compare our results with other state-of-the-art methods, such as FNO, CNO, WNO, and OFormer. Furthermore, a systematic ablation study is performed with respect to the number of Rowdy activation terms. Finally, Section~\ref{sec6} concludes the study, summarizing the key findings, limitations, and future directions.

\section{Governing Equations}
\label{sec3}
The physics of bubble dynamics is well characterized by mathematical models. We utilize two different equations: the Rayleigh–Plesset equation, a fundamental ordinary differential equation (ODE) that describes the idealized radial motion of a spherical bubble in an incompressible liquid~\cite{Hong2024,franc2007rayleigh,brennen2014cavitation}, and the Keller–Miksis equation, a more comprehensive model that accounts for liquid compressibility and acoustic effects to more accurately capture bubble behavior~\cite{keller1980bubble}.

\subsection{Rayleigh-Plesset (R-P) equation} 
Assuming a spherical bubble, the governing equation can be written as:
\begin{equation}
    \label{eq:R-P}
    R \ddot{R} + \frac{3}{2} \dot{R}^2 = \frac{1}{\rho} \left( P_G - P_{\infty} - 4\mu \frac{\dot{R}}{R}  - \frac{2S}{R} \right),
\end{equation}
where \( R \) denotes the bubble radius as a function of time \( t \), evolving under the influence of the far-field pressure \( P_{\infty} \). Thus, \(\dot{R}\) and \(\ddot{R}\) represent the first and second order derivative of \(R\) accordingly. Here, \( S \) represents the surface tension coefficient, \( \mu \) is the dynamic viscosity of the liquid, and \( \rho \) is the liquid density. \( P_G \) is the internal gas pressure, modeled as:
\begin{equation}
    P_G = P_{G0} \left(\frac{R_0}{R} \right)^{3k}.
\end{equation}
Here, \( P_{G0} \) corresponds to the initial internal gas pressure \( P_G(t=0) \). The exponent \( k = 1.4 \) is the polytropic coefficient, reflecting the adiabatic behavior of the gas. To ensure consistency between the training process and the governing physical laws, we employ the non-dimensional form of the R–P equation for normalization. This avoids potential numerical errors during mathematical operations. The derivation of the non-dimensional form and the numerical solution methodology using Runge-Kutta method can be found in~\ref{Nondim R-P}. 

\subsection{Keller-Miksis (K-M) equation}
 With the same assumptions used for the R-P equation, except for the incompressibility of the liquid medium, the K-M equation can be expressed as 
\begin{equation}
    \label{eq:K-M}
    \left(1 - \frac{\dot{R}}{c} \right) R \ddot{R} 
    + \frac{3}{2} \left(1 - \frac{\dot{R}}{3c} \right) \dot{R}^2 
    = \left(1 + \frac{\dot{R}}{c} \right) \frac{P_L - P_\infty}{\rho} 
    + R \frac{\dot{P}_L - \dot{P}_\infty}{\rho c},
\end{equation}
where
\[
P_L = P_G - 4\mu \frac{\dot{R}}{R} - \frac{2S}{R},
\]
\[
\dot{P}_L = \dot{P}_G + 4\mu \left( \frac{\dot{R}^2}{R^2} - \frac{\ddot{R}}{R} \right),
\]
Here, \( c \) denotes the speed of sound in the liquid, while all other variables are the same as in the R-P equation. The derivation of the non-dimensional form and the numerical solution methodology using Runge Kutta method can be found in~\ref{Nondim K-M}.

\section{Limitations of the Vanilla DeepONet Approach to Bubble Dynamics}
\label{sec2}
Lin \textit{et al.}~\cite{Lin2021_1,Lin2021_2} employed the Vanilla DeepONet architecture to investigate bubble dynamics governed by the Rayleigh–Plesset equation. In their framework, the branch network takes the pressure variation \(\Delta P\) as input, where the imposed pressure is defined as \(P_\infty = P_0 + \Delta P\), consistent with the assumption of small amplitude fluctuations. Here, \(P_0\) represents the initial pressure. The trunk network receives time as its input. Both the branch and trunk networks are implemented as standard feedforward neural networks (FFNNs), and their outputs are combined through a dot product to generate the final prediction.

\begin{table}[htbp]
\caption{Design of Experiments for generating training and validation data}
\label{tab:DoE}
\centering
\begin{tabular}{l|c|c}
\hline
Parameters & Range & $\#$ Samples \\
\hline
Initial radius, \( R_0 \) (\(\mu\)m) & \( 50 \) & 1 \\
Pressure amplitude, \( \text{amp} \) (Pa) & \( [1,10] \times 10^5 \) & 10 \\
Pressure frequency, \( f \) (KHz) & \( [200,2000] \) & 300 \\
\hline
\end{tabular}
\end{table}

\begin{figure}
\centering
\includegraphics[trim=1cm 1cm 1cm 2cm, clip =true,scale=0.35]{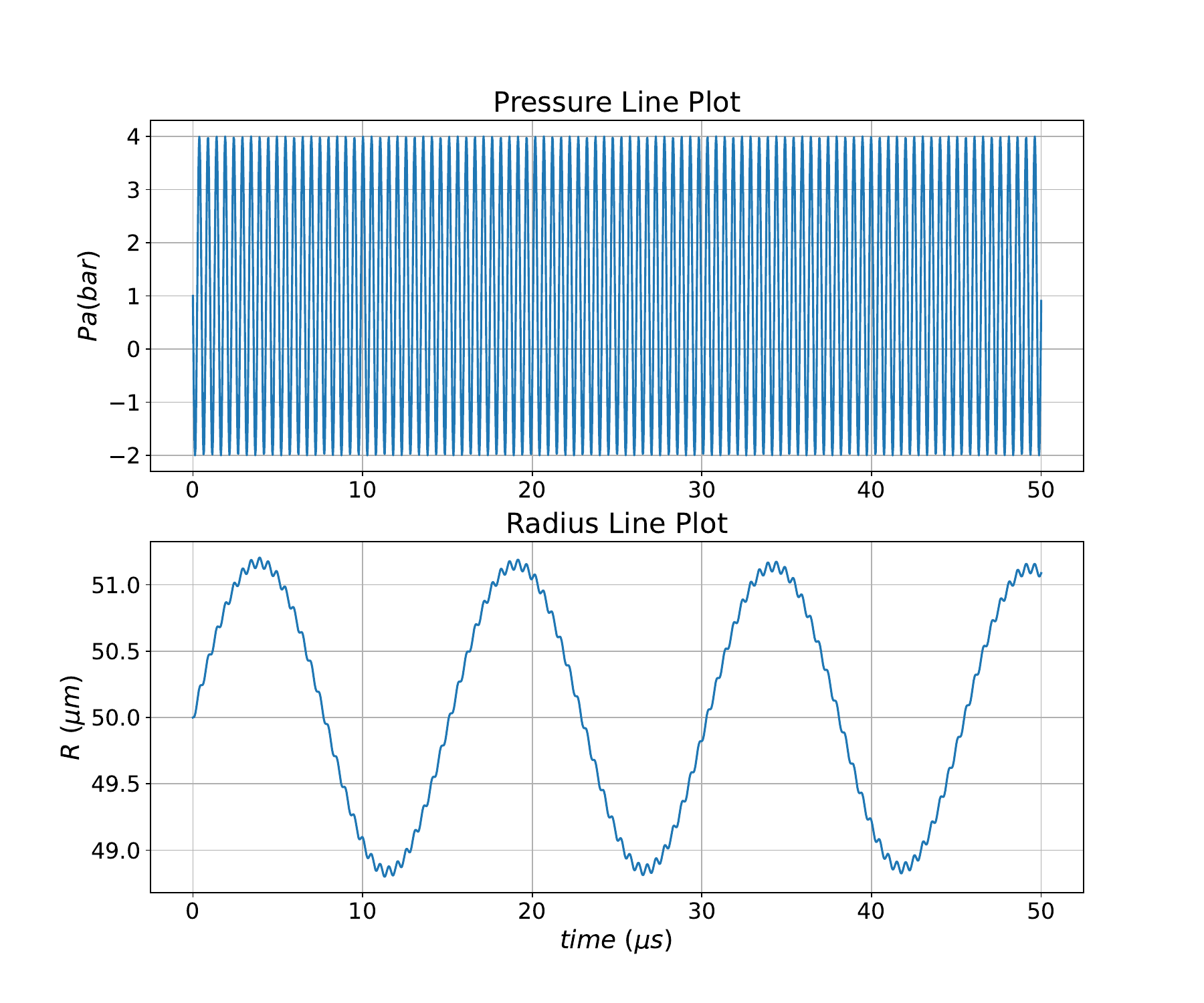}
\caption{A bubble dynamics simulation based on R-P equation obtained from APECSS for a bubble with \(R_0 = 50\mu m, f = 1970 \text{KHz}, ~\text{amp} = 3\times10^5Pa, t = 50\mu s\). Top: pressure profile. Bottom: Radius evolution}
\label{fig:database}
\end{figure}

To evaluate the generalization capability of the model, we initially employed this vanilla DeepONet architecture to simulate bubble dynamics. The training data were generated using the open-source code APECSS~\cite{Denner2023}, which numerically solves the Rayleigh–Plesset equation. A total of 3,000 datasets driven by sinusoidal pressure profiles were generated following the design of experiments (DoE) outlined in Table~\ref{tab:DoE}, with 2,400 datasets used for training and 600 for validation. An example from the dataset is shown in Figure~\ref{fig:database}, corresponding to a bubble with an initial radius of \(R_0 = 50\,\mu\text{m}\), subjected to an acoustic pressure field with a frequency of \(f = 1970\,\text{KHz}\) and an amplitude of \(3 \times 10^5\,\text{Pa}\) over a time span of \(50\,\mu\text{s}\). 

\begin{figure}[htbp]
    \centering
    \begin{subfigure}[b]{0.45\textwidth}
        \centering
        \includegraphics[trim=0 0 0 0, clip,width=\textwidth]{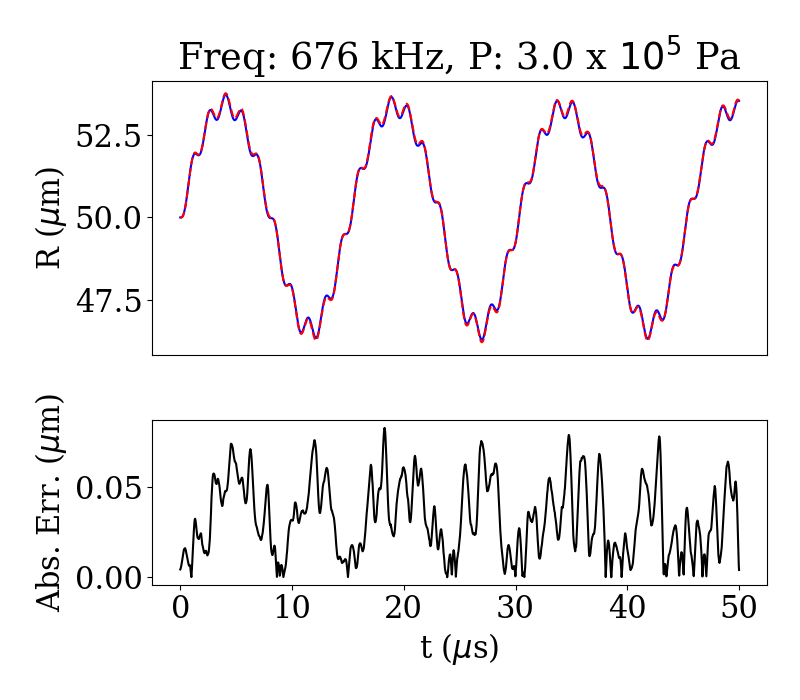}
        \caption{}
        \label{fig:old_low}
    \end{subfigure}
    \hspace{0.02\textwidth}
    \begin{subfigure}[b]{0.45\textwidth}
        \centering
        \includegraphics[trim=0 0 0 0, clip,width=\textwidth]{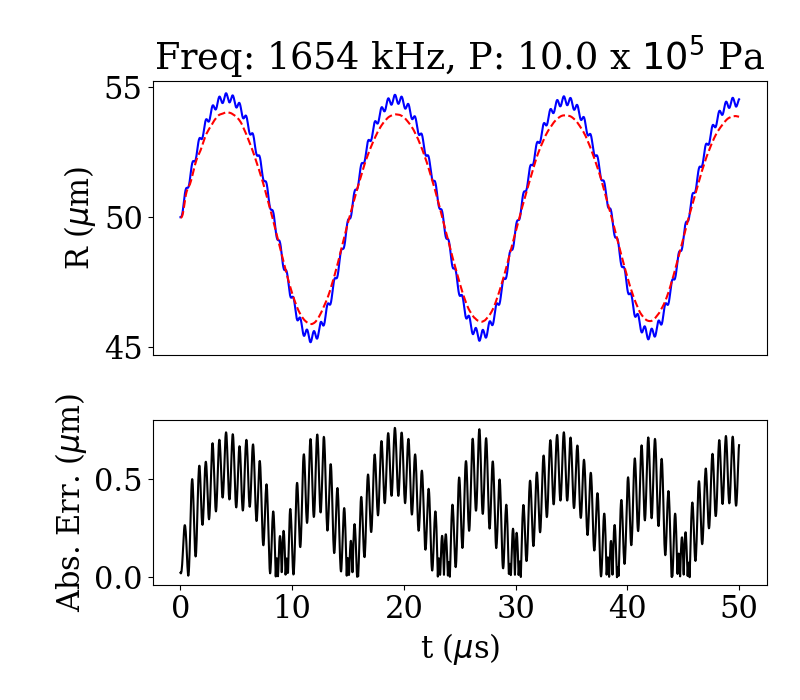}
        \caption{}
        \label{fig:old_high}
    \end{subfigure}
    \caption{Validation results from R-P equation using the vanilla DeepONet after 500,000 epochs. (a) Bubble dynamics driven by pressure with frequency of 676\,KHz and amplitude of \(3 \times 10^5\,\text{Pa}\), representing low frequency case. (b) Bubble dynamics driven by pressure with frequency of 1654\,KHz and amplitude of \(10 \times 10^5\,\text{Pa}\), representing high frequency case.}
    \label{fig:old_results}
\end{figure}

Figure~\ref{fig:old_results} presents the predictions generated by the vanilla DeepONet architecture after training. Two representative cases of pressure-driven bubble dynamics are shown, both with an initial radius of \(R_0 = 50\,\mu\text{m}\). Case 1 corresponds to a pressure signal with a frequency of 630.43\,KHz and an amplitude of \(9 \times 10^5\,\text{Pa}\), while Case 2 involves a relatively higher frequency of 1654\,KHz and a lower amplitude of \(10 \times 10^5\,\text{Pa}\). To optimize model performance, we explored multiple configurations of the branch and trunk networks, varying the depth from 3 to 8 layers and the width from 128 to 512 neurons. The results displayed in Figure~\ref{fig:old_results} correspond to the best-performing configuration, 8 layers with 512 neurons per layer, trained for 500,000 epochs, which yielded the lowest training and validation losses. Figure~\ref{fig:old_low} illustrates the online prediction for Case 1, where the model output closely aligns with the ground truth, with a low absolute error confirming the model’s ability to capture low-frequency bubble dynamics with high accuracy. In contrast, Figure~\ref{fig:old_high} shows the prediction for Case 2, where the model captures the overall trend but fails to resolve the finer details of the high-frequency oscillations. This behavior is further exacerbated in cases involving ultra-high frequencies (e.g., 1900\,KHz and 2000\,KHz). These observations highlight the spectral bias (a propensity to learn low-frequency components more readily than high-frequency ones) inherent to the original DeepONet architecture. More details on the preliminary study on the original model can be found in \cite{zhang2025bubble}.

\section{Operator Learning Architectures for High-Frequency Bubble Dynamics}
\label{sec4}
 Among the various neural operator architectures, DeepONets have shown remarkable flexibility and efficiency across a range of physical systems. In this section, we first discuss the physics-informed DeepONet architecture \cite{wang2021learning}, which integrates governing physical laws directly into the training process to enhance data efficiency and enforce physical consistency.  We then discuss the two-step DeepONet architecture \cite{lee2024training}, which is more accurate and efficient than vanilla DeepONet. To this end, we propose a more interpretable version based on Kolmogorov-Arnold Networks (KAN), named as, two-step DeepOKAN architecture.

\subsection{Single-Step DeepONet}
\begin{figure}[H]
\centering
\includegraphics[trim=0cm 0cm 0cm 0cm, clip, width=\linewidth]{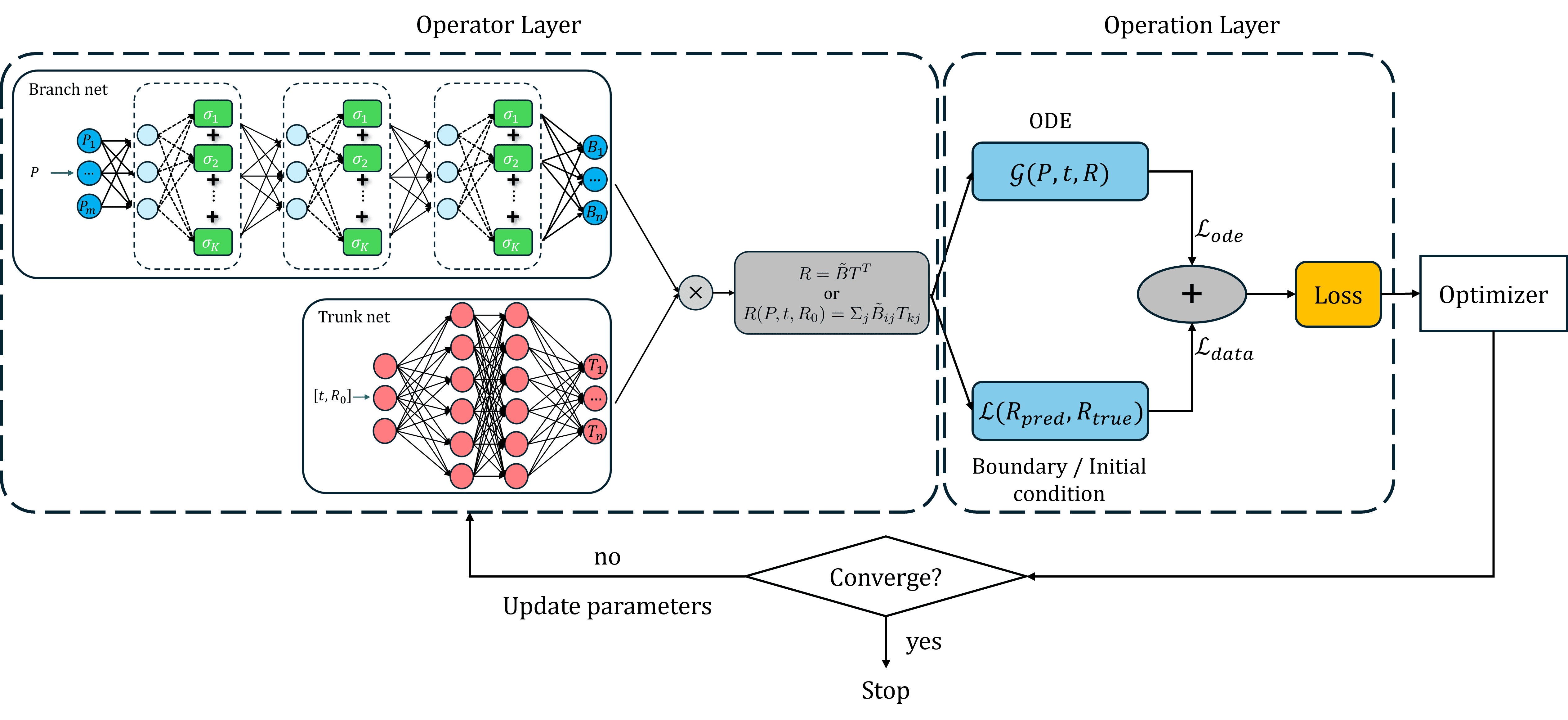}
\caption{Architecture of DeepONet: The operator layer employs a Kronecker-based branch net (input: full pressure profile) and a trunk net (input: time and initial radius). The operation layer enforces physical constraints on outputs, while training minimizes a combined data-ODE loss via iterative optimization.}
\label{fig:architecture}
\end{figure}

Figures \ref{fig:architecture} depict the schematic of the physics-informed DeepONet architecture, wherein physical constraints are seamlessly embedded alongside the vanilla DeepONet framework to enforce governing laws. 

Deep learning models are known to exhibit spectral bias  arising from both network parameterization and training dynamics \cite{Cao2021, Rahaman2019}. Adaptive activation functions have been proposed to mitigate this bias by introducing bounded, non-monotonic variations that reduce saturation and enhance sensitivity to high-frequency features \cite{Jagtap2022}. Related concepts include the noise activation function \cite{Gulcehre2016} and probabilistic activation function \cite{Shridhar2019}, both of which improve model expressivity. Such adaptive activations also promote smoother convergence and improved accuracy across diverse architectures, see, \cite{Jagtap2020, jagtap2020locally}, for more details; a comprehensive overview is provided by Jagtap and Karniadakis \cite{jagtap2023important}. In particular, Jagtap et al. \cite{Jagtap2022} proposed the Rowdy activation function, which augments conventional activation functions with sinusoidal terms to capture high-frequency behaviors in problems such as the Helmholtz equation and the Riemann problem featuring discontinuities \cite{Peyvan2024}. Building on these advances, the present work integrates the Rowdy activation within the branch network of the DeepONet architecture. This design introduces learnable frequency and phase parameters, thereby alleviating spectral bias and enhancing the model’s capacity to represent complex bubble dynamics.

As discussed previously, the large magnitude of the pressure amplitude renders \( \Delta P \) unsuitable as an input for the branch network, as it fails to adequately represent the global input function space. Instead, our model employs the full pressure profile as the branch network input, producing a latent variable \( B \) that encapsulates the extracted features. The trunk network encodes the temporal domain and initial radius into a latent representation \( T \) in a higher-dimensional space, for which a standard FFNN suffices. The predicted bubble radius \( R(P, t, R_0) \) is obtained via a linear operation on the latent variables:
\begin{equation}
    R(P, t, R_0) = BT^\top,
\end{equation}
where \( B \in \mathbb{R}^{m \times d} \) and \( T \in \mathbb{R}^{n \times d} \), with \( n \) denoting the temporal length, \( m \) the batch size, and \( d \) the latent dimension. To ensure physical consistency, a \textit{SmoothReLU} (\textit{softplus}) activation is applied to \( R \), enforcing positive radius predictions.

With physical constraints incorporated, the total loss comprises of two components: a \textit{data loss} (\(\mathcal{L}_{\text{data}}\)) from network predictions and an \textit{ODE loss} (\(\mathcal{L}_{\text{ode}}\)) enforcing the governing dynamics. Let \(\mathcal{L}\) denote the mean squared error (MSE). The total loss is expressed as
\begin{equation}
    \label{eq:loss}
    \begin{aligned}
        \mathcal{L} &= w_{\text{data}} \mathcal{L}_{\text{data}} + w_{\text{ode}} \mathcal{L}_{\text{ode}} \\
                    &= w_{\text{data}} \frac{1}{N_d} \sum_{i=1}^{N_d} \|R_i - \bar{R}_i\|_2^2
                    + w_{\text{ode}} \frac{1}{N_c} \sum_{i=1}^{N_c} \|\mathcal{G}(R_i)\|_2^2,
    \end{aligned}
\end{equation}
where \(\mathcal{L}_{\text{data}}\) measures the discrepancy between the predicted and reference radius profiles, and \(\mathcal{L}_{\text{ode}}\) represents the residual of the governing dynamics. Here, \(\mathcal{G}(\cdot)\) is the Runge--Kutta-based numerical operator,
\begin{equation}
    \mathcal{G}(R_N) := y_{N+1} - \left(y_N + dt \sum_{j=1}^{s} b_j k_j\right),
\end{equation}
with \(dt\) as the time step, \(s\) the number of stages, and \(b_j\) the corresponding weights. The state variable is $y = [R, ~\dot{R}]^T.$
Intermediate stages \(k_j\) and coefficients \(b_j\) follow the RK5(4)7M scheme \cite{Dormand1986}, featuring seven stages and fifth-order accuracy. The weights \(w_{\text{data}}\) and \(w_{\text{ode}}\) balance data fidelity and physics-informed regularization.

\subsection{Two-Step DeepONet with Rowdy Activations}
\label{sec:2sdeeponet}
The central concept of DeepONet lies in constructing functional \textit{basis} representations via the trunk network, while the branch network learns its corresponding \textit{coefficients}. 
\begin{figure}[htbp]
\centering
\includegraphics[trim=0cm 0cm 0cm 0cm, clip, width=0.8\linewidth]{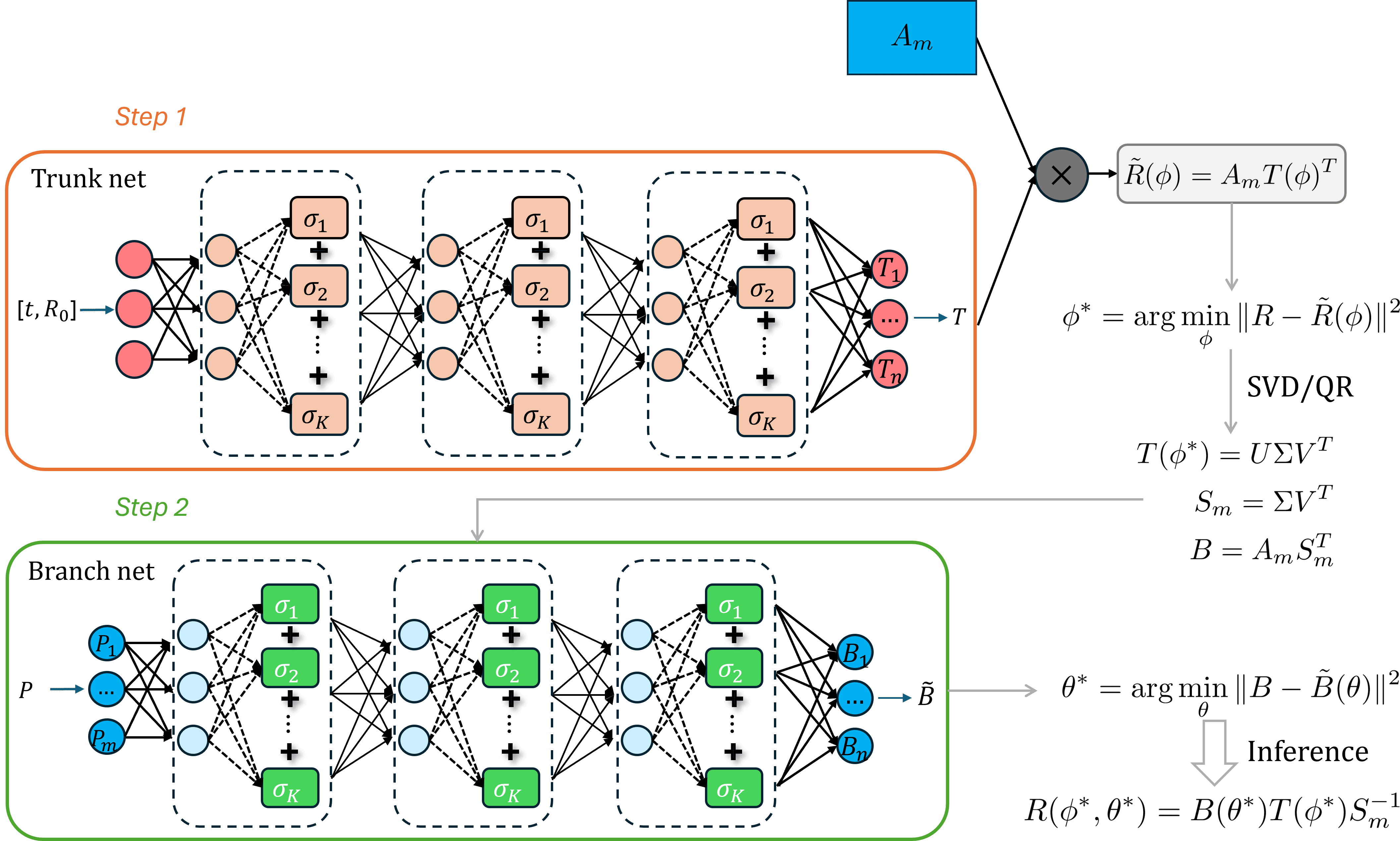}
\caption{Two step training architecture for DeepONet.}
\label{fig:twostepDON}
\end{figure}
To improve training efficiency and generalization, Lee \textit{et al.}~\cite{Lee2024} introduced a two-step training strategy in which QR factorization is applied to the pretrained basis functions of the trunk network before training the branch network. Building on this framework, Peyvan \textit{et al.}~\cite{Peyvan2024} further enhanced the approach by employing singular value decomposition (SVD) instead of QR decomposition and by using Rowdy activation functions. Adopting this idea, we incorporate the two-step training methodology into our model; readers are referred to~\cite{Lee2024,Peyvan2024} for additional implementation details.

The overall training workflow and network architecture are illustrated in Figure~\ref{fig:twostepDON}. In the first stage, a learnable variable \( A_m \) is introduced to emulate the latent representation of the branch network in conventional single-step training. The trunk network is first optimized to obtain its parameters \( \phi^* \), after which the branch network is trained using the SVD-transformed outputs of the pretrained trunk network to determine its optimal parameters \( \theta^* \).

\subsection{Two-Step DeepOKAN}
Kolmogorov–Arnold Neural Networks (KANs), recently proposed by Liu et al. \cite{liu2024kan} is build upon the classical Kolmogorov–Arnold representation theorem, which originally proposed for shallow networks with limited hidden units. However, the recently formulation of KAN extends this foundational concept to contemporary deep learning architectures by leveraging training through backpropagation similar to multi-layer perceptron (MLP). This enables the framework to scale naturally to networks with any arbitrary number of layers, $L$. In this formulation, a KAN with $L$ layers is represented as a composition of $L$ learnable basis functions:
$$
f(\mathbf{x}) = (\Phi_{L-1} \circ \Phi_{L-2} \circ \dots \circ \Phi_1 \circ \Phi_0)(\mathbf{x}) ,   
$$
where $\mathbf{x}\in\mathbb{R}^d$ and $f(\mathbf{x})$ is a multivariate continuous function. Also, the functionality of basis function $\Phi$ in each layer may be defined as follows:
$$
\mathbf{x}_{l+1} = \Phi_l \mathbf{x}_{l},
$$
where $\mathbf{x}_l \in \mathbb{R}^{n_l}$ and $\mathbf{x}_{l+1} \in \mathbb{R}^{n_{l+1}}$ represent the input and output of the $l$-th layer, respectively, in a KAN architecture with an arbitrary number of $L$ layers, and $n_l$ denotes number of neurons in the $l $-th layer. The transformation $\Phi_l $ associated with the $l $-th layer can then be expressed in matrix form as follows:

$$
\Phi_l = \begin{bmatrix}
\phi_{l,1,1}(\cdot) & \phi_{l,1,2}(\cdot) & \dots & \phi_{l,1,n_l}(\cdot)\\
\phi_{l,2,1}(\cdot) & \phi_{l,2,2}(\cdot) & \dots & \phi_{l,2,n_l}(\cdot)\\
\vdots & \vdots & \ddots & \vdots\\
\phi_{l,n_{l+1},1}(\cdot) & \phi_{l,n_{l+1},2}(\cdot) & \dots & \phi_{l,n_{l+1},n_l}(\cdot)\\
\end{bmatrix}.
$$

\noindent

In both the original formulation of the Kolmogorov–Arnold representation theorem \cite{liu2024kan} and its recent extensions (\cite{10763509}, \cite{Sidharth2024ChebyshevPK}, \cite{shukla2024comprehensive}), KAN have been constructed using spline basis functions based on the original formulation of KAN as well as with various alternative formulations of the basis functions with sufficient internal parameterizations. For the spline basis function, two important hyperparameters, namely, polynomial order ($k$) and grid size ($G$), are pivotal in determining the performance and expressivity of KAN architecture. 
The function approximation capability of KAN is fundamentally governed by the choice of basis functions, denoted as $\Phi$. A key advantage of KAN lies in their intrinsic interpretability, an aspect largely absent in conventional MLPs. KANs are adept at capturing both compositional structures and univariate functional relationships, often outperforming MLPs in terms of predictive accuracy. Moreover, they demonstrate superior parametric efficiency, achieving comparable or better results with significantly fewer trainable parameters. Despite these strengths, several practical challenges persist. Notably, the selection of important hyperparameters, such as grid resolution and spline order in the context of a spline basis, is highly problem-specific and non-trivial, highlighting the need for further research to enable more robust and automated deployment of the KAN architecture.

There has been a steep rise in utilization of the KAN architecture by leveraging their aforementioned benefits for solving PDEs in a physics-informed manner for low-dimensional \cite{shukla2024comprehensive} as well as for high-dimensional \cite{menon2025anant} PDEs. In the context of operator learning, Abueidd et al. \cite{abueidda2025deepokan} proposed DeepOKAN that utilizes a basis formulation comprising of radial basis functions (RBF) as opposed to the conventional spline basis, mainly owing to their enhanced computational efficiency and better approximation properties compared to the B-splines. They evaluated their model on several mechanics problems and showed the superior performance of DeepOKAN over the DeepONet architecture with respect to the model accuracy.

\begin{figure}[htbp]
\centering
\includegraphics[trim=0cm 0cm 0cm 0cm, clip, width=0.7\linewidth]{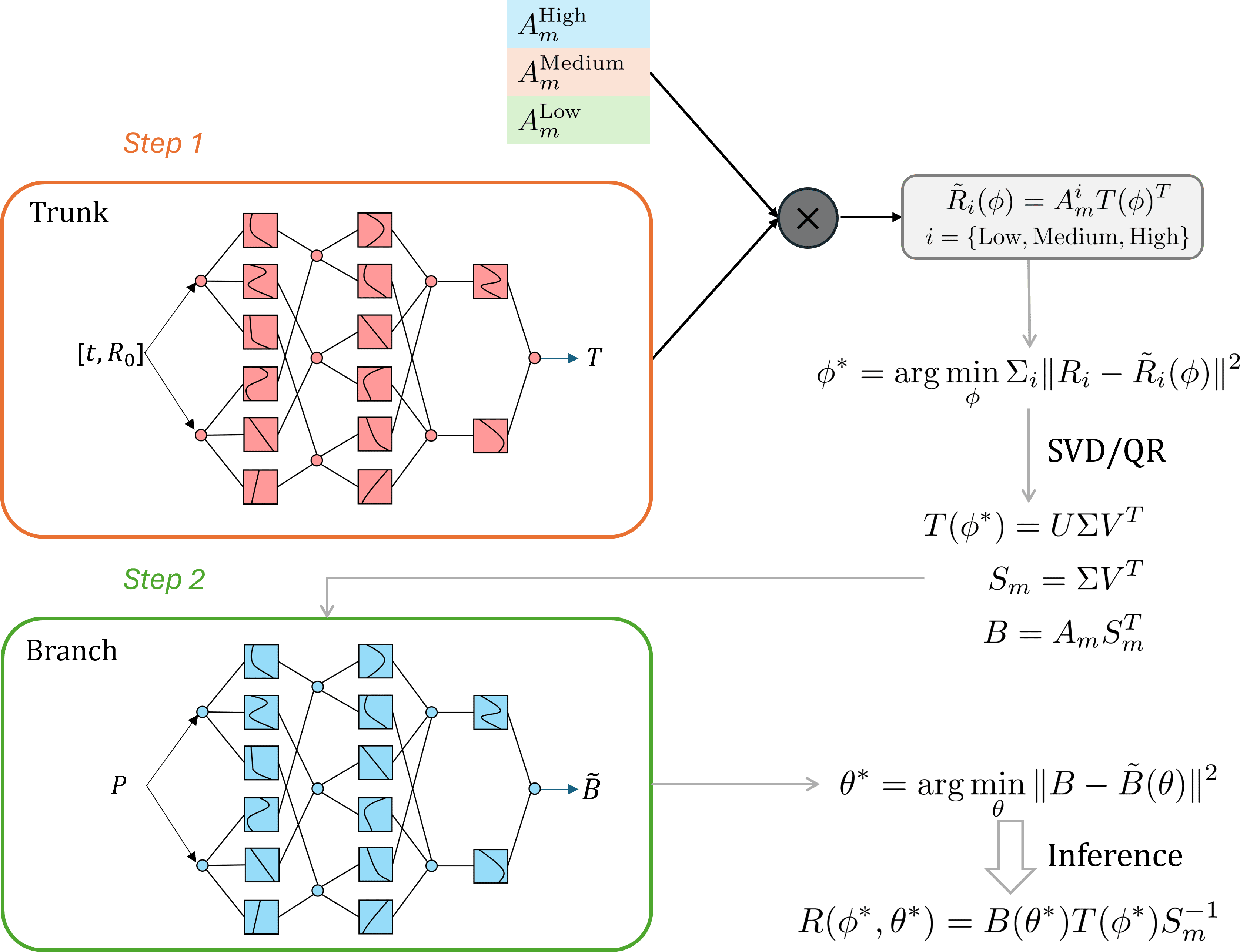}
\caption{Two step training architecture for DeepOKAN.}
\label{fig:twostep_KAN}
\end{figure}
In this work, we propose a two-step DeepOKAN architecture that leverages the favorable properties of KAN such as interpretability and parametric efficiency while training the trunk and branch networks independently which can potentially lead to reduction in the overall training time as showcased by the previous work \cite{Peyvan2024}. To this end, an important contribution of our work is that we provide an alternate strategy for overcoming the spectral bias while learning the operator governing the high-frequency bubble dynamics solely through continual learning using the two-step DeepOKAN architecture as opposed to utilizing the adaptive activations \cite{jagtap2020locally} and Rowdy activations \cite{jagtap2022deep}, which are inevitable while tackling spectral bias using MLP based architectures. While Abueidd et al. \cite{abueidda2025deepokan} proposed vanilla DeepOKAN with RBF basis, we show the need to carefully select the basis for a KAN based architecture depending on the problem that is being addressed as opposed naively selecting a fixed basis for all the problems. Figure~\ref{fig:twostep_KAN} presents the schematic of the two-step DeepOKAN, which follows the two-step DeepONet framework but replaces the MLP components with more interpretable KAN architecture. The discussions henceforth will elaborate on the training procedures that were adopted for the two-step DeepOKAN architecture independently for the trunk network and the branch network.

\subsubsection{Trunk Network: Overcoming spectral bias through continual learning}

The trunk network is responsible for constructing a common basis that can effectively represent the bubble dynamics across a wide range of frequencies present in the training data. Furthermore, the training of the branch network is built upon the performance of the trunk network for the two-step framework, and poor representation of the frequencies present in the training data by the trunk network can propagate forward and adversely influence the training of the branch network, thereby leading to poor generalization of the overall model during inference. Hence, an efficient training procedure for the trunk network is always beneficial and leads to better generalization of the model during inference. To this end, the MLP-based two-step DeepONet (discussed in Section \ref{sec:2sdeeponet}) utilized Rowdy activation and trained using the following formulation of the objective function.

\begin{equation}
    \phi^* = \text{arg min}_{\phi}\|R - A_mT(\phi)^{T} \|^2
    \label{eq:trunkobj}
\end{equation}
Here, $\tilde{R}(\phi) = A_mT(\phi)^{T}$ where $A_m$ is a trainable tensor that acts as a proxy for the branch network in its absence while $T(\phi)$ is the prediction by the trunk network and the reconstructed output prediction $\tilde{R}(\phi)$ is obtained through the tensor product of $A_m$ and $T(\phi)$. 

However, in the context of two-step DeepOKAN, training the trunk network by utilizing the formulation of the objective shown in Equation \ref{eq:trunkobj} is observed to be not very effective for overcoming the spectral bias. Continual learning is a prominent learning paradigm utilized in machine learning for sequentially training a model for new tasks while preserving the old knowledge gathered from previous tasks without forgetting. In the context of KAN, Lu et al.\cite{liu2024kan} showcased the effectiveness of the KAN architecture over the MLP counterpart for continual learning of simple 1D functions. They showed that MLP forgets old knowledge while KAN preserves them without catastrophic forgetting for a toy problem. While this is not entirely true based on the current problem we are solving, we allude to utilizing continual learning with \textit{partial reminders}  for overcoming the spectral bias in the context of two-step DeepOKAN. For bubble dynamics, we sort the dataset based on their dominant frequencies in \text{High}, \text{Medium}, and \text{Low} frequency ranges. After the data segregation, the model is fed with data from specific frequency ranges sequentially at regular intervals during model training. In the current scenario, the 'partial reminders' ensure that the model does not completely forget the old knowledge from the previous frequency ranges while learning a new window of frequency range. It is also worthwhile to note that, the two step DeepOKAN solely utilizes the richness of the basis functions and doesn't utilize any additional activation functions or learnable parameters outside of the KAN architecture for overcoming the spectral bias.

\begin{equation}
    \phi^* = \arg \min_{\phi} \Sigma_i \|R^{i} - A_m^{i}T(\phi)^{T} \|^2 
\end{equation}

Here, $i = \{\text{High, Medium, Low}\}$ and $T(\phi)$ is prediction by the trunk network. Also note that, if $A_m \in \mathbb{R}^{p\times q}$ and $a + b + c = p$, then $\text{Concatenate}\{A_m^{\text{High}} \in \mathbb{R}^{a \times q}, A_m^{\text{Medium}} \in \mathbb{R}^{b \times q}, A_m^{\text{Low}} \in \mathbb{R}^{c \times q}\} = A_m \in \mathbb{R}^{p \times q}$. Also, $\text{Concatenate}\{R^{\text{High}} \in \mathbb{R}^{a \times q \times r}, R^{\text{Medium}} \in \mathbb{R}^{b \times q \times r}, R^{\text{Low}} \in \mathbb{R}^{c \times q \times r}\} = R \in \mathbb{R}^{p \times q \times r}$.
The procedure for continual learning may be summarized as follows.
\\\\
Learning High Frequencies:
\begin{equation}
    \phi^* = \arg \min_{\phi} \|R^{\text{High}} - A_m^{\text{High}}T(\phi)^{T} \|^2 
\end{equation}
Learning Medium Frequencies while preserving the knowledge from High Frequencies:
\begin{equation}
    \phi^* = \arg \min_{\phi} \|R^{\text{Medium}} - A_m^{\text{Medium}}T(\phi)^{T} \|^2 + \lambda_{\text{High}} \|R^{\text{High}} - A_m^{\text{High}}T(\phi)^{T} \|^2
\end{equation}
Learning Low Frequencies while preserving the knowledge from both Medium and High Frequencies:
\begin{equation}
    \phi^* = \arg \min_{\phi}  \|R^{\text{Low}} - A_m^{\text{Low}}T(\phi)^{T} \|^2 +\lambda_{\text{Medium}} \|R^{\text{Medium}} - A_m^{\text{Medium}}T(\phi)^{T} \|^2 + \lambda_{\text{High}} \|R^{\text{High}} - A_m^{\text{High}}T(\phi)^{T} \|^2
\end{equation}
Here, the $\lambda$ parameter controls the 'partial reminders' from the previous step of the continual learning that preserves knowledge gathered from the previous frequency range while learning a new frequency range. Also, it is worthwhile to note that the training for the current step is warm-started using trained parameters from the previous step. Therefore, when $\lambda = 0$ the process is similar to transfer learning. However for the current problem of learning the bubble dynamics, the aim is to learn the entire frequency range present in the training data by partitioning them into manageable frequency ranges while preserving the knowledge across the entire range. Hence, $\lambda > 0$ in the current work.

\subsubsection{Branch Network}
While the previous section discussed the training process of trunk network through continual learning, the current section discusses the details regarding training of the branch network. Here, training is performed using the full batch with the utilization of the RBF \cite{abueidda2025deepokan} basis function.

\begin{equation}
    \theta^* = \arg \min_{\theta} \|B - \tilde{B}(\theta) \|^2 
\end{equation}

\noindent For training the branch network, we utilize the RBF basis owing to its superior computational efficiency over the spline basis function of equivalent representational capacity. However, this was not the case for the trunk network discussed in the previous section since the RBF basis could only learn the global low-frequency structure of the bubble dynamics and failed to learn the dominant high-frequency components irrespective of the number of terms used within the RBF basis or the size of the overall KAN architecture. In a way, this emphasizes the need for careful selection of an appropriate basis function based on the underlying problem that is being solved.

\section{Results and Discussion}
\label{sec5}
In this section, we consider three cases: bubble dynamics governed by the R-P equation with a single initial radius \(R_0 = 50~\mu\text{m}\), representing the simplest scenario; the K-M equation with a single initial radius, introducing liquid compressibility effects relative to the R-P model; and the K-M equation with multiple initial radii, aimed at evaluating the model’s capability in capturing bubble dynamics across higher-dimensional parameter spaces. 
We employed both single-step (Sec. 5.1 and 5.2) and two-step (Sec. 5.3) DeepONet architectures with the Rowdy activation function. Morover, the proposed two-step DeepOKAN framework (Sec. 5.4) is employed to solve the most complicated K-M equation with multiple initial radii. To emphasize its use for bubble dynamics, we refer to it as \textit{BubbleOKAN}.

Model training was performed on an NVIDIA GeForce RTX 3070 GPU with 8~GB of memory, paired with an Intel Core i7-12700 CPU and 25.4~GB of system RAM. The Adam optimizer was employed to advance the training process. The hyperparameters used for each case are summarized in Table~\ref{tab:hyperparameter}.

\begin{table}[H]
\caption{Hyperparameters used in each subsection}
\label{tab:hyperparameter}
\centering
\begin{tabular}{l|c|c|c}
\hline
Subsection & Learning rate & Batch size & Architecture \\
\hline
Section~\ref{session:R-P} & 0.0005 & 200 & Branch net: [2000, 512, 512, 512, 512, 512, 512, 512] \\
                          &        &     & Trunk net: [1, 512, 512, 512, 512, 512, 512, 512] \\
Section~\ref{session:K-M} & 0.0005 & 200 & Branch net: [2000, 512, 512, 512, 512, 512, 512, 512] \\
                          &        &     & Trunk net: [1, 512, 512, 512, 512, 512, 512, 512] \\
Section~\ref{session:multR} & 0.0001 & 25  & Branch net: [2000, 512, 512, 512, 512, 512, 512, 512] \\
                          &        &     & Trunk net: [2, 512, 512, 512, 512, 512, 512, 512] \\
\hline
\end{tabular}
\end{table}

\subsection{Bubble Dynamics Governed by the R-P Equation with Single Initial Radius}
\label{session:R-P}
In this section, we evaluate the model's learning capability and performance on this simplest scenario.
The model is trained on data simulated according to the design of experiments in Table~\ref{tab:DoE}. Of 3000 datasets, 2400 are used for training and 600 for validation. Each simulation runs for \(50~\mu\text{s}\) and is segmented into samples of size 2000. Training proceeds for 500{,}000 epochs. Since this case involves single-radius prediction, the initial condition is not explicitly enforced, reducing the loss function to
\begin{equation}
    \label{eq:singleRloss}
    \mathcal{L} = \mathcal{L}_{\text{data}} + 100\,\mathcal{L}_{\text{ode}}.
\end{equation}
The weights (1 and 100) are chosen based on the initial epoch losses to balance the reduction of both terms.

\begin{figure}[htpb]
\centering
{
\centering
\includegraphics[width=0.3\linewidth]{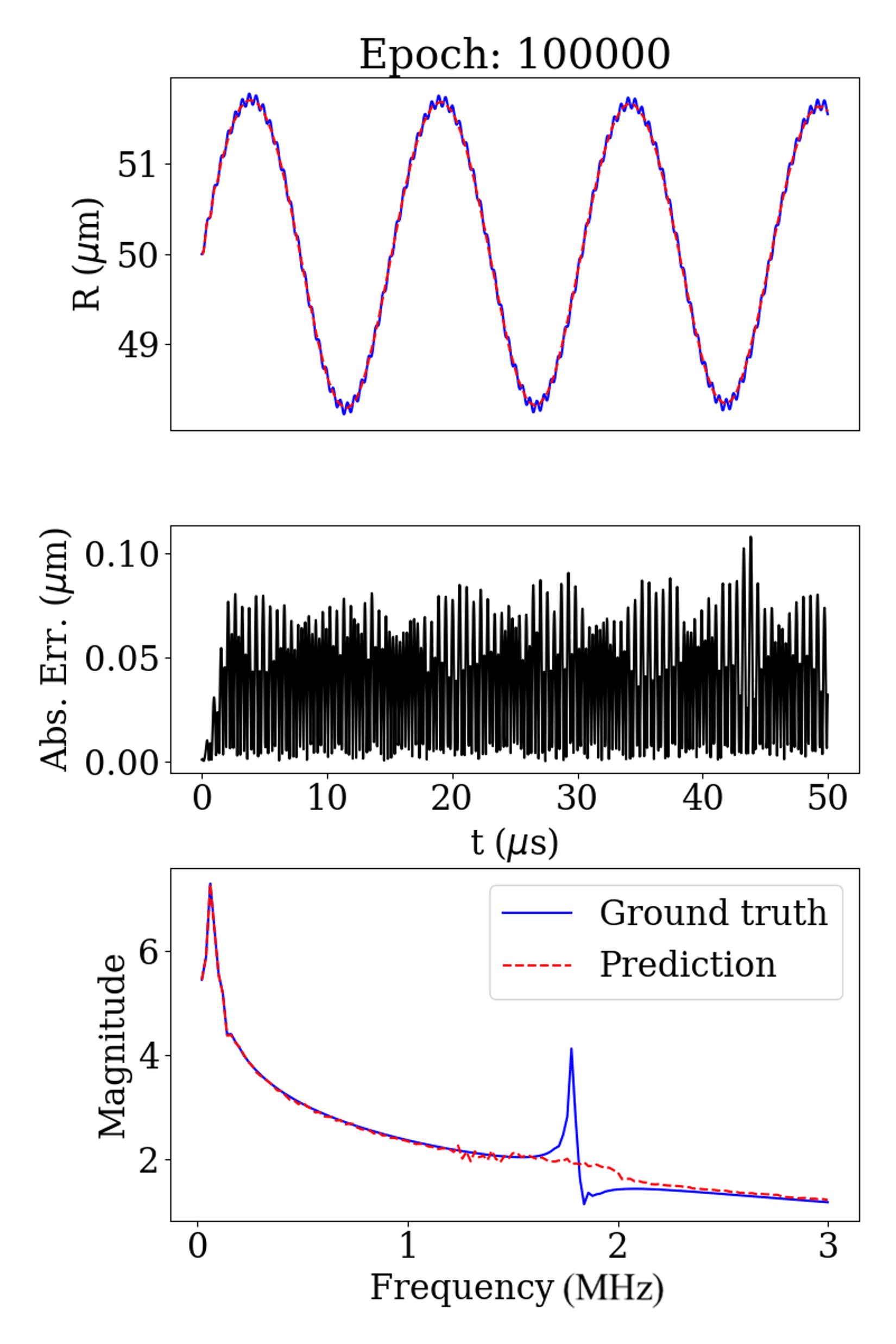}
}
\hspace{0.001cm}
\centering
{
\centering
\includegraphics[width=0.3\linewidth]{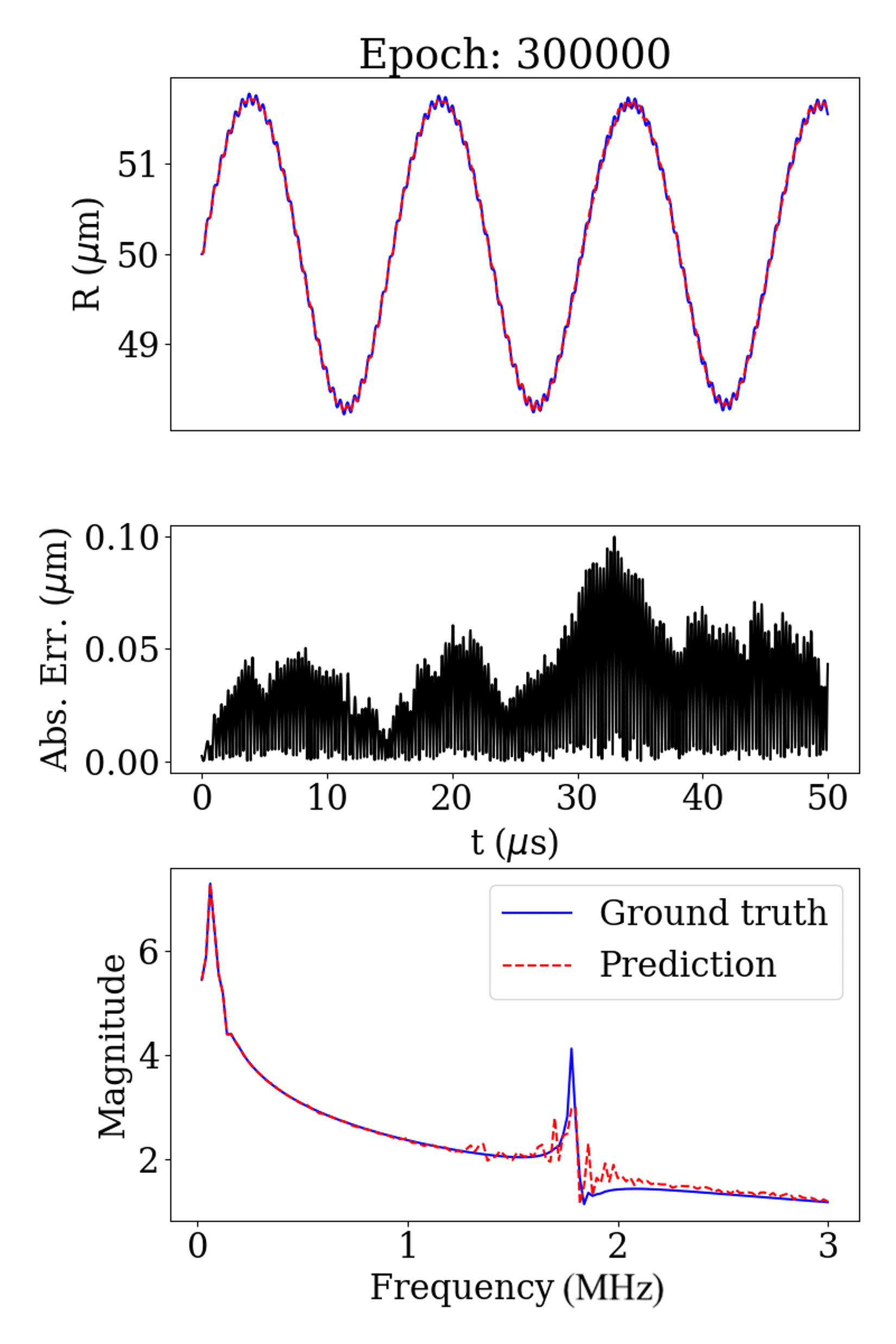}
}
\hspace{0.001cm}
\centering
{
\centering
\includegraphics[width=0.3\linewidth]{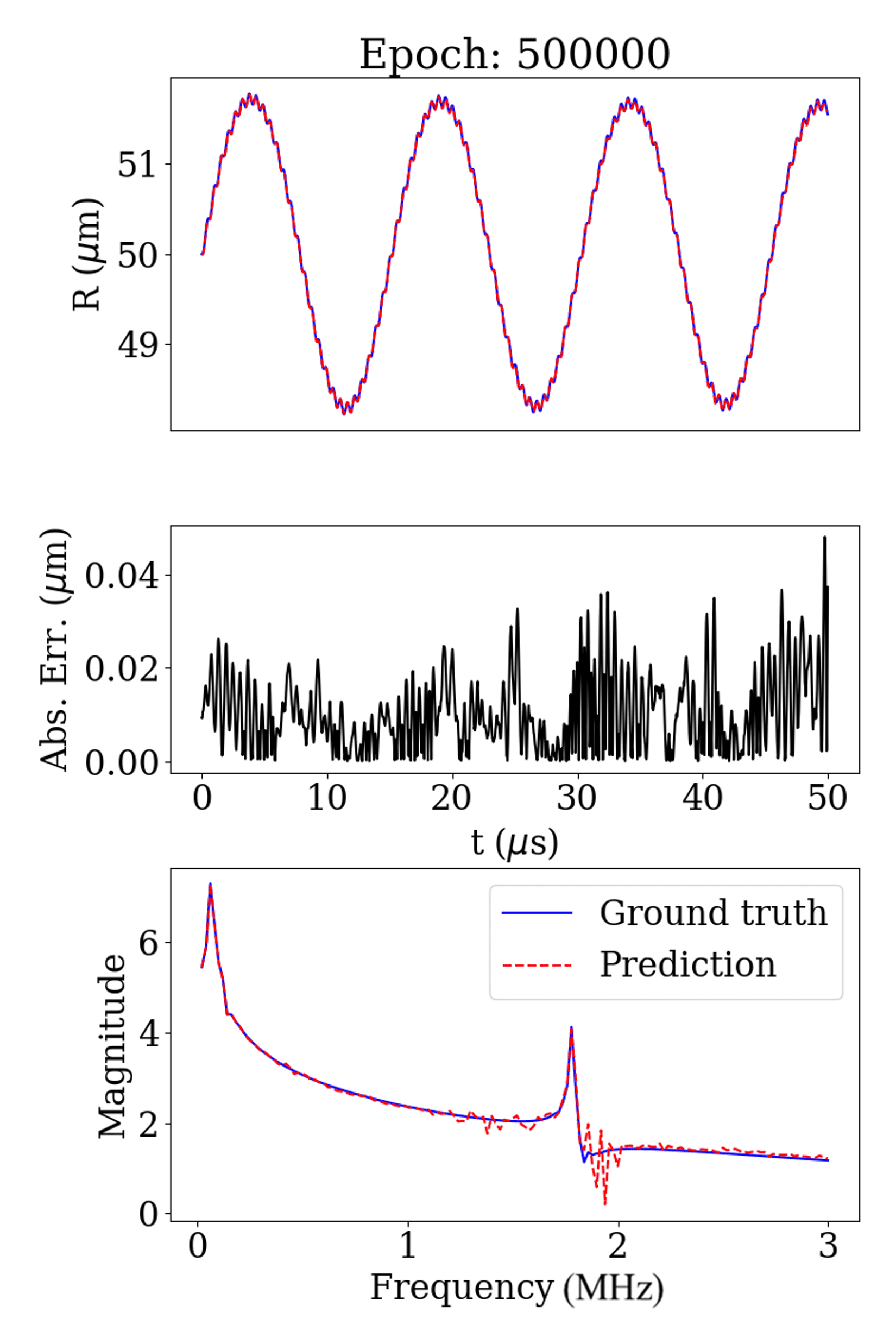}
}
\caption{Validation of bubble dynamics simulated with a driving frequency of \(1800~\text{KHz}\) and an amplitude of \(4 \times 10^5~\text{Pa}\) after 100{,}000, 300{,}000, 500{,}000 epochs and the corresponding data in frequency space.}
\label{fig:validation_RP_highf}
\end{figure}

A high-frequency validation case with a driving frequency of \(1800~\text{kHz}\) and an amplitude of \(4 \times 10^5~\text{Pa}\) is used to assess model performance at different training epochs (Figure~\ref{fig:validation_RP_highf}). The figure illustrates how the model progressively learns key features during training. Each validation includes three plots: (1) comparison between the ground truth and the online prediction, (2) the corresponding absolute error, and (3) the frequency-domain response obtained via Fast Fourier Transform (FFT). In bubble dynamics, two peaks typically appear in the frequency spectrum: the first corresponds to the natural frequency, while the second arises from the driving frequency of the external pressure field \(P_\infty\). 
At epoch 100{,}000, the model accurately captures the natural frequency but underestimates the driving frequency, yielding an absolute error within \([0, 0.1]\,\mu\text{m}\), approximately \(0.2\%\) of the initial radius (\(50\,\mu\text{m}\)). The second FFT peak is shifted relative to the ground truth. By epoch 300{,}000, the model begins to learn the driving frequency, reducing the early-time absolute error and moving the second FFT peak closer to the correct position. At epoch 500{,}000, the second FFT peak aligns well with the driving frequency, and the absolute error decreases further to below \(0.05\,\mu\text{m}\), indicating that the predicted bubble radius closely matches the ground truth.

\subsection{Bubble Dynamics Governed by the K-M Equation with Single Initial Radius}
\label{session:K-M}
Compared with the R–P equation, the K–M equation incorporates liquid compressibility, providing a more accurate description of bubble dynamics, particularly under high-amplitude oscillations. This section focuses on the model’s ability to learn such complex behavior.

A similar dataset is constructed based on the DoE shown in Table~\ref{tab:DoE}, with each dataset representing bubble dynamics simulated over \(55~\mu\text{s}\) and comprising 2000 sampled data points. The loss function follows the same formulation as in Eq.~\ref{eq:singleRloss}. The number of training epochs is increased to 600{,}000 to account for the additional complexity introduced by liquid compressibility in the K–M equation. Both the neural network loss and the governing equation loss exhibit learning trends comparable to those observed in the R–P case.

\begin{figure}[htpb]
\centering
{
\centering
\includegraphics[width=0.3\linewidth]{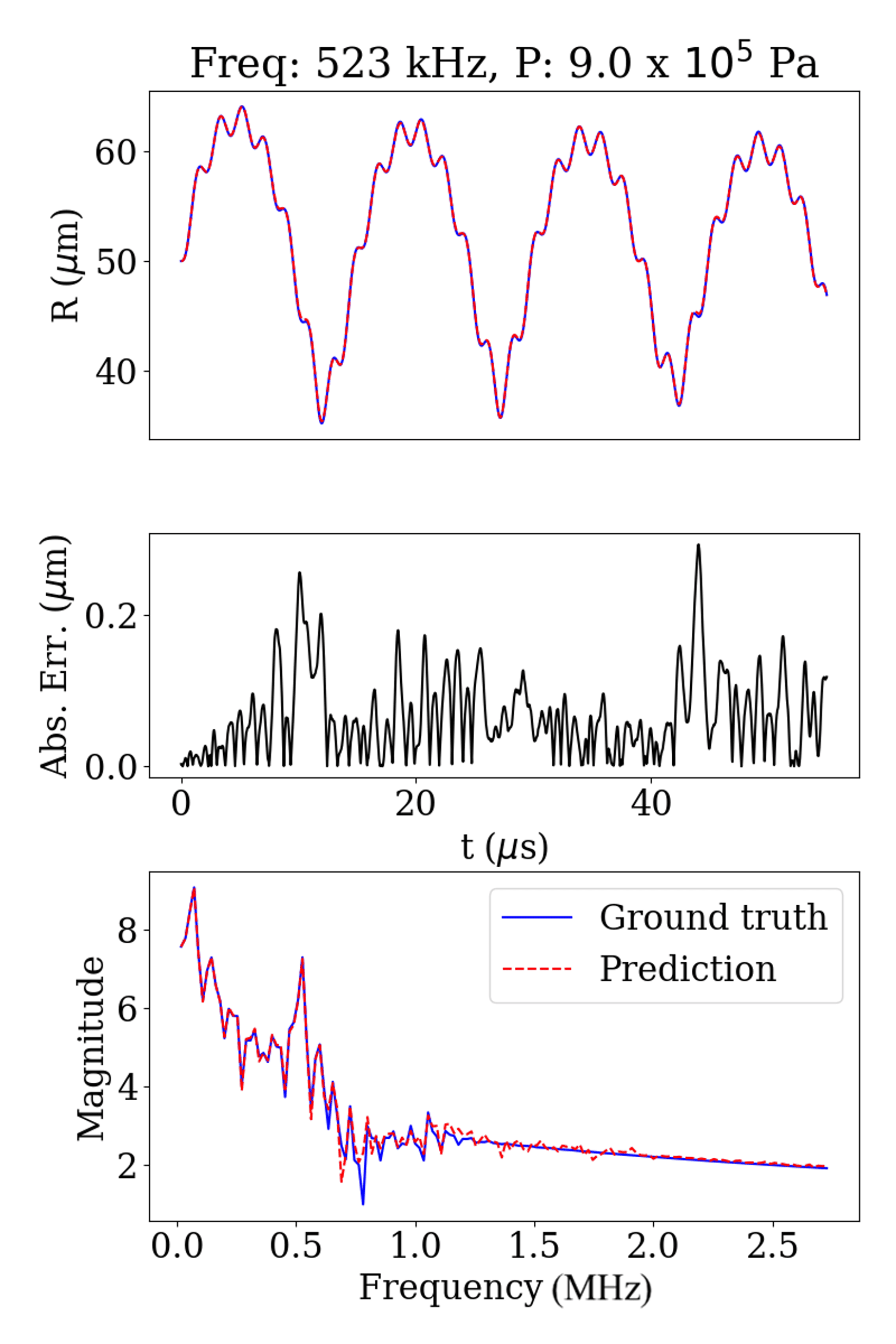}
}
\hspace{0.001cm}
\centering
{
\centering
\includegraphics[width=0.3\linewidth]{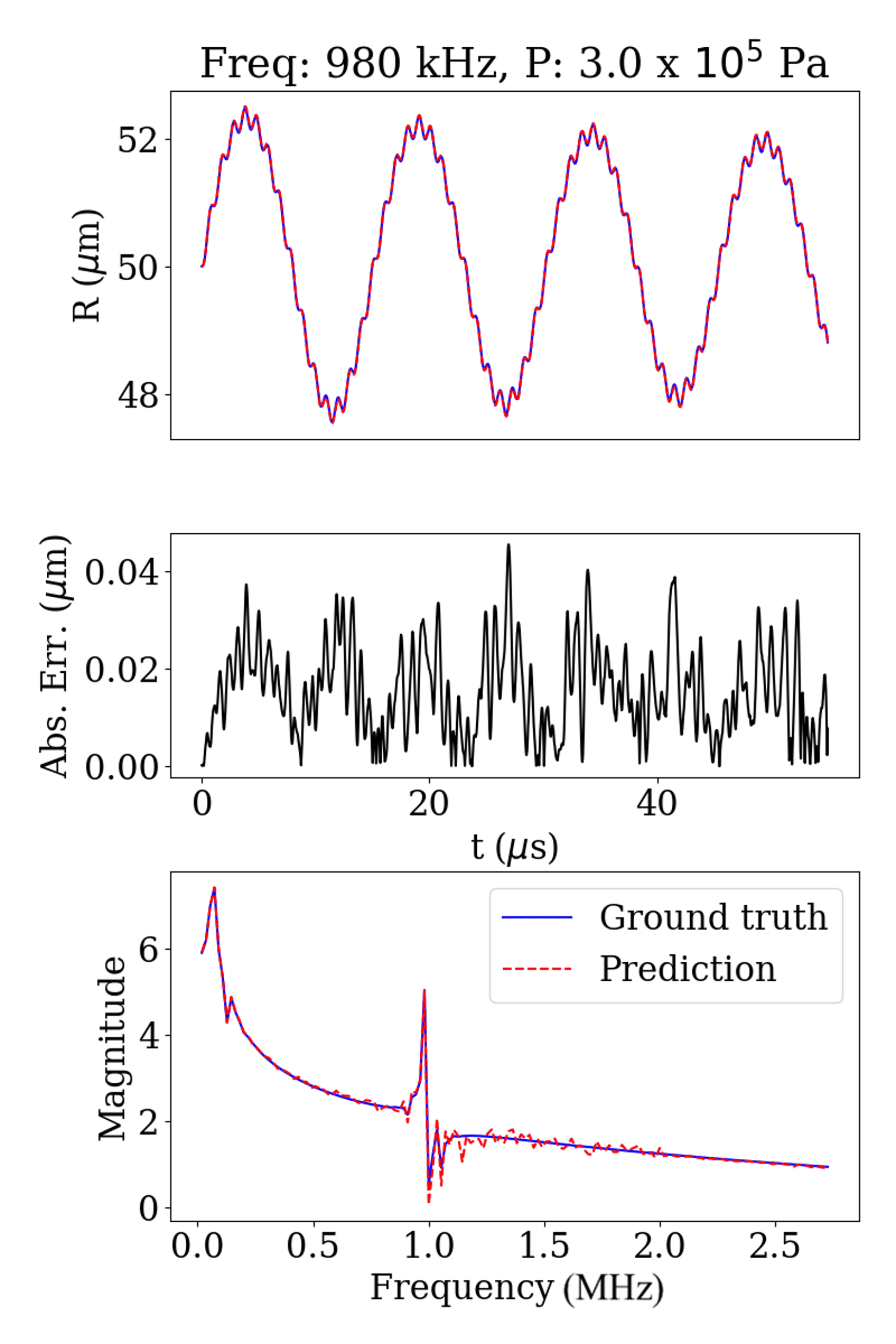}
}
\hspace{0.001cm}
\centering
{
\centering
\includegraphics[width=0.3\linewidth]{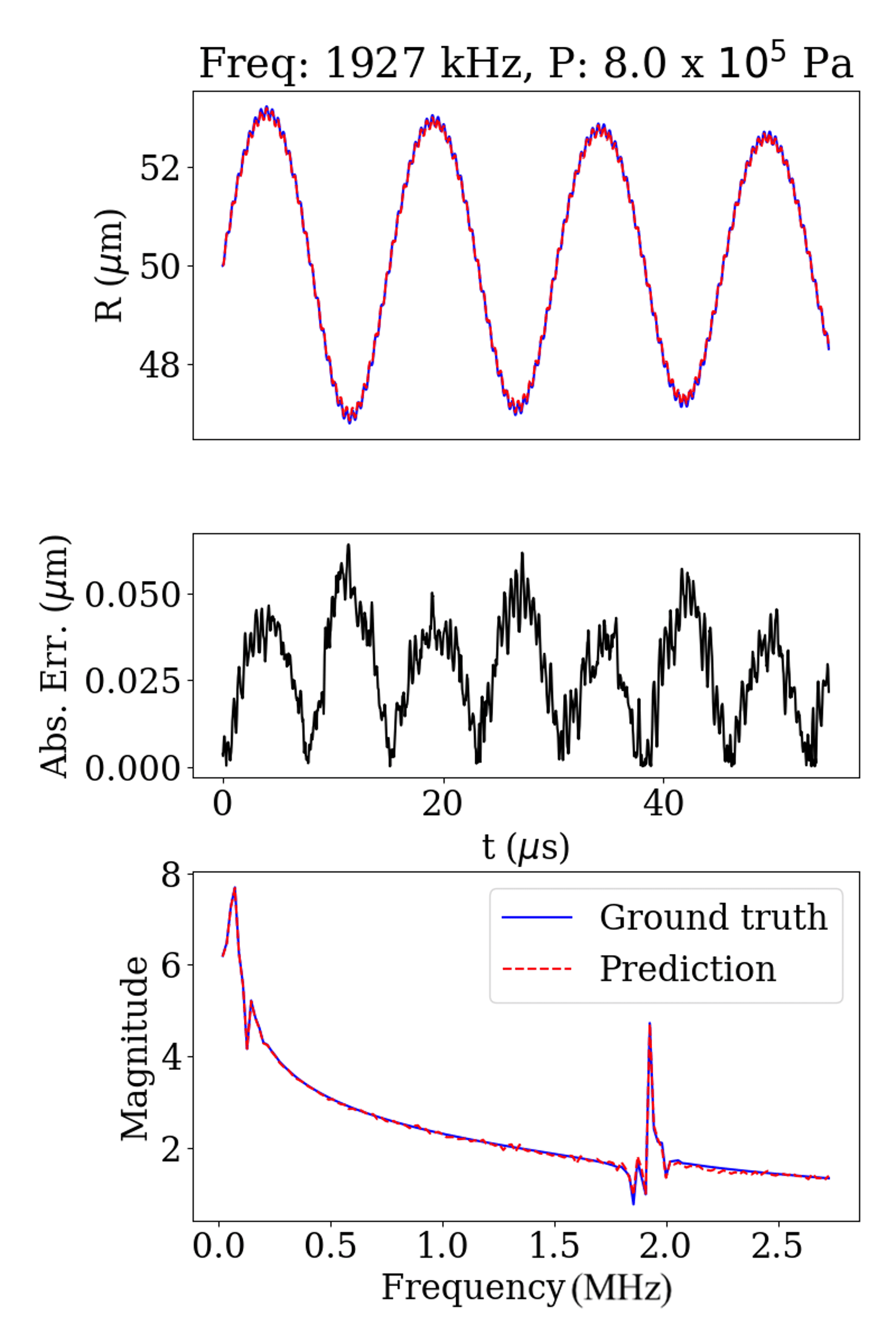}
}
\caption{Validation of bubble dynamics (using K--M) simulated with: (a) a frequency of \(523~\text{KHz}\) and an amplitude of \(9 \times 10^5~\text{Pa}\), (b) a frequency of \(980~\text{KHz}\) and an amplitude of \(3 \times 10^5~\text{Pa}\), and (c) a frequency of \(1927~\text{KHz}\) and an amplitude of \(8 \times 10^5~\text{Pa}\), after 600{,}000 training epochs, along with the corresponding data in frequency space.}
\label{fig:validation_KM}
\end{figure}
Three validation cases representing driving frequencies from different ranges are presented in Figure~\ref{fig:validation_KM}. The first case corresponds to a low frequency of \(523~\text{kHz}\) and an amplitude of \(9 \times 10^5~\text{Pa}\). Noticeable discrepancies are observed in the radius profile, with a maximum absolute error of approximately \(0.2~\mu\text{m}\). The second validation case, simulated at a frequency of \(980~\text{kHz}\) and an amplitude of \(3 \times 10^5~\text{Pa}\), represents a mid-range frequency in the design of experiments. Here, the prediction aligns closely with the ground truth, with a maximum absolute error of \(0.04~\mu\text{m}\). The final case, corresponding to a high driving frequency of \(1927~\text{kHz}\) and an amplitude of \(8 \times 10^5~\text{Pa}\), demonstrates strong performance in this high-frequency regime, with the absolute error remaining below \(0.05~\mu\text{m}\). Across all cases, the model consistently captures the damping effects arising from liquid compressibility and accurately identifies the driving frequency in the frequency domain.

\begin{figure}[]
    \centering
    \includegraphics[trim=0cm 0cm 0cm 0cm, clip, width=0.8\linewidth]{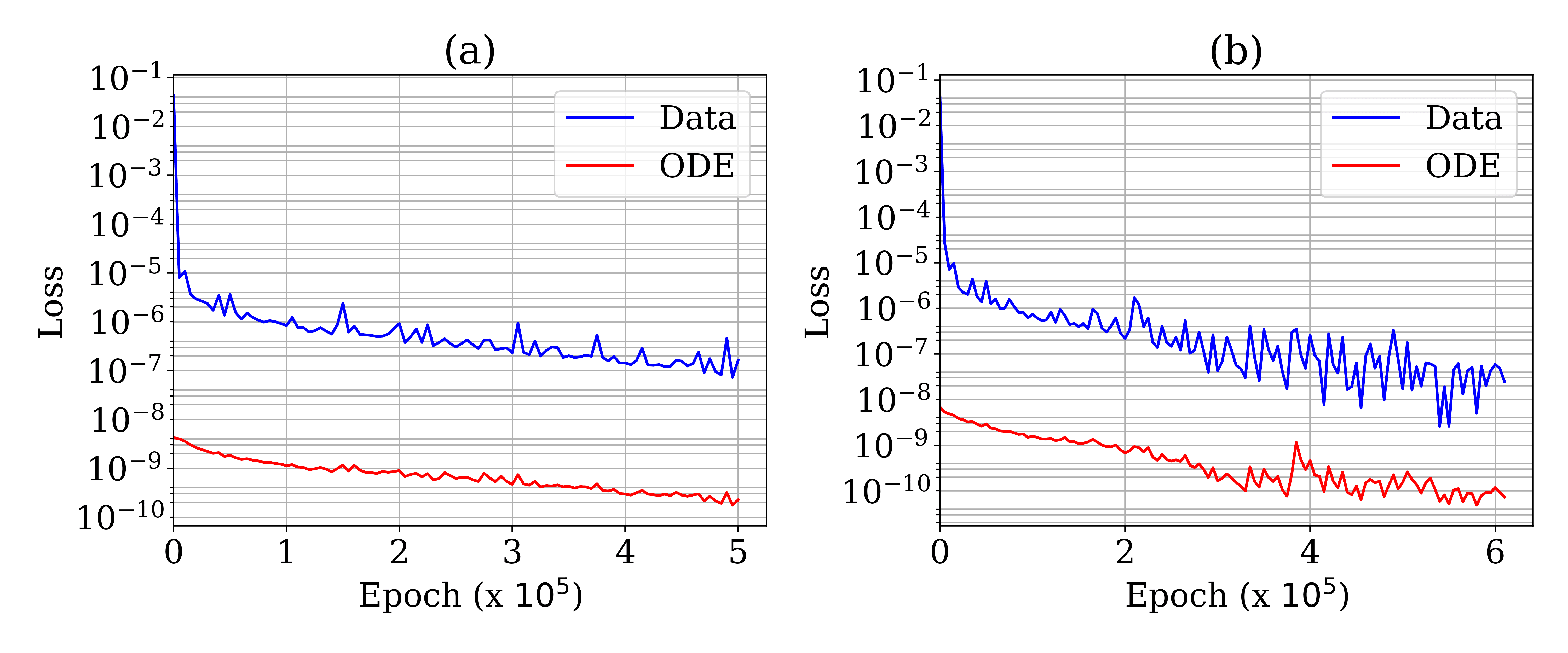}
    \caption{Convergence history showing both data loss and residual loss from ODE for bubbles simulated using (a) R-P equation (b) K-M equation}
    \label{fig:convergenceRPKM}
\end{figure}
\noindent The training loss is shown in Figure~\ref{fig:convergenceRPKM}, where the neural network loss \(\mathcal{L}_{data}\) decreases rapidly at the beginning and gradually continues to decline as training progresses, while the governing equation loss \(\mathcal{L}_{\text{ode}}\) steadily decreases throughout.

\subsubsection{Cross validation for generalization}
\begin{figure}[htpb]
\centering
\includegraphics[width=0.3\linewidth]{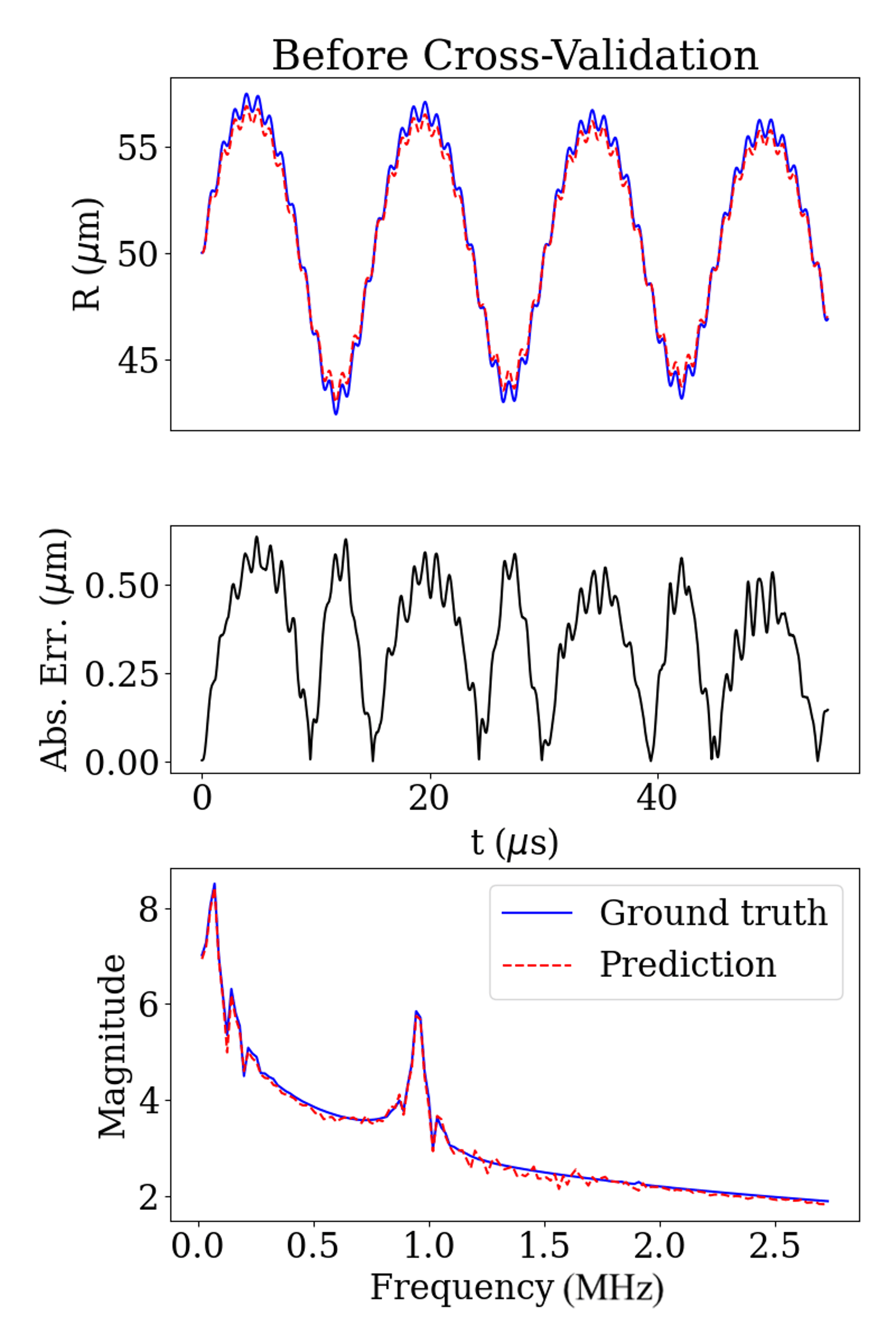}
\includegraphics[width=0.3\linewidth]{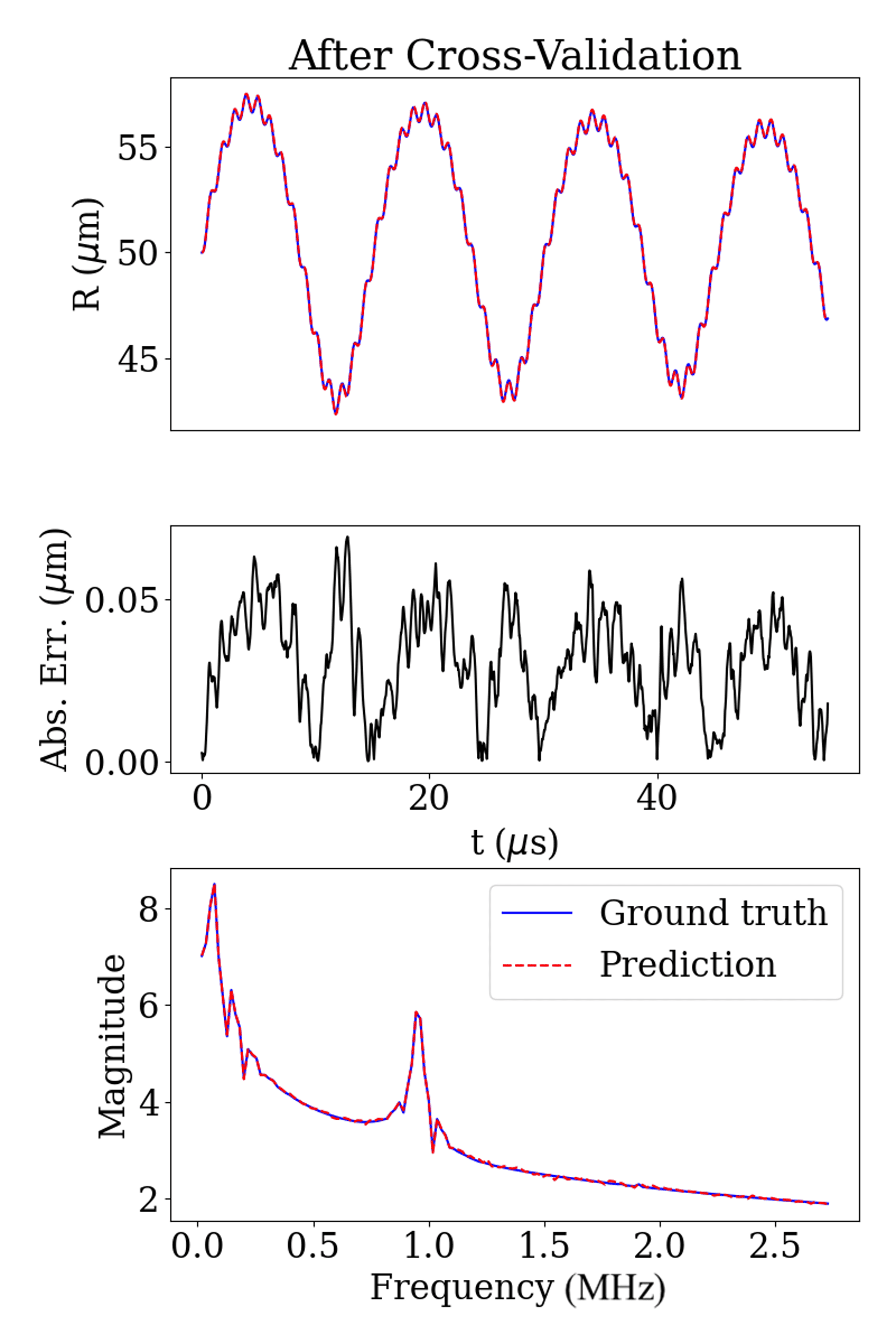}
\caption{Effect of cross-validation on model generalization. The case of a bubble driven at \(954~\text{KHz}\) and \(9 \times 10^5~\text{Pa}\) shows significant improvement in prediction accuracy after applying 5-fold cross-validation.}
\label{fig:CV}
\end{figure}
Some validation cases, such as the bubble simulated with a frequency of \(954~\text{kHz}\) and an amplitude of \(9 \times 10^5~\text{Pa}\), exhibit noticeable discrepancies between the ground truth and the model prediction (Figure~\ref{fig:CV} (left)). This may result from a lack of similar features in the training data, leading to poor generalization in this specific case. To address this limitation, a 5-fold cross-validation (CV) was performed to enhance the model’s robustness.

Comparison between the directly trained model and the CV-trained model shows a marked improvement: the absolute error range decreases by an order of magnitude, from \([0, 0.6]\) to approximately \([0, 0.06]\). Notably, the frequency-domain results indicate that even the directly trained model provides sufficient phase and frequency resolution to capture the driving frequency, demonstrating the effectiveness of the Rowdy activation function. However, it fails to accurately predict the magnitude of the driving-frequency component-that is, the strength of the frequency response-resulting in a mismatch between the predicted and true amplitudes. These findings suggest that the primary source of discrepancy arises from data sparsity rather than an inherent limitation in learning frequency characteristics. Accordingly, applying cross-validation mitigates this issue by improving the model’s generalization performance in underrepresented regions of the parameter space; see, Figure~\ref{fig:CV} (right).

\subsubsection{Out-of-distribution Results}
We perform a systematic study of frequency-domain, time-domain, and pressure-amplitude extrapolation results. 
\begin{figure}[htpb]
\centering
\centering
\includegraphics[width=0.2\linewidth]{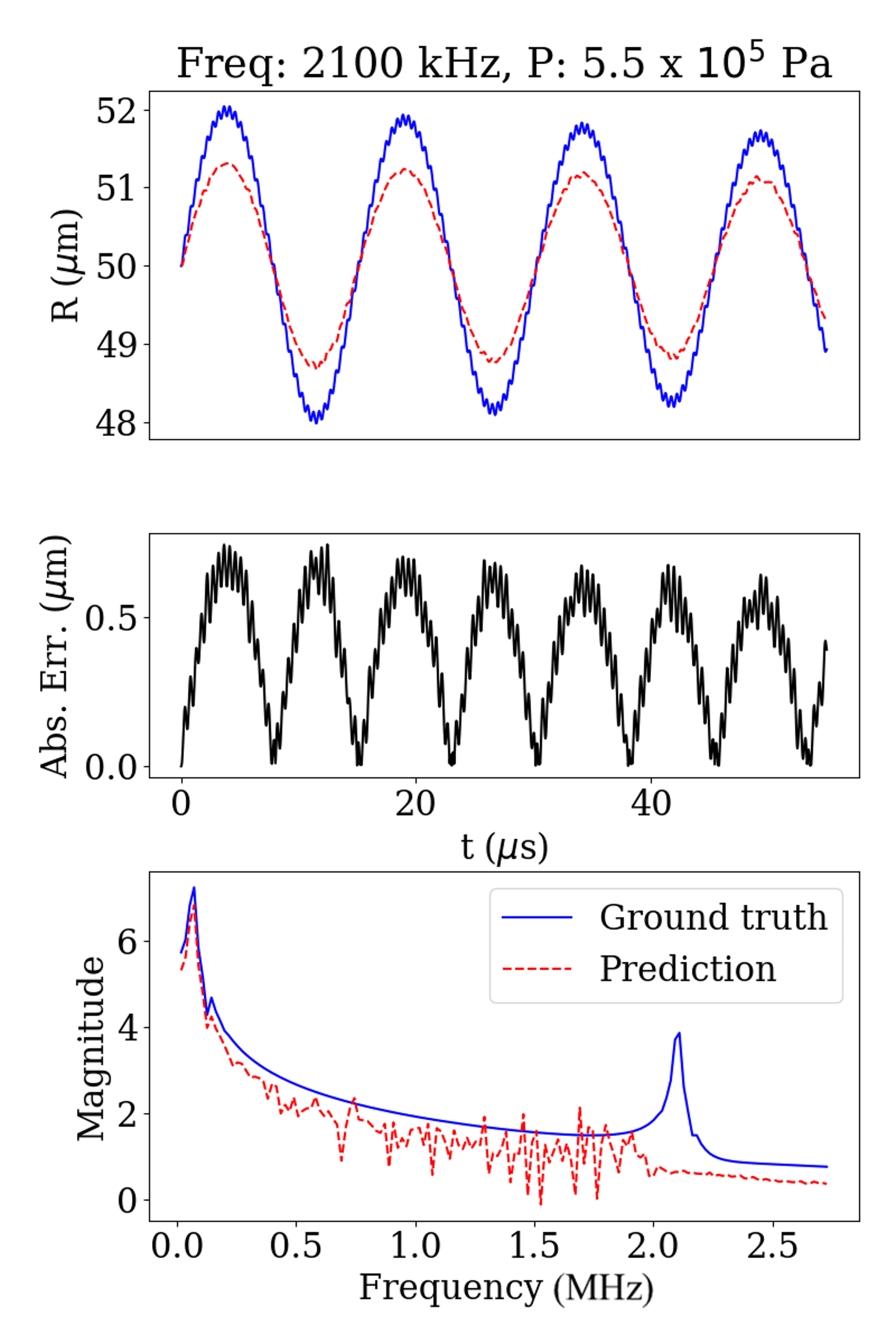}
\centering
\includegraphics[width=0.2\linewidth]{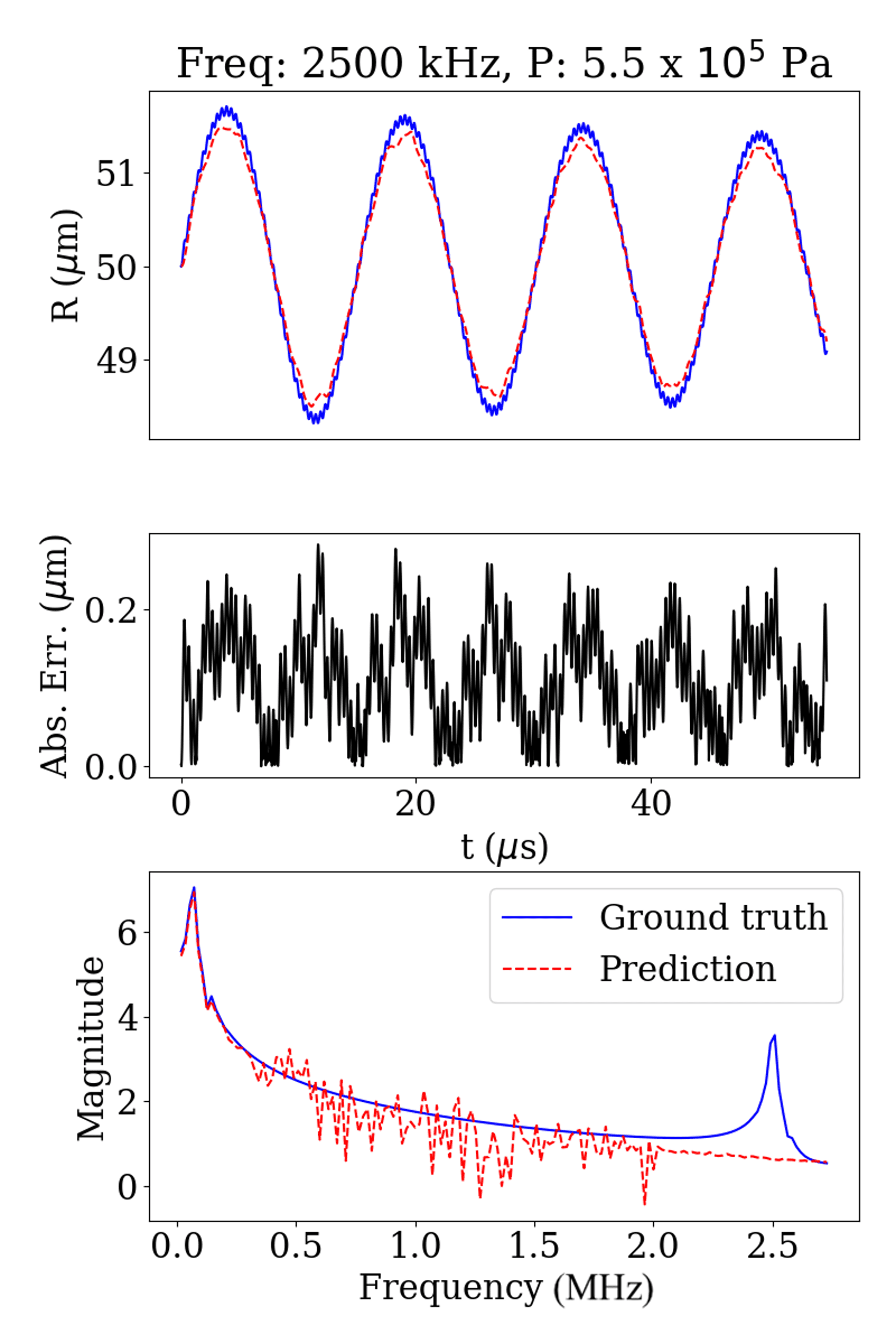}
 \includegraphics[trim=1cm 17cm 1cm 1cm, clip, scale=0.45]{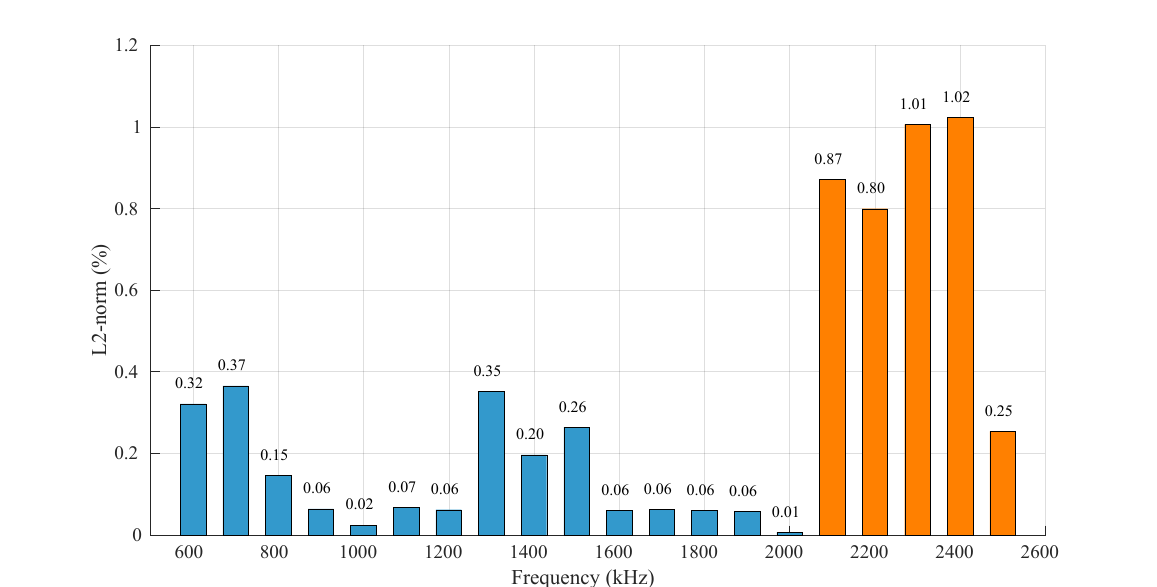}
\caption{First two columns: Frequency-domain extrapolation results for bubbles driven at \(2100~\text{KHz}\) and \(2500~\text{KHz}\), with a fixed amplitude of \(5.5 \times 10^5~\text{Pa}\). Right figure: L2-norm errors for different frequencies. Orange bars represents out-of-distribution frequencies.}
\label{fig:extrapolation5.5}
\end{figure}
Two out-of-distribution cases at an amplitude of \(5.5\times10^{5}~\text{Pa}\) are presented in Figure~\ref{fig:extrapolation5.5} (first two columns), corresponding to driving frequencies of \(2100~\text{kHz}\) and \(2500~\text{kHz}\). Since these cases fall outside the function space used for operator learning, it is anticipated that the model may struggle to generalize. Indeed, both cases fail to accurately capture the driving frequency in the frequency domain. Figure~\ref{fig:extrapolation5.5} (Right) shows L2-norm errors for different frequencies.
\begin{figure}[htpb]
\centering
\centering
\includegraphics[width=0.2\linewidth]{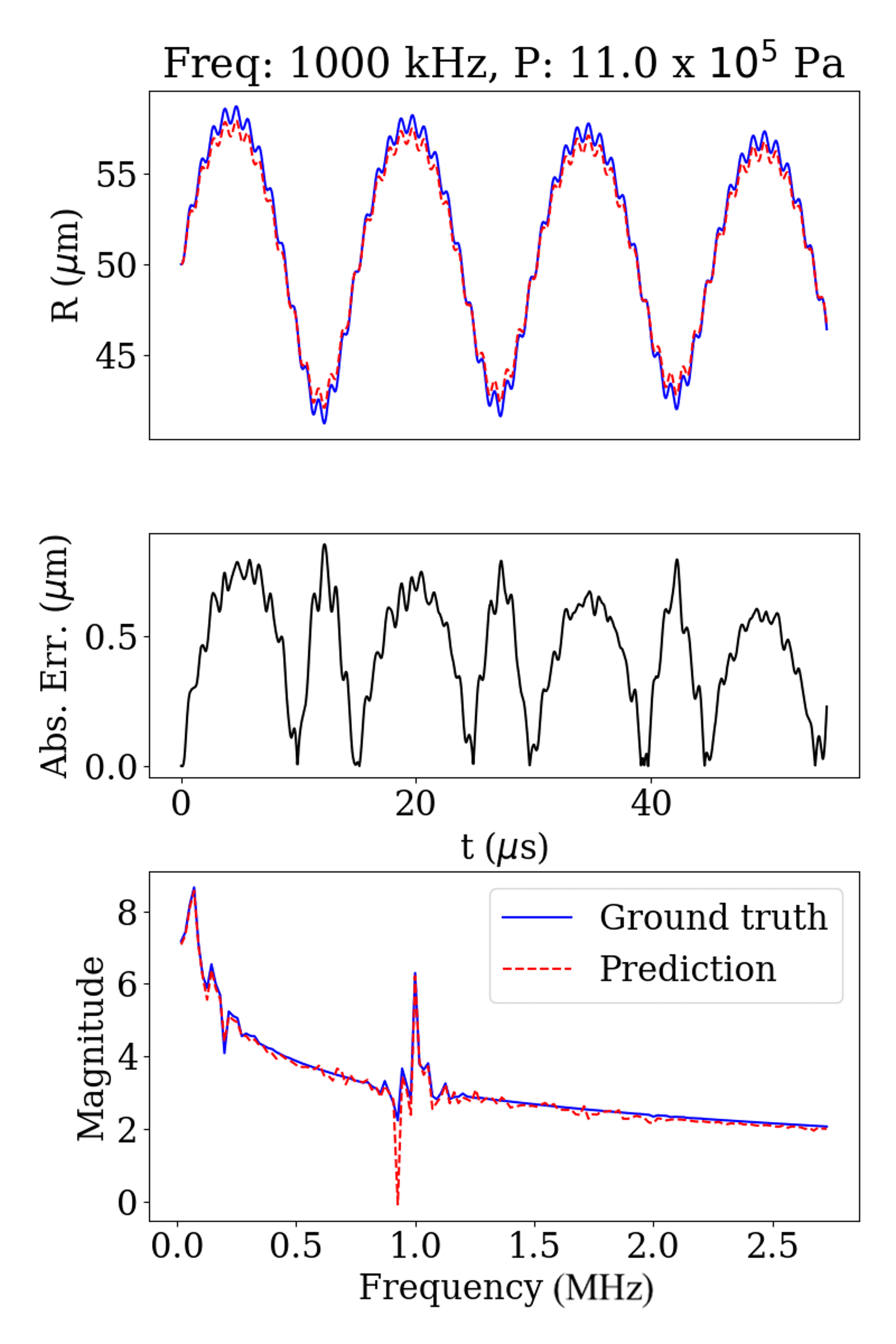}
\centering
\includegraphics[width=0.2\linewidth]{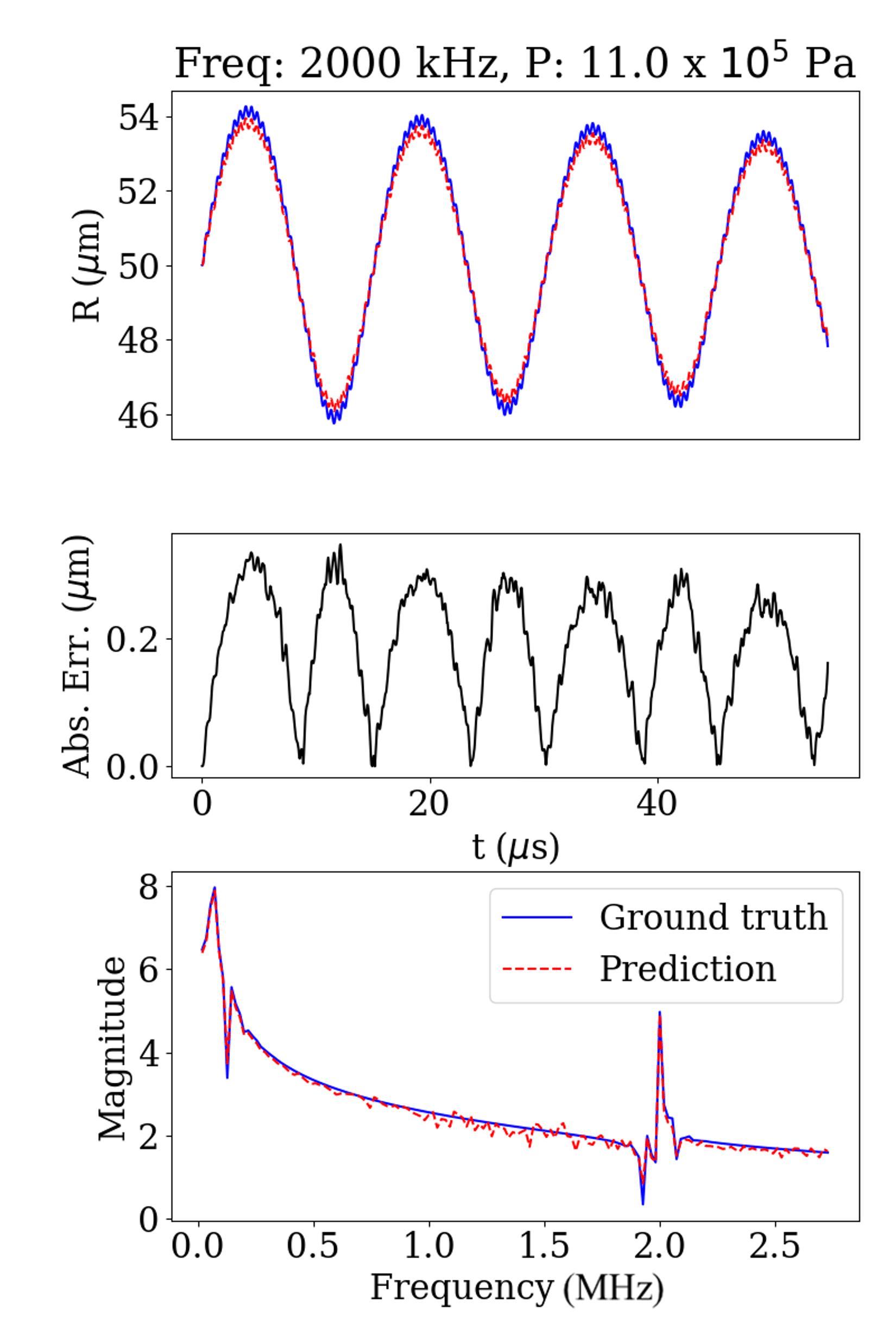}
\includegraphics[width=0.2\linewidth]{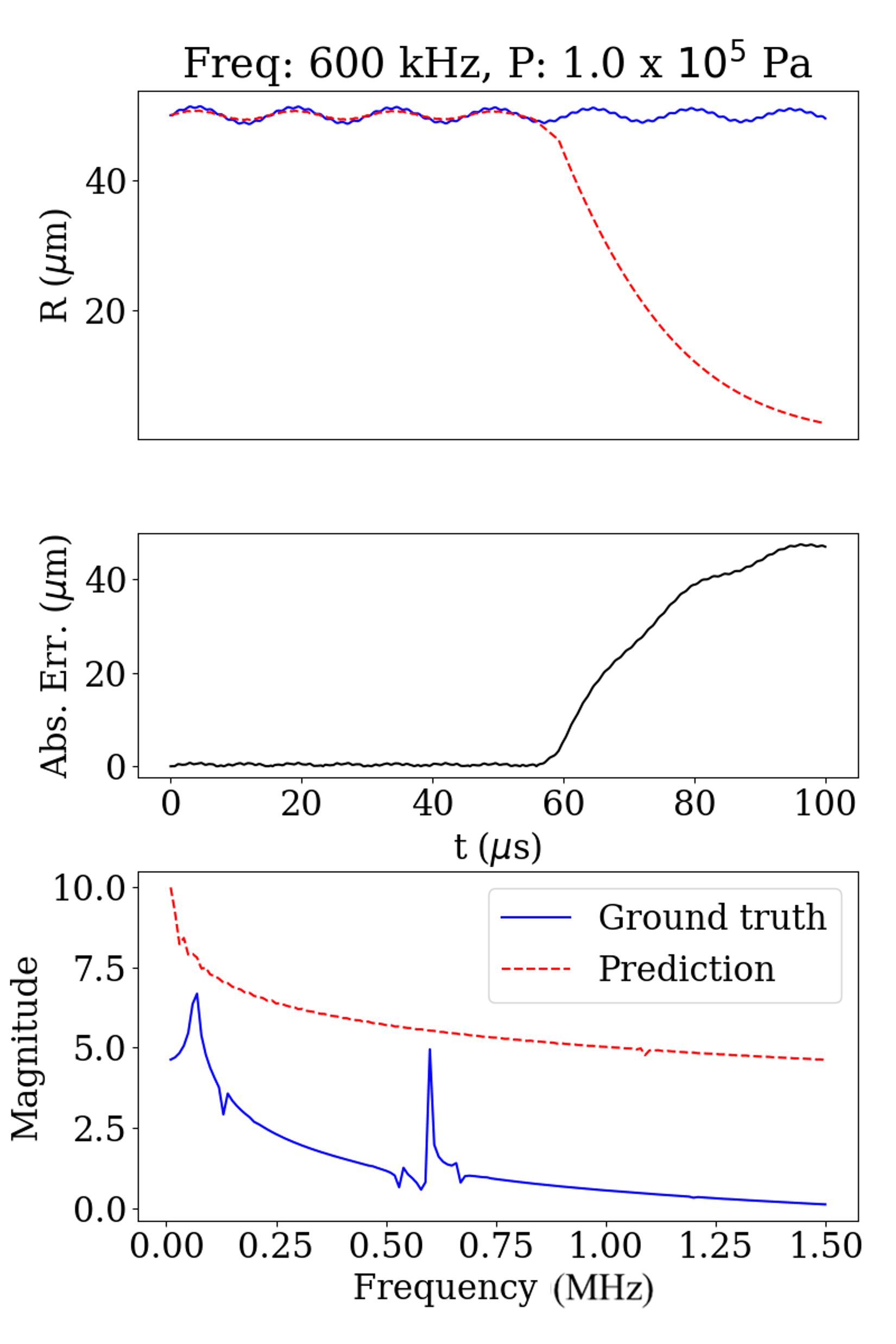}
\centering
\includegraphics[width=0.2\linewidth]{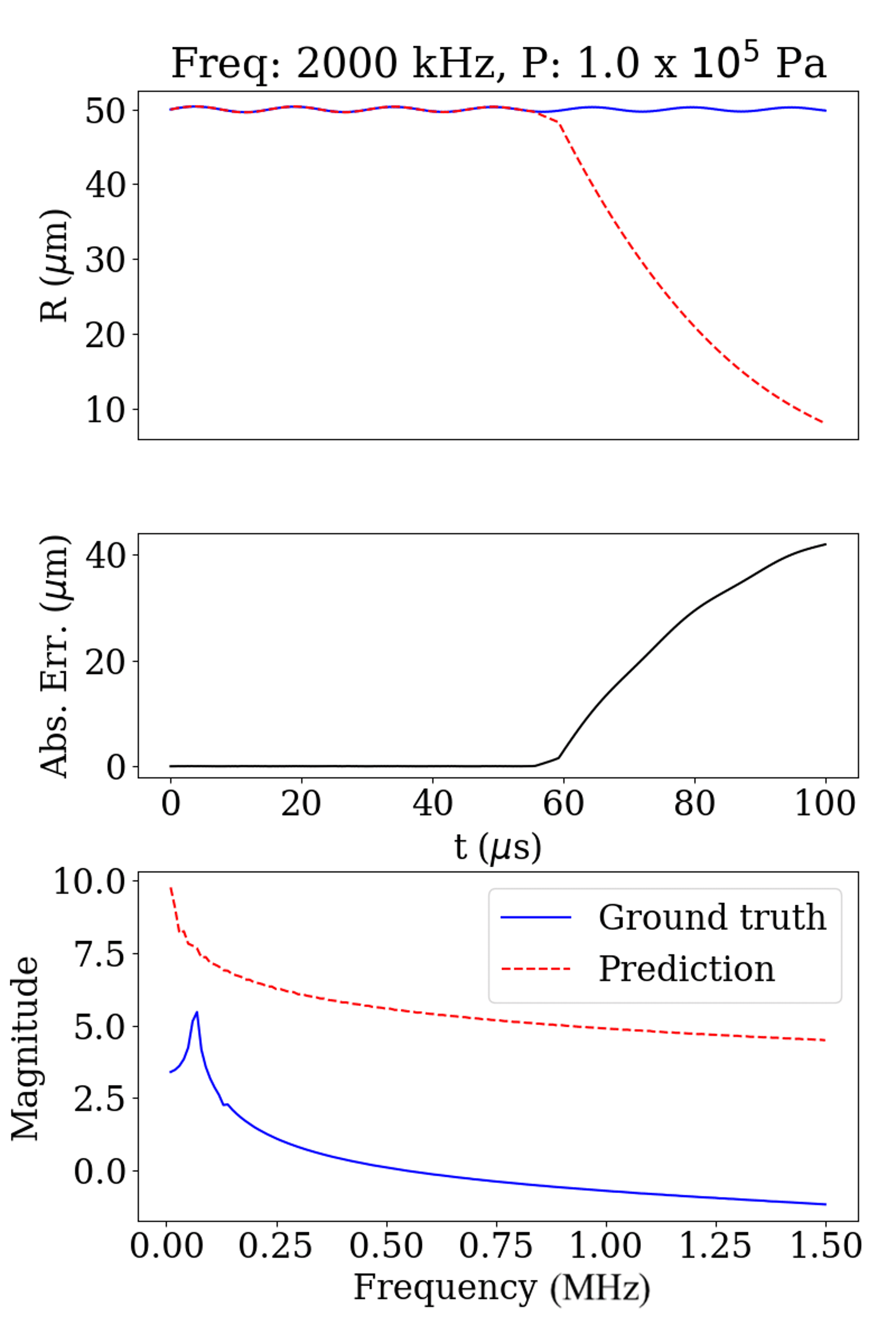}
\caption{First and Second Columns: Pressure amplitude extrapolation results for bubbles driven at \(1000~\text{KHz}\) and \(2000~\text{KHz}\), with a fixed amplitude of \(11 \times 10^5~\text{Pa}\). Third and Forth Columns: Time-domain extrapolation results for bubble dynamics over \(100\mu s\) driven at \(600~\text{KHz}\) and \(2000~\text{KHz}\), with a fixed amplitude of \(1 \times 10^5~\text{Pa}\).}
\label{fig:extrapolation_11}
\end{figure}
For the amplitude-based out-of-distribution test, two cases with an amplitude of \(1.1\times10^{6}~\text{Pa}\) at frequencies of \(1000~\text{kHz}\) and \(2000~\text{kHz}\) are shown in Figure~\ref{fig:extrapolation_11} (the first and second columns). In both cases, the absolute errors are reasonable, ranging approximately within \([0,\,0.8]\) and \([0,\,0.4]\). However, the driving frequencies are still accurately captured in the frequency domain. This suggests that the source of error is primarily due to the amplitude, as the model has not encountered similar features during training. These findings further confirm the sensitivity of the model to unseen amplitude ranges.

A final out-of-distribution study is conducted for an extended simulated time range, increasing the original duration from \(55~\mu\text{s}\) to \(100~\mu\text{s}\). The results are shown in Figure~\ref{fig:extrapolation_11} (the third and fourth columns). In this scenario, the model fails to make accurate predictions even within the time interval it was previously trained on (i.e., from 0 to \(55~\mu\text{s}\)). Beyond this point, the predicted radius begins to shrink significantly, indicating that the model has entered a region outside its learned function space. Further evidence of this failure is observed in the frequency domain, where the model is unable to capture either the natural or driving frequency, let alone their magnitudes. These results highlight the model’s limitation in generalizing to extended temporal domains.

\subsection{Two-step DeepONet Training for K-M Equation with Multi Initial Radii}
\begin{figure}[htpb]
    \centering
    \includegraphics[trim=0cm 0cm 0cm 0cm, clip, width=1\linewidth]{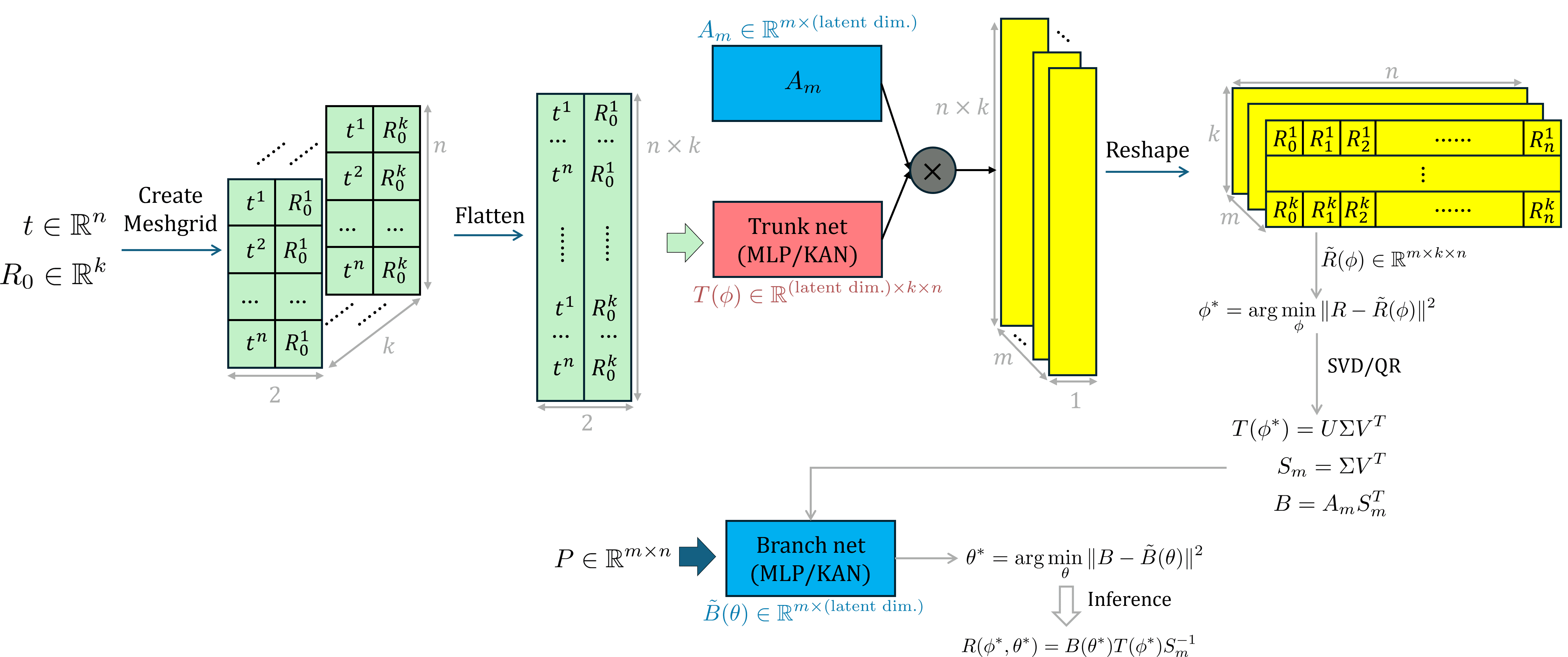}
    \caption{Implementation of multi-initial radii scenario.}
    \label{fig:multiR_SKE}
\end{figure}

In this section, we focus on the two-dimensional basis constructed by the trunk network, as illustrated in Figure~\ref{fig:multiR_SKE}. Given \(n\) time sampling points and \(k\) different initial radii \(R_0\), meshgrids of size \(n \times k\) (time and radius) are generated for each \(R_0\), resulting in \(k\) meshgrids. These meshgrids are flattened and stacked into a single input of size \((n \times k) \times 2\) for the trunk network.
The branch network receives pressure profiles of size \(m \times n_p\), where \(m\) is the batch size and \(n_p\) is the number of pressure sampling points. The model outputs predictions of size \(m \times (n \times k)\), which are then reshaped to \(m \times k \times n\), representing bubble dynamics over time for each initial radius.

\label{session:multR}
To demonstrate the model's capability, five initial radii, \(50\,\mu\text{m}\), \(60\,\mu\text{m}\), \(70\,\mu\text{m}\), \(80\,\mu\text{m}\), and \(90\,\mu\text{m}\), are considered. Pressure amplitudes and frequencies are sampled within the same ranges specified in Table~\ref{tab:DoE}, using 10 amplitude values and 30 frequency values, resulting in 300 distinct pressure profiles. For each initial radius, 300 bubble profiles are generated, yielding a total of 1500 samples. Among these, 80\% are used for training and 20\% for validation. Consequently, the network architecture is adjusted to account for the varying natural frequencies associated with different initial radii. The Rowdy activation function incorporates four sinusoidal components to better capture these dynamics. The loss function weights are set to \(w_{\text{data}} = 1\) and \(w_{\text{phys}} = 1000\), and an additional term enforcing the initial radius as an initial condition (IC) is introduced with a weight of \(w_{\text{IC}} = 1\).

\begin{figure}[htpb]
    \centering
    \includegraphics[trim=0.5cm 10cm 1cm 1cm, clip=true, scale=0.45]{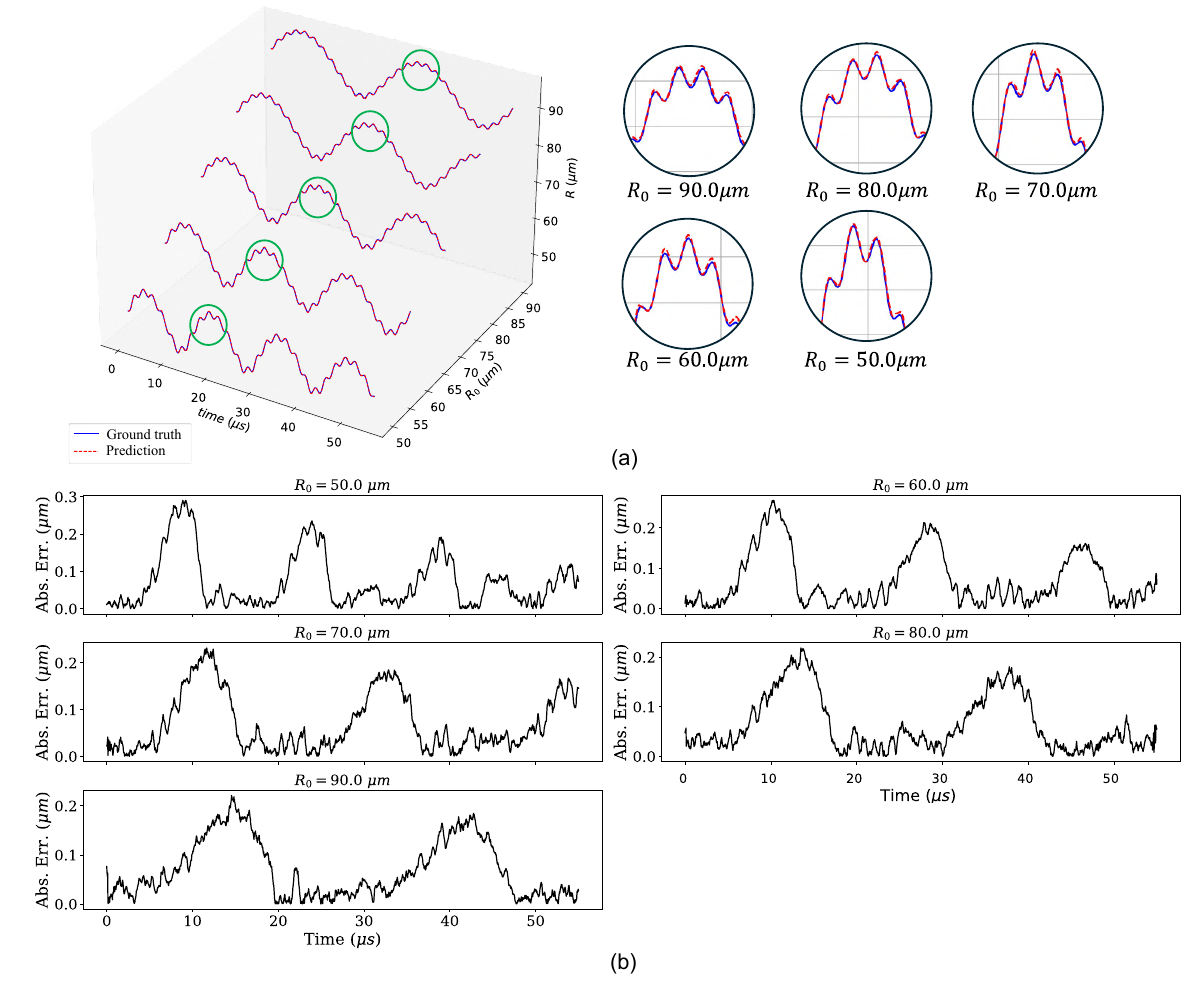}
    \caption{Prediction of bubble dynamics with five different initial radii, under pressure with a frequency of \(467~\text{KHz}\) and an amplitude of \(4\times 10^5~\text{Pa}\). (a) Bubble profile with the second peak highlighted in green circles and shown on the right. (b) Absolute error.}
    \label{fig:multiR_lowf}
\end{figure}
A prediction for bubbles subjected to a pressure oscillation with a frequency of \(467~\text{kHz}\) and an amplitude of \(4 \times 10^5~\text{Pa}\) is presented in Figure~\ref{fig:multiR_lowf}. In Figure~\ref{fig:multiR_lowf}(a), the second peaks of all bubble profiles are highlighted with green circles and magnified on the right for clearer visualization. As shown in Figure~\ref{fig:multiR_lowf}(b), the predicted profiles exhibit excellent agreement with the ground truth, with maximum absolute deviations of approximately \(0.3\,\mu\text{m}\) (0.40\%) for the \(50\,\mu\text{m}\) radius and \(0.2\,\mu\text{m}\) (0.22\%) for the \(90\,\mu\text{m}\) radius.

\begin{figure}[htpb]
    \centering
    \includegraphics[trim=0.5cm 10.5cm 1cm 1cm, clip=true,scale=0.45]{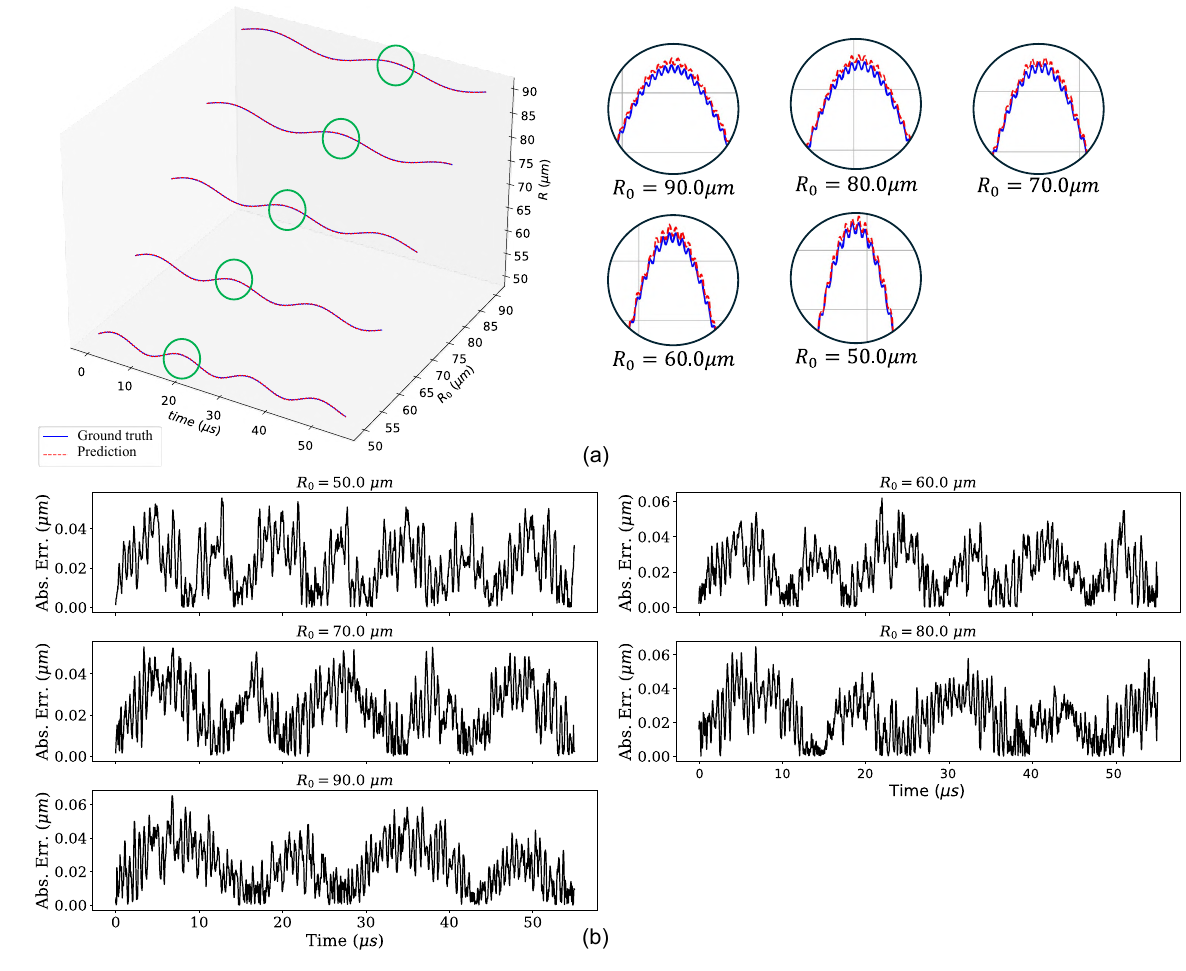}
    \caption{Prediction of bubble dynamics with five different initial radii, under pressure with a frequency of \(1800~\text{KHz}\) and an amplitude of \(3\times 10^5~\text{Pa}\). (a) Bubble profile with the second peak highlighted in green circles and shown on the right. (b) Absolute error.}
    \label{fig:multiR_highf}
\end{figure}

\begin{figure}[htpb]
    \centering
    \includegraphics[trim=0cm 0cm 0cm 0cm, clip=true, scale=0.4]{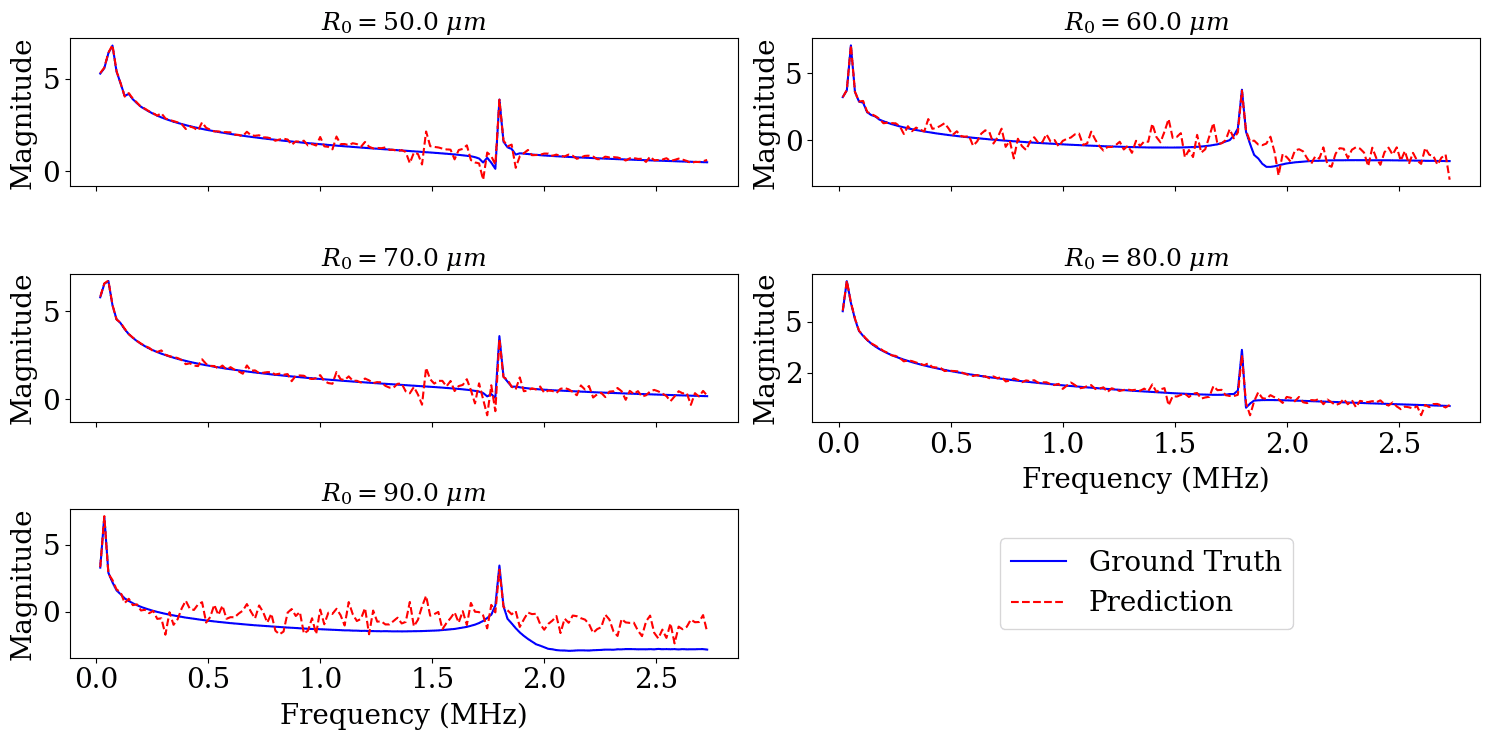}
    \caption{Fast Fourier Transform of predicted bubble dynamics with five different initial radii, under pressure with a frequency of \(1800~\text{KHz}\) and an amplitude of \(3\times 10^5~\text{Pa}\).}
    \label{fig:multiR_highf_fft}
\end{figure}

A second prediction, shown in Figure~\ref{fig:multiR_highf}, considers bubbles subjected to a higher-frequency pressure oscillation of \(1800~\text{KHz}\) with an amplitude of \(3 \times 10^5~\text{Pa}\). As highlighted in Figure~\ref{fig:multiR_highf}(a), discrepancies in amplitude between the predicted and true profiles are observed at the second peaks (highlighted by green circles and magnified views). Nevertheless, the maximum absolute errors remain below \(0.061\,\mu\text{m}\), corresponding to 0.12\% and 0.07\% for the \(50\,\mu\text{m}\) and \(90\,\mu\text{m}\) radii, well within an acceptable range. Furthermore, FFT in Figure~\ref{fig:multiR_highf_fft} confirms that the model accurately captures both the natural and driving frequencies.

\subsubsection{Ablation Study}
\begin{table}[htbp]
\centering
\caption{Ablation study of Rowdy terms in the Two-Step DeepONet under a pressure amplitude of \(1\times10^{6}\,\text{Pa}\) and a frequency of \(2000\,\text{kHz}\).}
\label{tab:Rowdy_ablation}
\begin{tabular}{lc}
\hline
\textbf{Model} & \textbf{L2-norm Error} \\ 
\hline
2-Step DeepONet & 0.287\% \\
2-Step DeepONet + 1 term (Rowdy) & 0.043\% \\
2-Step DeepONet + 2 terms (Rowdy) & 0.020\% \\
2-Step DeepONet + 3 terms (Rowdy) & 0.051\% \\
\hline
\end{tabular}
\end{table}
An ablation study is conducted to assess the effectiveness of the Rowdy activation terms; see Table \ref{tab:Rowdy_ablation}. Four experiments are performed using the two-step DeepONet architecture with varying numbers of Rowdy activations, validated against bubble dynamics subjected to a pressure amplitude of \(1\times10^{6}\,\text{Pa}\) and a high driving frequency of 2000\,kHz. When no Rowdy activation is employed, the model captures only the natural frequency of the bubble but fails to reproduce the driving frequency, as also reported in Ref.~\cite{zhang2025bubble}, resulting in the highest prediction error of 0.287\%. As the number of Rowdy terms increases, the model’s accuracy improves and the prediction error decreases. However, when three Rowdy terms are introduced, overfitting occurs, leading to a slight increase in error.

\subsection{Two-step DeepOKAN with Multi Initial Radii}

Spectral bias was introduced by Rahaman et al. \cite{rahaman2019spectral} in the context of deep networks where these networks in spite of following the universal approximation theorem solely aims to approximate the low frequency components of a function often neglecting the essential high frequency features present in an underlying function. The previous section dealt with overcoming the spectral bias while learning the high frequency bubble dynamics using two-step DeepONet framework with the help of Rowdy activation functions. Likewise, several recent developments have focused on addressing the spectral bias mainly within the context of MLP; due to the nascency of KAN there has not been any work that tackled spectral bias using the KAN architecture. The two-step DeepOKAN model was trained on an NVIDIA V100 GPU with 64 GB of system RAM.

\begin{figure}[htpb]
\centering
{
\centering
\includegraphics[width=0.5\linewidth]{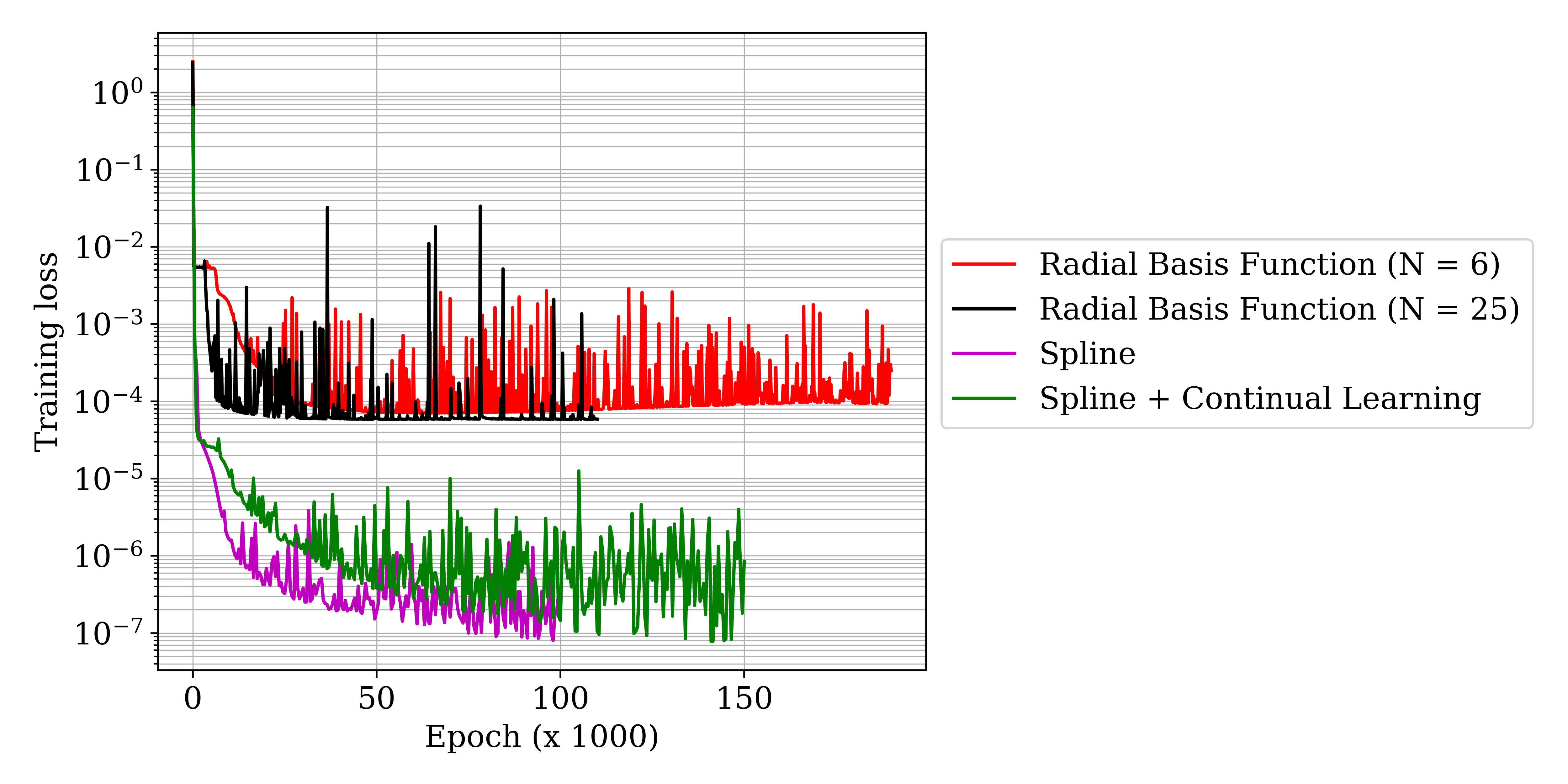}
}
    \caption{The training performance of different basis functions for trunk network training for high frequency bubble dynamics. Here, $N$ in the context of radial basis functions (RBF) means the number of functions at a given node. Spline basis has polynomial order ($k$) = 2 and grid size ($G$) = 40. The training undergoes early stopping under convergence.}
\label{fig:trunk_train_2sdeepokan}
\end{figure}

\begin{figure}[htpb]
\centering
{
\centering
\includegraphics[width=0.3\linewidth]{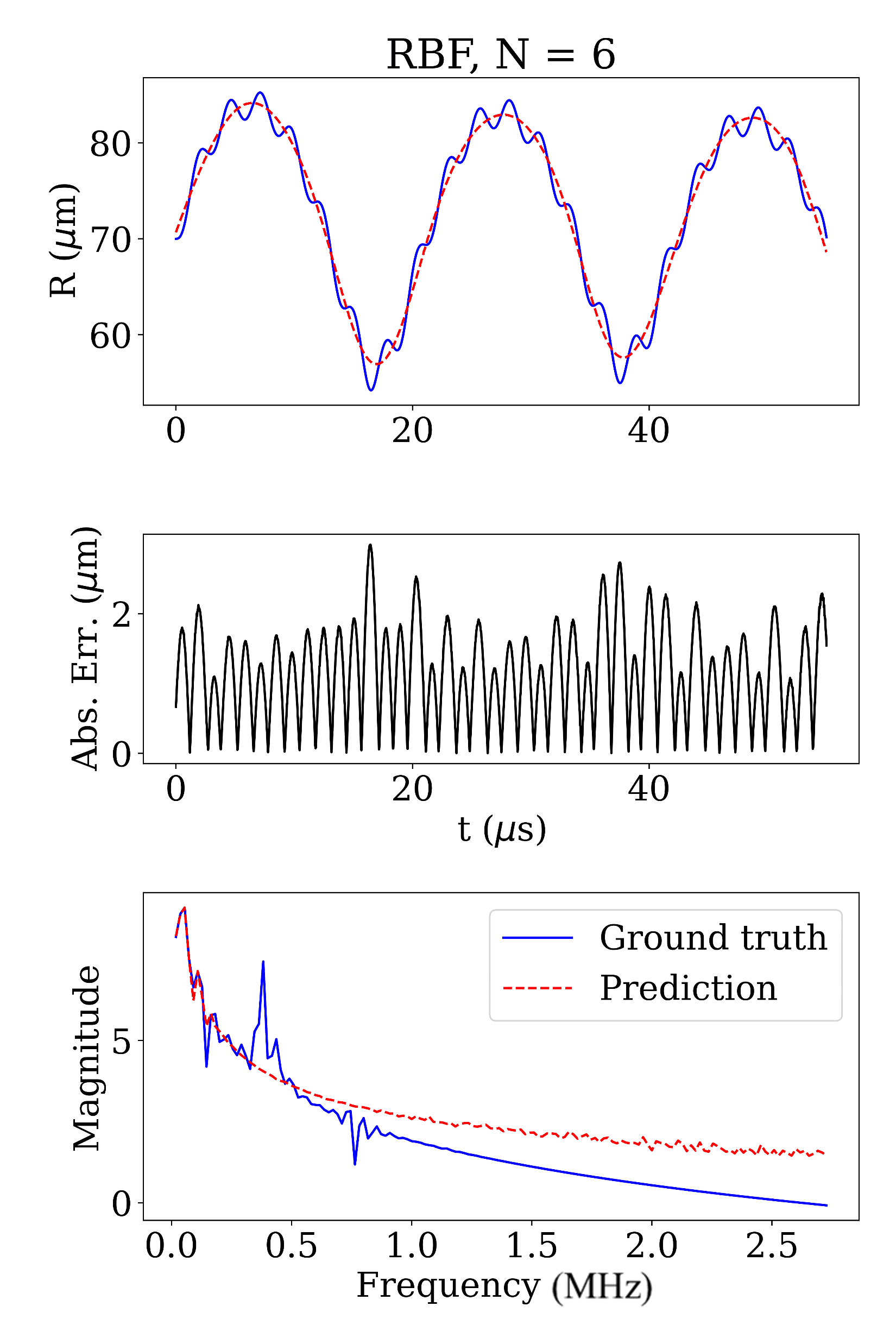}
}
\hspace{0.001cm}
\centering
{
\centering
\includegraphics[width=0.3\linewidth]{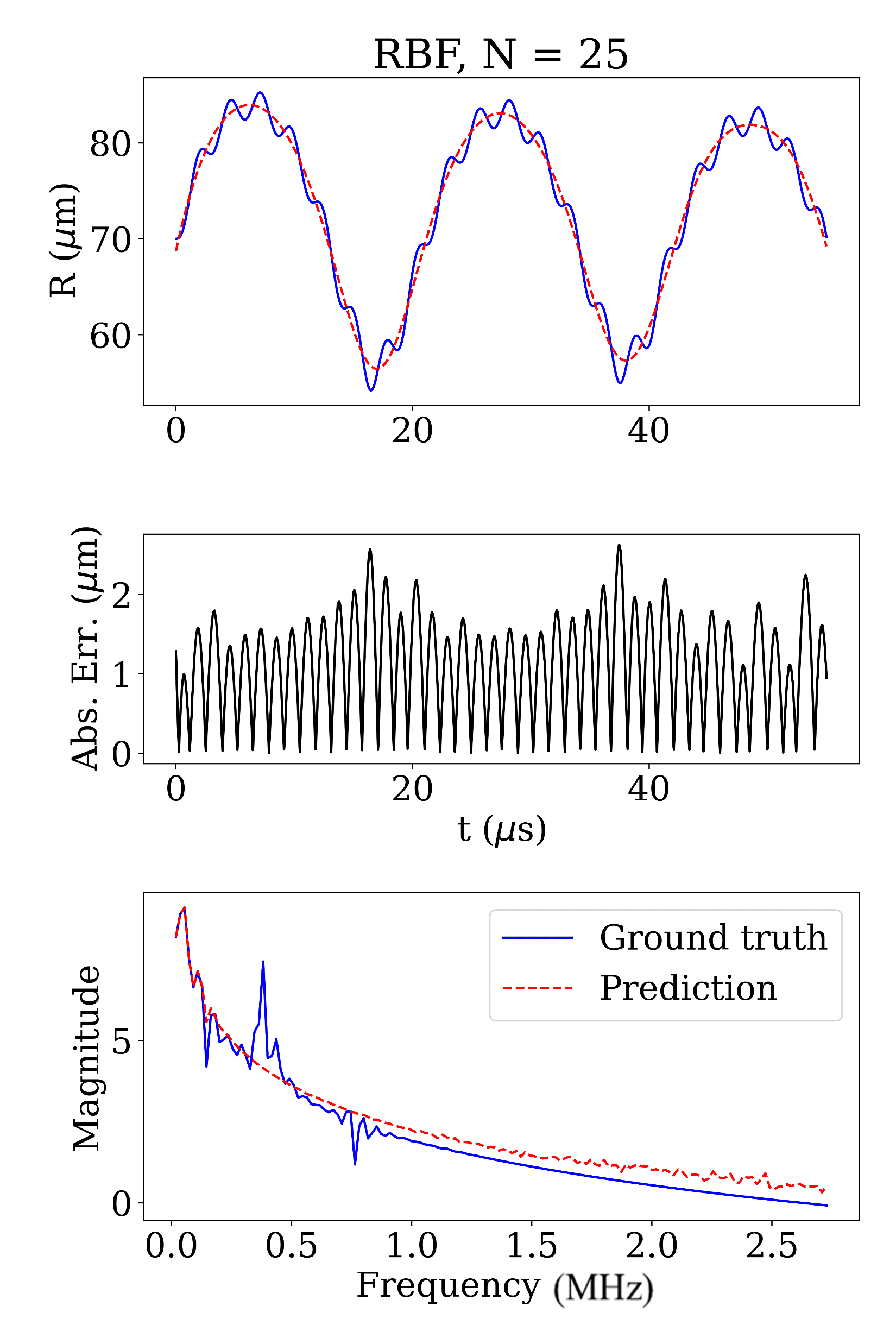}
}
\hspace{0.001cm}
\centering
{
\centering
\includegraphics[width=0.3\linewidth]{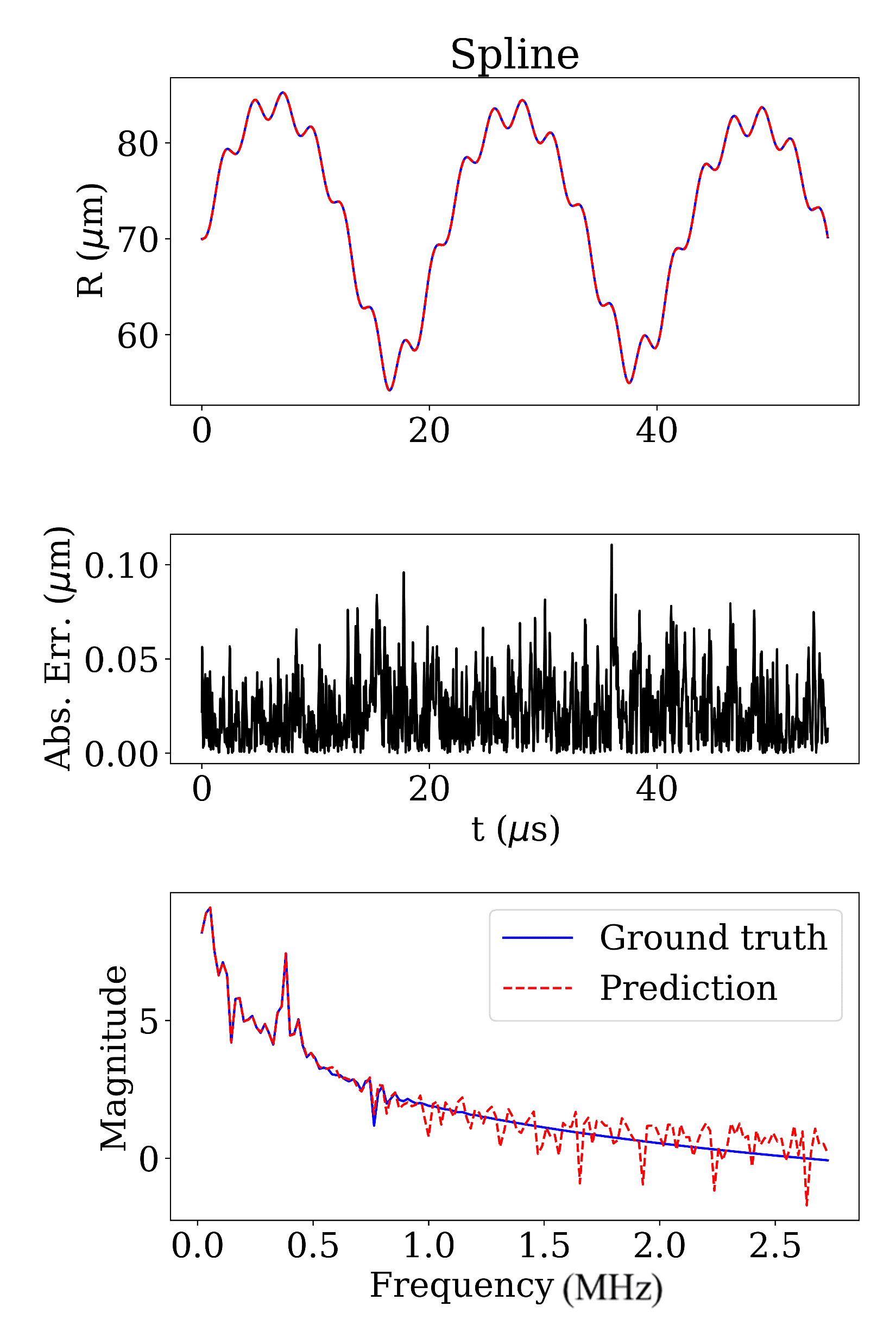}
}
\caption{The performance of different basis functions while learning the bubble dynamics under pressure with a frequency of $491$ kHz and amplitude of $8 \times 10^5$ Pa. (\textit{left}) RBF with $N=6$ (\textit{middle}) RBF with $N=25$ (\textit{right}) Spline with $k=2$ and $G=40$. Note that these are intermediate results ($\tilde{R}(\phi)$) solely from trunk net, $\tilde{R}(\phi) = A_mT(\phi)^T$ to justify the utilization of a specific formulation of the basis function within the KAN architecture.}
\label{fig:basis_lf}
\end{figure}

\begin{figure}[htpb]
\centering
{
\centering
\includegraphics[width=0.3\linewidth]{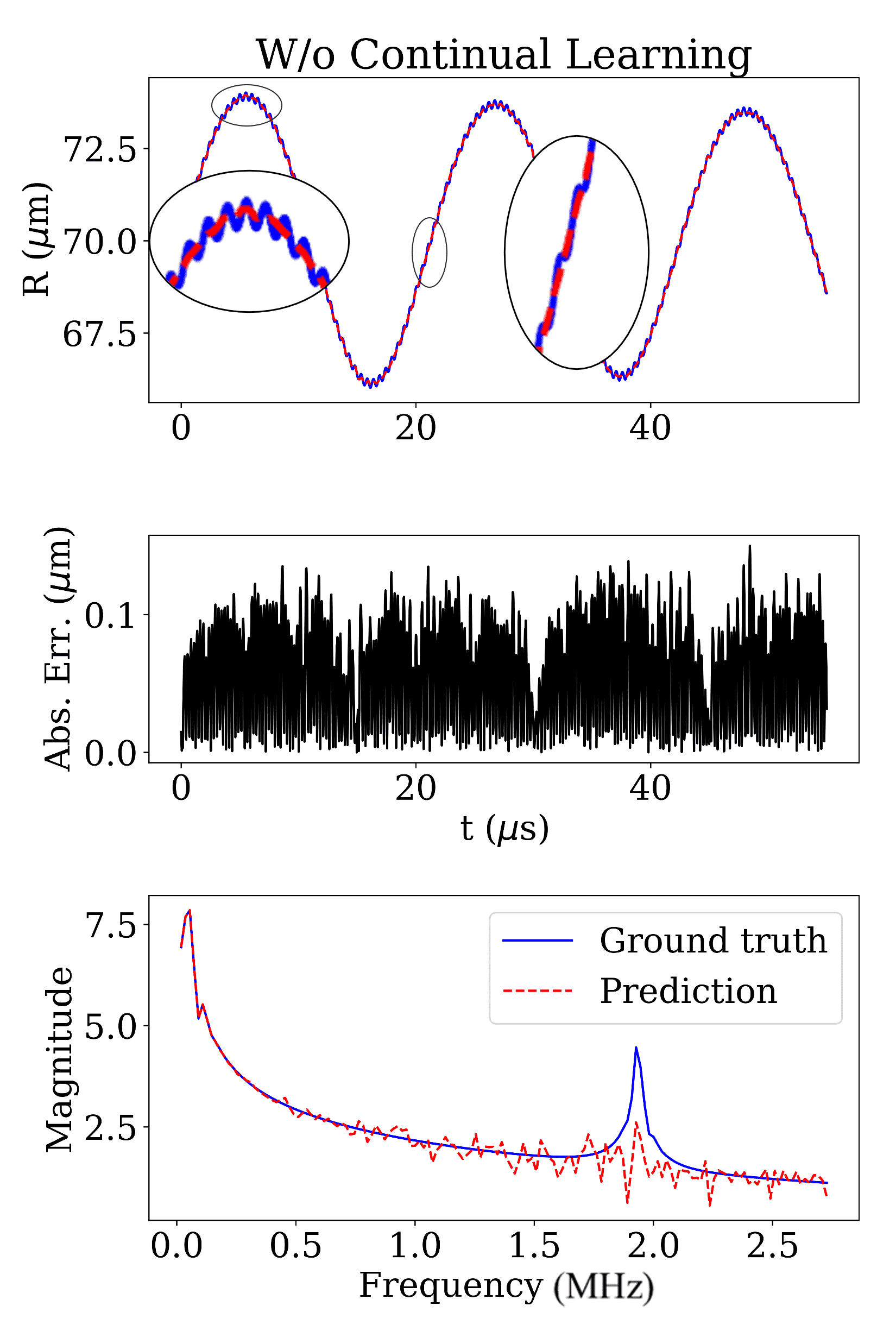}
}
\hspace{0.001cm}
\centering
{
\centering
\includegraphics[width=0.3\linewidth]{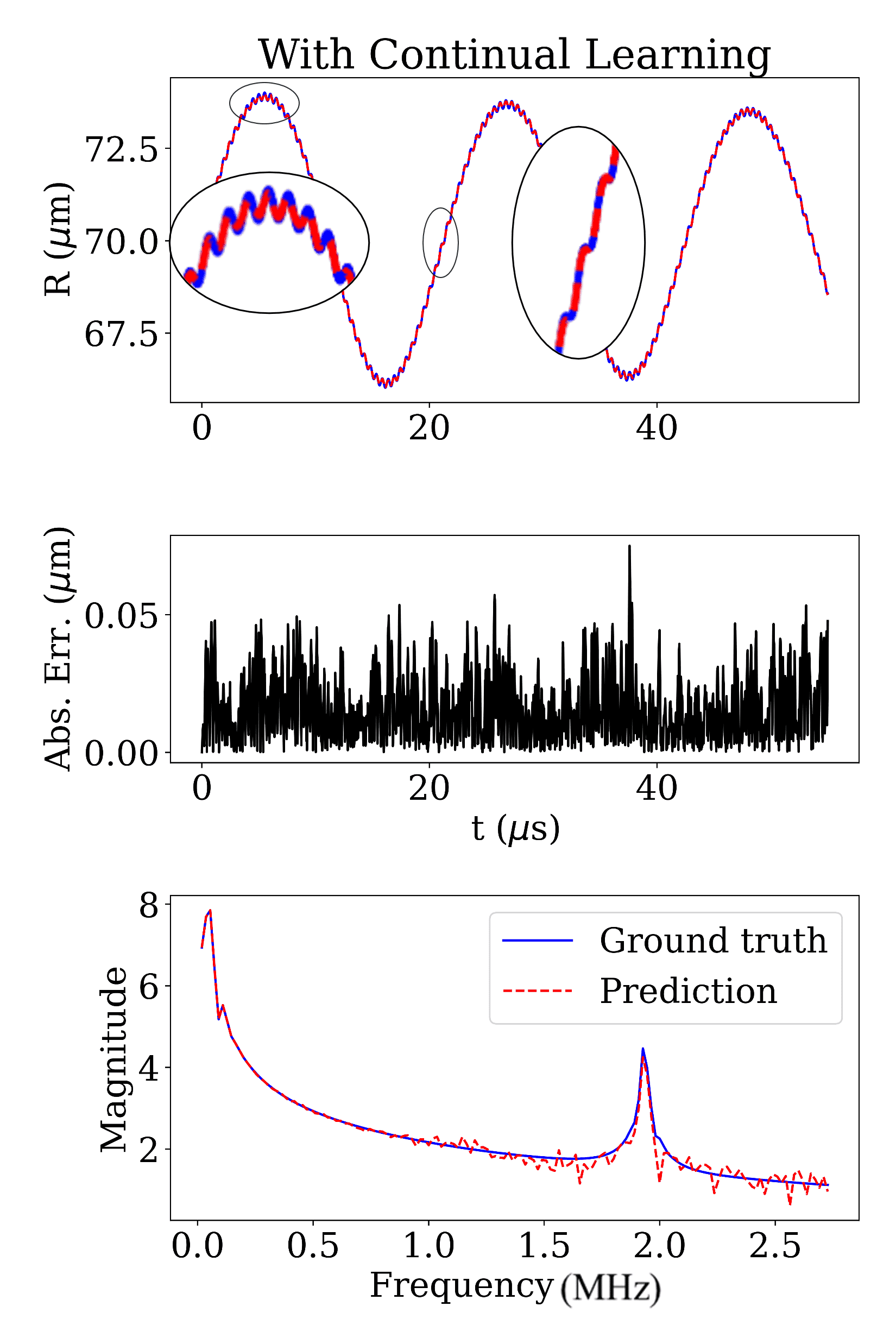}
}
\caption{Comparison of the training performance of trunk network based on spline basis with continual learning and without continual learning while learning bubble dynamics under pressure with a frequency of $1928$ kHz and amplitude of $11 \times 10^5$ Pa. Note that these are intermediate results ($\tilde{R}(\phi)$) for trunk net, $\tilde{R}(\phi) = A_mT(\phi)^T$.}
\label{fig:spline_cont}
\end{figure}

Figure \ref{fig:trunk_train_2sdeepokan} showcases the training performance of the trunk network for different basis functions utilized in the KAN framework. Here, we note that spline basis with continual learning outperforms all the other configurations of the basis functions. Increasing the number of funtions $N$ from 6 to 25 for the RBF \cite{abueidda2025deepokan} basis is observed to have negligible effects on the training performance with added computational cost incurred from over-parametrization. In a way, this is confirmed by the results shown in Figure \ref{fig:basis_lf} where RBF basis unlike the spine basis fails to learn the internal high frequency features of the bubble dynamics even at lower frequency ranges. Although, Figure \ref{fig:trunk_train_2sdeepokan} suggests almost similar training performance for spline basis with and without continual training, it is does not reveal the actual improvement. Figure \ref{fig:spline_cont} showcases this improvement for a high frequency, high pressure scenario where the spline basis without continual learning only captures the localized global frequencies while avoiding the high-frequency features of the bubble dynamics. On the other hand, with the use of continual training, we observe that the model captures all the dominant frequencies, with an order-of-magnitude improvement in the training error, as shown in Figure \ref{fig:spline_cont}.

Based on the above observations, it can be inferred that KAN as an independent architecture like MLP, may not be able to overcome spectral bias solely by itself. However, with a slight modification in the training via continual learning we are able to overcome the spectral bias without the utilization of any additional trainable parameters via activation functions unlike the two-step DeepONet discussed in the previous section. We also note that this is made possible through the internal parametrization of the basis functions used in the KAN architecture  that substantially improves the representation capacity of the KAN-based framework in the high frequency setting. However, an equivalent MLP-based model do not have this favorable characteristic and requires additional parameterizations through adaptive or Rowdy activation functions to overcome the spectral bias as shown in the previous sections. It is also worthwhile to note that, in scenarios where the segregation of the dataset based on their dominant frequency is not feasible prior to model training, it may not be possible to implement continual learning. In such cases one may have to devise alternate strategies to overcome spectral bias using KAN or resort to using MLP with adaptive activation functions to overcome the spectral bias as discussed in the previous sections.

\subsubsection{Inference}
\begin{figure}[H]
\centering
{
\centering
\includegraphics[width=0.3\linewidth]{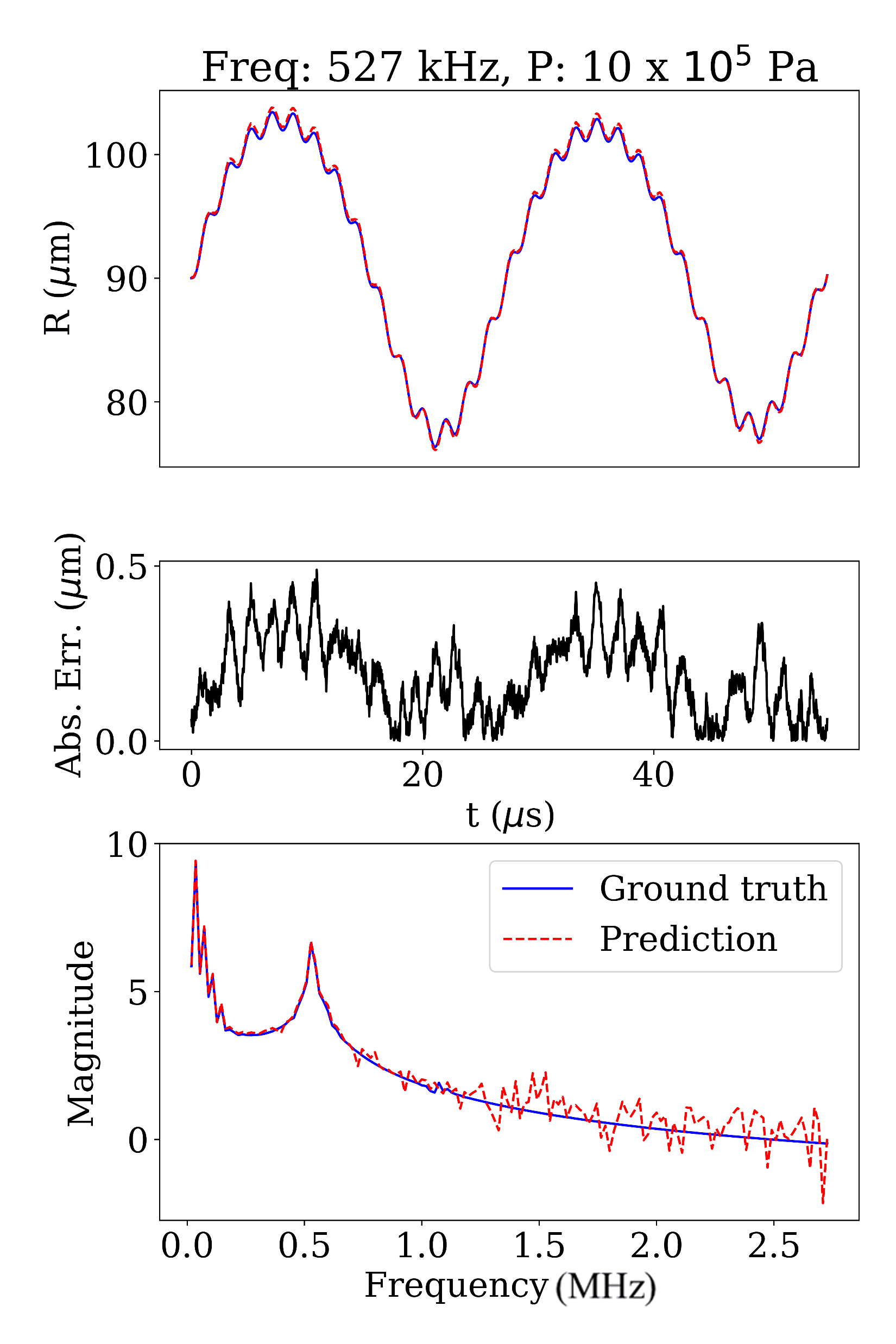}
}
\hspace{0.001cm}
\centering
{
\centering
\includegraphics[width=0.3\linewidth]{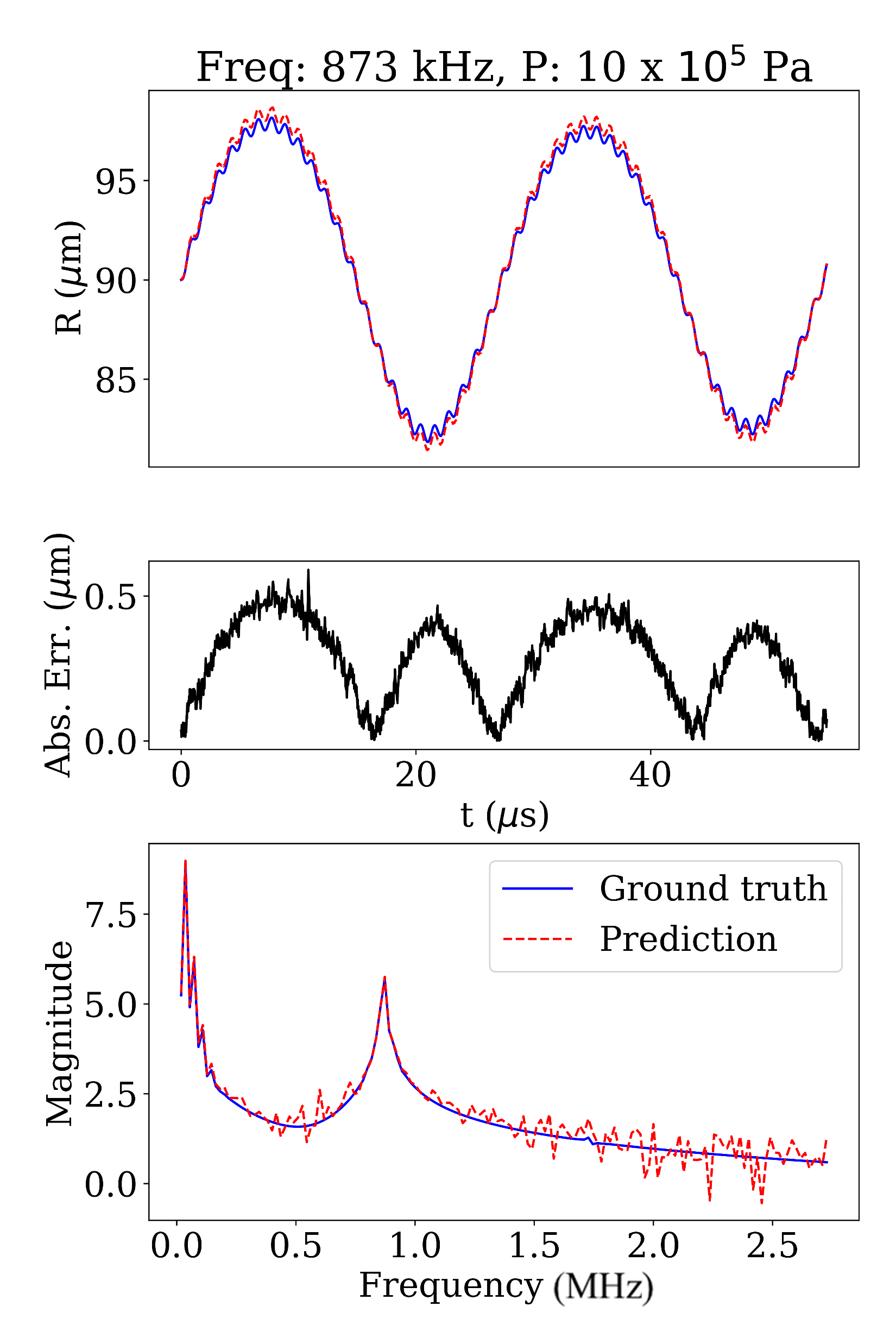}
}
\hspace{0.001cm}
\centering
{
\centering
\includegraphics[width=0.3\linewidth]{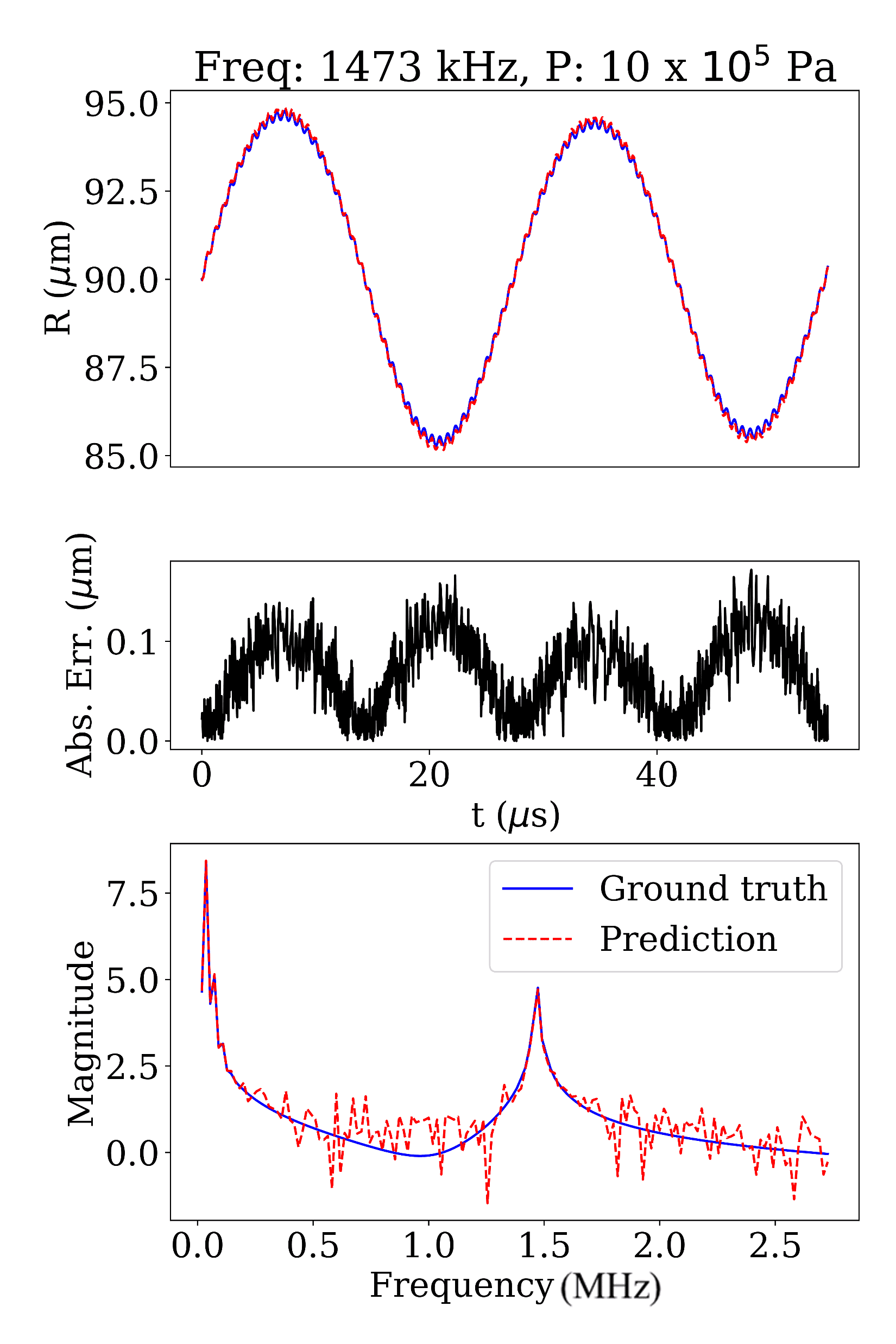}
}
\caption{Validation of two-step training method for bubbles simulated using K-M equation with: a frequency of (a) $527$ kHz, (b) $873$ kHz, and (c) $1473$ kHz for a fixed amplitude of $1 \times 10^6$ Pa.}
\label{fig:KAN_highpres}
\end{figure}

\begin{figure}[H]
\centering
{
\centering
\includegraphics[width=0.3\linewidth]{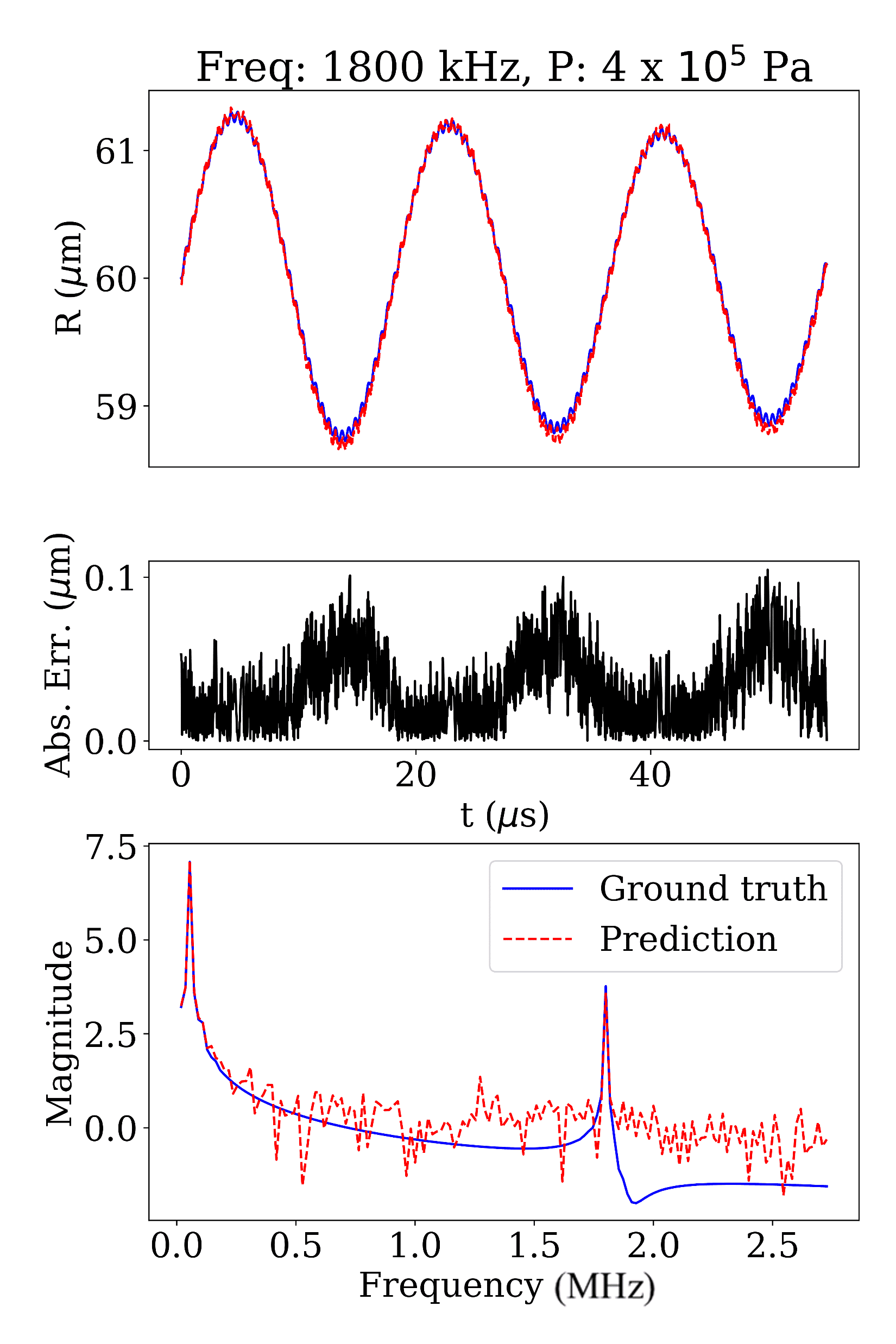}
}
\hspace{0.001cm}
\centering
{
\centering
\includegraphics[width=0.3\linewidth]{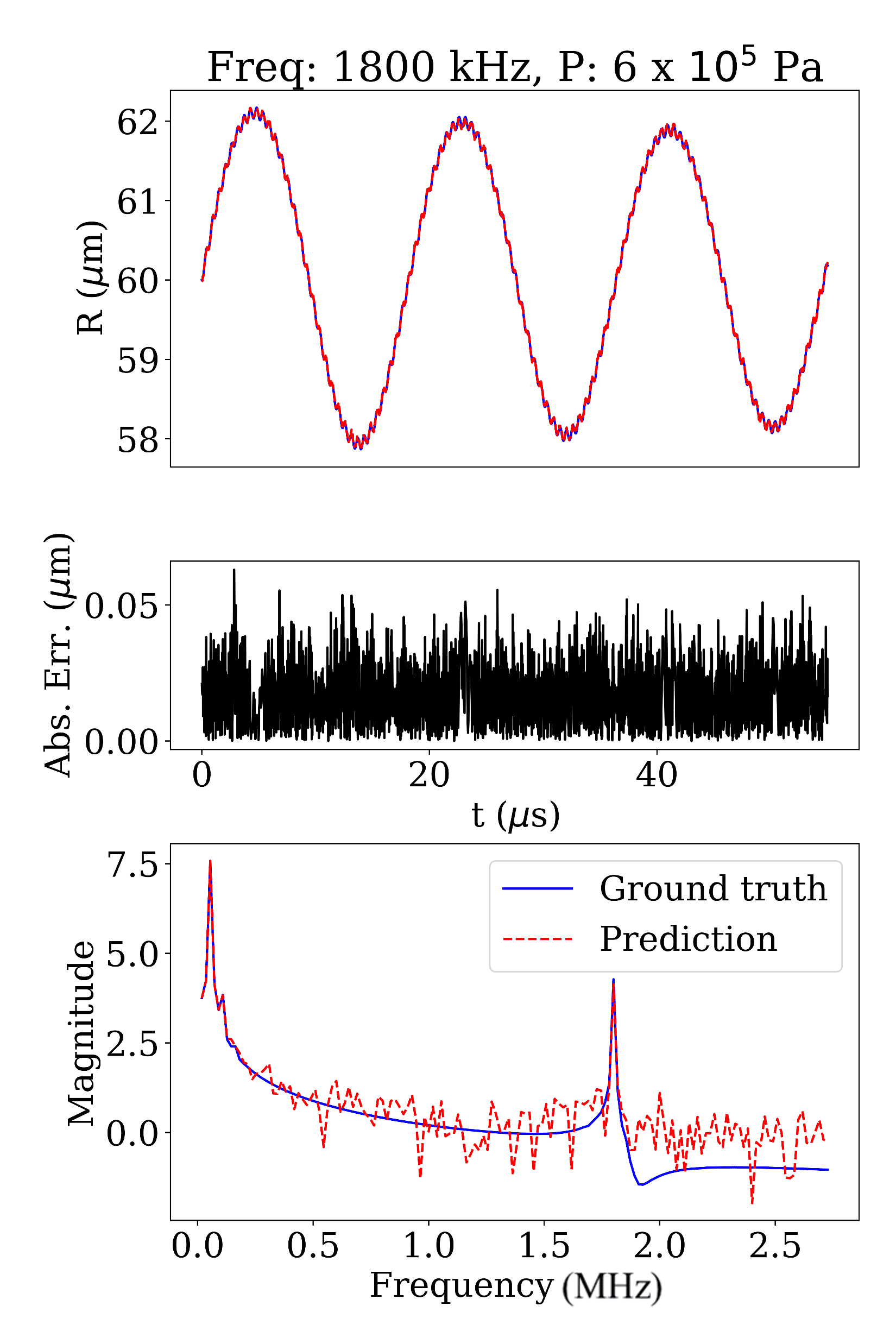}
}
\hspace{0.001cm}
\centering
{
\centering
\includegraphics[width=0.3\linewidth]{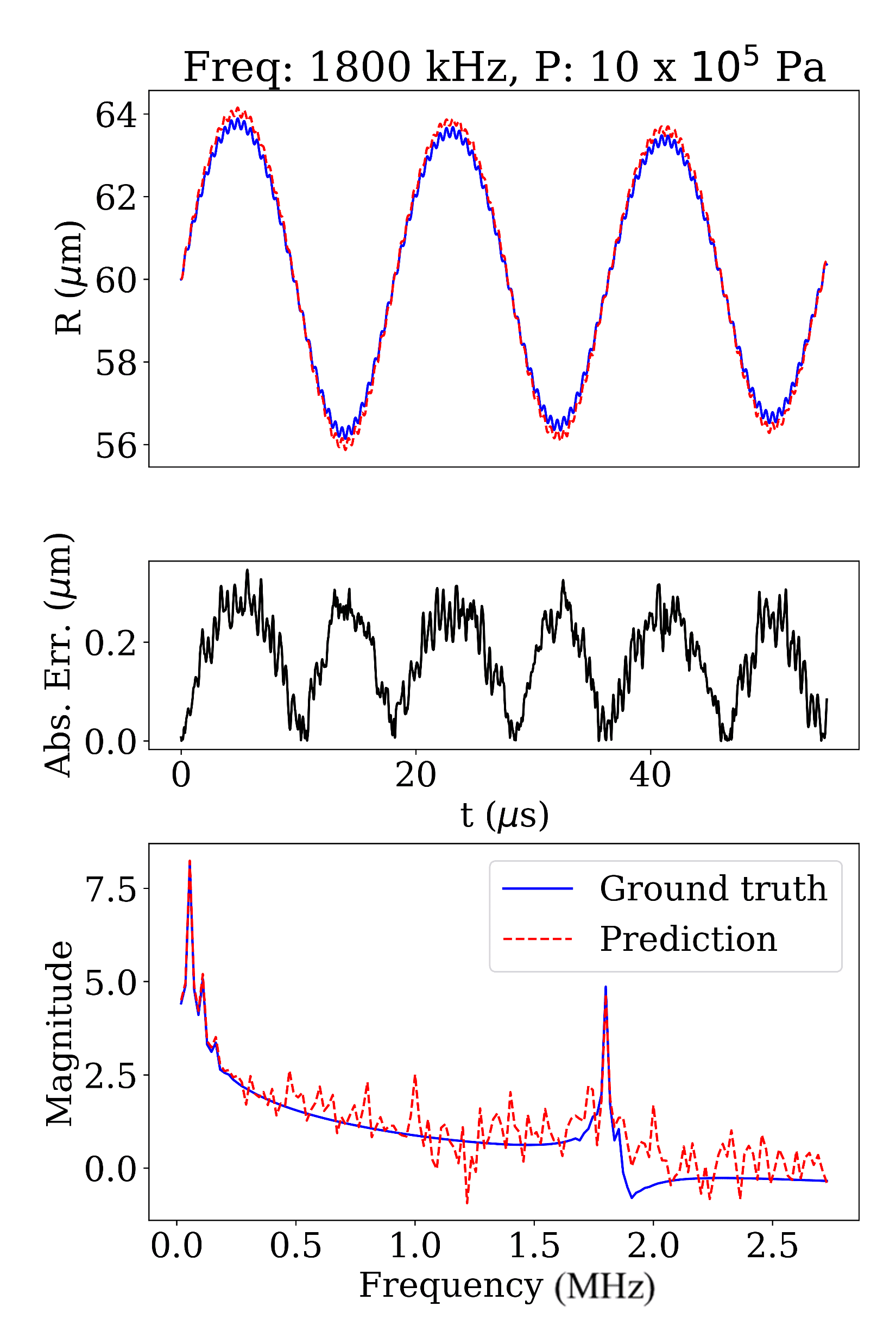}
}
\caption{Validation of two-step training method for bubbles simulated using K-M equation with: a frequency of $1800$ kHz and an amplitude of (a) $4 \times 10^5$ Pa, (b) $6 \times 10^5$ Pa, and (c) $1 \times 10^6$ Pa.}
\label{fig:KAN_highfreq}
\end{figure}

\begin{table}[H]
\caption{Hyperparameters for 2-step DeepOKAN}
\label{tab:hyperparameter_KAN}
\centering
\begin{tabular}{l|c|c|c}
\hline
Network & Learning rate & Basis & Architecture \\
\hline
Trunk & 0.0001 & Spline ($k = 2$, $G = 40$) & [2, 150, 150, 150, 150] \\
\hline
Branch & 0.0001 & RBF ($N = 6$) & [2000, 150, 150, 150, 150, 150, 150] \\
\hline
\end{tabular}
\end{table}
While the trunk network utilizes spline basis over the RBF \cite{abueidda2025deepokan} basis for reasons justified in the previous subsection, we utilize the RBF basis for the branch network mainly owing to its computational efficiency, leading to reduced training time compared to the spline basis. The architecture for trunk and branch networks is shown in Table \ref{tab:hyperparameter_KAN}. Figure \ref{fig:KAN_highpres} and Figure \ref{fig:KAN_highfreq} provide an overview of the two-step DeepOKAN model during inference. For a fixed amplitude of pressure, it can be observed that the model captures the localized dominant high frequency features along with the low frequency bubble dynamics for varying magnitudes of frequency. However, for extremely large frequencies such as $1800$ kHz, it can be observed that although the model predicts the dominant frequency with good accuracy, it poorly approximates the higher frequencies as shown in Figure \ref{fig:KAN_highfreq}. This may be improved further through continual learning of the trunk network where such extremely high frequencies may be treated as a separate batch ($1800 - 2000$ kHz) which can be learned in progression after the model learns a specific batch of frequencies with comparatively lower magnitudes ($<1800$ kHz).

\subsubsection{Comparison with vanilla DeepOKAN}

Here, we compare the results for the multi-radii case with the vanilla DeepOKAN \cite{abueidda2025deepokan} framework. While Liu et al. \cite{liu2024kan} proposed B-splines as the potential basis for the KAN architecture, Abueidda et al. \cite{abueidda2025deepokan} proposed DeepOKAN with the RBF basis owing to their superior approximation properties and computational efficiency. Here, we compare the multi-radii two-step DeepOKAN with the vanilla DeepOKAN complying with the layout shown in Figure \ref{fig:multiR_SKE} and the architecture shown in Table \ref{tab:hyperparameter_KAN} (with RBF basis) for fair comparison of both the architectures.

\begin{figure}[H]
\centering
{
\centering
\includegraphics[width=0.3\linewidth]{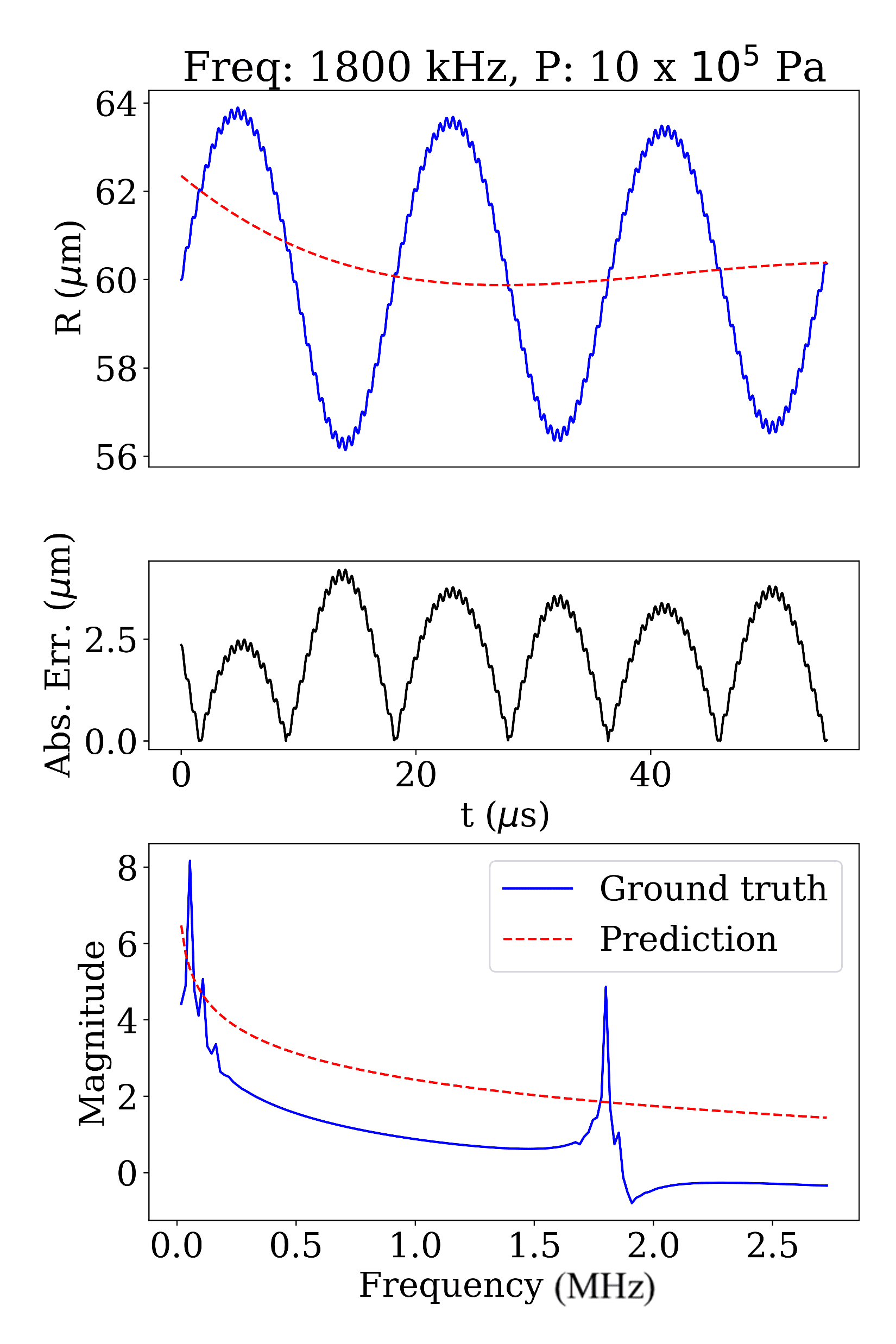}
}
\hspace{0.001cm}
\centering
{
\centering
\includegraphics[width=0.3\linewidth]{Figures/00_pres_1000000.0_Pa_freq_1800_kHz_error__18_1.png}
}
\caption{Comparison of (\textit{left}) DeepOKAN \cite{abueidda2025deepokan} and (\textit{right}) 2-step DeepOKAN for bubble dynamics simulated using K-M equation at a frequency of $1800$ kHz and an amplitude of $1 \times 10^6$ Pa.}
\label{fig:KAN_highfreq_comp}
\end{figure}

Figure \ref{fig:KAN_highfreq_comp} shows the comparison of both the architectures for a high frequency, high pressure test case. Here, we note a clear degradation of performance using the DeepOKAN \cite{abueidda2025deepokan} framework in the multi-radii setting and the utilization of the RBF basis doesn't favor the operator learning framework in the context of high frequency bubble dynamics. This observation also complies with the results in Figure \ref{fig:basis_lf} where it was shown that the performance of spline basis surpasses the RBF for the trunk network in the two-step DeepOKAN framework. To this end, we emphasize the need for careful selection of a suitable basis function based on the type of problem being solved as opposed to selecting a universal basis for the KAN architecture. Here, we also note the benefit of using a two-step operator learning framework which allows independently tuning the trunk and branch networks after taking into account factors like accuracy and computational efficiency of different basis functions as opposed to training them together as a single framework at least in the context of KAN. Furthermore, while the single step operator learning frameworks have showcased exceptional performance for various types of problems, it may be beneficial to transition to a two step operator learning framework based on the level of difficulty (such as overcoming the spectral bias shown in the current work). There are also exceptions to this. For example, FNO~\cite{li2020fourier} operates directly in the frequency domain and can effectively address issues related to spectral bias in that domain; however, it is difficult to implement for high-dimensional problems. The comparison of the proposed two-step operator frameworks with such state-of-the-art neural operators will be compared and discussed in the next section.

\subsection{Comparison with Different Neural Operators}
A fair benchmark study is conducted to compare several operator learning models, including Two-Step DeepONet, the Fourier Neural Operator (FNO)~\cite{li2020fourier}, the Wavelet Neural Operator (WNO)~\cite{tripura2023wavelet}, the Operator Transformer (OFormer)~\cite{li2022transformer}, the Convolutional Neural Operator (CNO)~\cite{raonic2023convolutional} and the proposed Two-Step DeepOKAN. Training is performed using bubble dynamics data corresponding to a single initial radius simulated by the Keller–Miksis equations. Validation is carried out across three different pressure amplitudes, \(1\times10^{5}\), \(6\times10^{5}\), and \(1\times10^{6}\,\text{Pa}\), and three driving frequencies of 600, 1300, and 2000\,kHz, representing low, medium, and high parameter regimes, respectively. 
\begin{sidewaystable}[htbp]
\centering
\caption{Comparison of relative $L_2$ errors (mean $\pm$ standard deviation over 10 realizations) for different neural operator models of bubble dynamics under varying frequencies and pressure amplitudes. The best results are highlighted in \textbf{bold}.}
\label{tab:model_comparison}
\setlength{\tabcolsep}{2.5pt}
{\footnotesize
\begin{tabular}{@{}cccccccc@{}}
\hline
\textbf{Amp (Pa)} & \textbf{Freq (KHz)} & \textbf{FNO} & \textbf{WNO} & \textbf{OFormer} & \textbf{CNO} & \textbf{2-Step DeepONet} & \textbf{2-step DeepOKAN} \\
\hline
\multirow{3}{*}{$1\times10^{5}$}
 & 600 & \(0.0479 \pm 0.0076\%\) & \(0.115 \pm 0.071\%\) & \(0.0948 \pm 0.0178\%\) & \(0.0418 \pm 0.0004\%\)  & \(0.0234 \pm 0.0156\%\) & \(\mathbf{0.0208 \pm 0.0070\%}\) \\ 
 & 1300 & \(0.0261 \pm 0.0088\%\) & \(0.083 \pm 0.180\%\) & \(0.0573 \pm 0.0297\%\) & \(0.0247 \pm 0.0024\%\) & \(\mathbf{0.0102 \pm 0.0028\%}\) & \(0.0169 \pm 0.0050\%\)\\ 
 & 2000 & \(0.0254 \pm 0.0069\%\) & \(0.091 \pm 0.086\%\) & \(0.0439 \pm 0.0157\%\) & \(0.0239 \pm 0.0032\%\) & \(0.0230 \pm 0.0276\%\) & \(\mathbf{0.0188 \pm 0.0055\%}\)\\ 
\hline
\multirow{3}{*}{$6\times10^{5}$}
 & 600 & \(0.0681 \pm 0.1605\%\) & \(0.158 \pm 0.058\%\) & \(0.0891 \pm 0.0157\%\) & \(0.0952 \pm 0.0271\%\) & \(\mathbf{0.0366 \pm 0.0183\%}\) & \(0.204 \pm 0.0460\%\)\\ 
 & 1300 & \(0.0732 \pm 0.0066\%\) & \(0.118 \pm 0.054\%\) & \(0.1189 \pm 0.0991\%\) & \(0.0390 \pm 0.0128\%\) & \(0.0788 \pm 0.0231\%\) & \(\mathbf{0.0209 \pm 0.0061\%}\)\\ 
 & 2000 & \(0.0567 \pm 0.0421\%\) & \(0.152 \pm 0.058\%\) & \(0.0954 \pm 0.0036\%\) & \(0.0301 \pm 0.0070\%\) & \(0.0546 \pm 0.0475\%\) & \(\mathbf{0.0226 \pm 0.0039\%}\)\\ 
\hline
\multirow{3}{*}{$1\times10^{6}$}
 & 600 & \(0.0858 \pm 0.2218\%\) & \(0.335 \pm 0.119\%\) & \(0.0950 \pm 0.0203\%\) & \(0.1916 \pm 0.0813\%\) & \(0.0485 \pm 0.0189\%\) & \(\mathbf{0.0305 \pm 0.0128\%}\)\\ 
 & 1300 & \(0.1141 \pm 0.0215\%\) & \(0.185 \pm 0.065\%\) & \(0.0601 \pm 0.0157\%\) & \(0.0560 \pm 0.0269\%\) & \(0.0320 \pm 0.0145\%\) & \(\mathbf{0.0271 \pm 0.0076\%}\)\\ 
 & 2000 & \(0.0898 \pm 0.0666\%\) & \(0.262 \pm 0.125\%\) & \(0.1079 \pm 0.0414\%\) & \(0.0437 \pm 0.0129\%\) & \(\mathbf{0.0379 \pm 0.0137\%}\) & \(0.0383 \pm 0.0068\%\)\\ 
\hline
\end{tabular}
}
\end{sidewaystable}
As shown in Table~\ref{tab:model_comparison}, all six models are evaluated over ten independent initializations, and their ensemble performance is compared using the distribution of \(L_{2}\)-norm errors (mean $\pm$ standard deviation). The two-step DeepOKAN demonstrates the highest accuracy and precision in most cases, followed by the two-step DeepONet. 
The results indicate that the two-step DeepOKAN maintains consistent performance in most cases within the specified frequency and pressure ranges, as summarized in Table~\ref{tab:model_comparison}.

\section{Conclusions}
\label{sec6}
High-frequency bubble dynamics plays an important role in many scientific and engineering domains, where rapid and accurate modeling is essential for understanding phase interactions and energy transfer in multiphase systems. While mathematical models such as the Rayleigh--Plesset and Keller--Miksis equations provide a well-characterized description of bubble dynamics, the lack of efficient prediction methods for multi-bubble simulations limits their applicability in real-time and in-situ scenarios. In this work, we employed a physics-informed two-step DeepONet model, which leverages the function approximation capabilities of DeepONet to model bubble dynamics. The framework accommodates bubbles governed by both the Rayleigh-Plesset and Keller-Miksis equations and handles scenarios involving single and multiple initial radii. In particular, we solved bubble dynamics over a large range of amplitudes (\( [1,10] \times 10^5 \) Pa) and frequencies (\( [200,2000] \times 10^6 \) KHz), representing a wide range of physically realistic scenarios. Our key findings show that the Rowdy activation function effectively mitigates spectral bias, enabling accurate prediction of bubble dynamics across varying pressure amplitudes and frequencies. The two-step training approach improves efficiency, achieving faster convergence while maintaining comparable accuracy in validation, interpolation, and extrapolation. The model is more sensitive to driving frequency than amplitude, capturing natural and driving frequencies when extrapolating over amplitude but facing challenges with frequency extrapolation. While these findings hold for the two-step DeepONet architecture, several important insights emerge for the proposed \textit{two-step DeepOKAN} framework:
\begin{enumerate}
    \item The proposed two-step DeepOKAN framework leverages the KAN architecture, providing parametric efficiency and interpretability absent in equivalent MLP-based models. However, the KAN architecture still exhibits spectral bias similar to MLPs.
    \item To address high-frequency spectral bias, we introduce a strategy that continually trains the trunk network, improving generalization across a range of frequencies and pressure amplitudes. This procedure can be extended to other problems without loss of generality using the KAN architecture.
    \item Although the RBF basis \cite{abueidda2025deepokan} is computationally efficient, it is limited in learning high-frequency phenomena such as bubble dynamics. We therefore employ a spline basis in the trunk network for improved training performance, while RBF is used in the branch network. In the two-step framework, the trunk network constructs the basis to represent high-frequency dynamics accurately, and the branch network weights this basis according to the input pressure distribution. The branch network’s quality is dependent on the trunk network, given the sequential training procedure.
\end{enumerate}

Despite its promising performance, the proposed two-step DeepOKAN model exhibits limitations. Importantly, it fails to capture bubble dynamics at resonance frequency, restricting its applicability in turbulent flows where a broad spectrum of frequencies (including resonance) is present. Furthermore, predictive capability is constrained to the time domain represented in the training data, leading to significant errors on out-of-distribution samples. Future research aims to enhance the model’s adaptability and generalization to unseen scenarios.

\section*{Acknowledgement}
We would like to thank all the reviewers for their valuable feedback and constructive suggestions, which have helped improve the quality and clarity of this work.

\appendix
\section{Non-dimensionalization of Rayleigh-Plesset Equation and Runge Kutta solution}
\label{Nondim R-P}
Starting from the original R-P equation, written as:
\begin{equation}
    \label{eq:fullR-P}
    R \ddot{R} + \frac{3}{2} \dot{R}^2 = \frac{1}{\rho} \left( P_B - P_{\infty} - 4\mu \frac{\dot{R}}{R}  - \frac{2S}{R} \right),
\end{equation}
where the internal bubble pressure is given by 
\[
P_B = P_G = P_{G0} \left( \frac{R_0}{R} \right)^{3k}.
\]

To nondimensionalize the R-P equation, we introduce the characteristic length \( a = R_0 \), time scale \( \tau = t_{\text{max}} \), and pressure scale \( P^* \). The corresponding non-dimensional variables are defined as:
\[
\bar{t} = \frac{t}{\tau}, \quad \bar{R} = \frac{R}{a}, \quad \bar{P} = \frac{P}{P^*}.
\]

Differentiating the dimensional radius with respect to time yields:
\begin{equation}
    \dot{R} = \frac{dR}{dt} = \frac{d(a\bar{R})}{d(\tau\bar{t})} = \frac{a}{\tau} \frac{d\bar{R}}{d\bar{t}} = \frac{a}{\tau} \dot{\bar{R}},
\end{equation}
and similarly,
\begin{equation}
    \ddot{R} = \frac{d^2R}{dt^2} = \frac{a}{\tau^2} \ddot{\bar{R}}.
\end{equation}

Substituting these expressions into Eq.~\eqref{eq:fullR-P} gives:
\begin{equation}
    \label{eq:nondimR-P_temp}
    \frac{a^2}{\tau^2} \left( \bar{R} \ddot{\bar{R}} + \frac{3}{2} \dot{\bar{R}}^2 \right) 
    = \frac{1}{\rho} \left( P_{G0} \left( \frac{1}{\bar{R}} \right)^{3k} - P^* \bar{P}_{\infty} - \frac{4\mu}{\tau} \frac{\dot{\bar{R}}}{\bar{R}} - \frac{2S}{a} \frac{1}{\bar{R}} \right).
\end{equation}

Dividing both sides by \( \frac{a^2}{\tau^2} \), we obtain the nondimensional Rayleigh–Plesset equation:
\begin{equation}
    \bar{R} \ddot{\bar{R}} + \frac{3}{2} \dot{\bar{R}}^2 
    = \frac{P_{G0} \tau^2}{\rho a^2} \left( \frac{1}{\bar{R}} \right)^{3k} 
    - \frac{P^* \tau^2}{\rho a^2} \bar{P}_{\infty}
    - \frac{4\mu \tau}{\rho a^2} \frac{\dot{\bar{R}}}{\bar{R}} 
    - \frac{2S \tau^2}{\rho a^3} \frac{1}{\bar{R}}.
\end{equation}

We define the pressure scale \( P^* \) as:
\[
P^* = n \frac{\rho a^2}{\tau^2},
\]
where \( n \) is a dimensionless scaling constant.

To simplify further, we introduce the Reynolds and Weber numbers as:
\begin{equation}
\begin{split}
    \text{Reynolds number:} \quad Re &= \frac{\rho a^2}{\mu \tau}, \\
    \text{Weber number:} \quad We &= \frac{\rho a^3}{S \tau^2}.
\end{split}
\end{equation}

Substituting these into the nondimensional equation, we obtain the final form:
\begin{equation}
    \label{eq:nondimR-P}
    \bar{R} \ddot{\bar{R}} + \frac{3}{2} \dot{\bar{R}}^2 
    = n\frac{P_{G0}}{P^*} \left( \frac{1}{\bar{R}} \right)^{3k} 
    - n \bar{P}_{\infty}
    - \frac{4}{Re} \frac{\dot{\bar{R}}}{\bar{R}} 
    - \frac{2}{We} \frac{1}{\bar{R}}.
\end{equation}

The Runge–Kutta solver for the non-dimensional R–P equation is written as:
\begin{equation}
    \begin{aligned}
        \bar{y} &= \begin{bmatrix} \bar{R} \\ \dot{\bar{R}} \end{bmatrix}, \\
        \dot{\bar{y}} &= \begin{bmatrix} \dot{\bar{R}} \\ \ddot{\bar{R}} \end{bmatrix} \\
        &= \begin{bmatrix} 
        \dot{\bar{R}} \\
        \frac{1}{\bar{R}} \left(
        n\frac{P_{G0}}{P^*} \left( \frac{1}{\bar{R}} \right)^{3k}
        - n \bar{P}_{\infty}
        - \frac{4\mu \tau}{\rho a^2} \frac{\dot{\bar{R}}}{\bar{R}}
        - \frac{2S \tau^2}{\rho a^3} \frac{1}{\bar{R}}
        - \frac{3}{2} \dot{\bar{R}}^2 \right)
        \end{bmatrix}.
    \end{aligned}
\end{equation}

\section{Non-dimensionalization of Keller-Miksis Equation and Runge Kutta solution}
\label{Nondim K-M}
Staring from Eq.~\ref{eq:K-M}:
\begin{equation}
\label{eq:fullK-M}
    \left(1 - \frac{\dot{R}}{c} \right) R \ddot{R} 
    + \frac{3}{2} \left(1 - \frac{\dot{R}}{3c} \right) \dot{R}^2 
    = \left(1 + \frac{\dot{R}}{c} \right) \frac{P_L - P_\infty}{\rho} 
    + R \frac{\dot{P}_L - \dot{P}_\infty}{\rho c},
\end{equation}
where the liquid pressure at the bubble wall is defined as
\[
P_L = P_G - 4\mu \frac{\dot{R}}{R} - \frac{2S}{R},
\]
and its time derivative is
\[
\dot{P}_L = \dot{P}_G + 4\mu \left( \frac{\dot{R}^2}{R^2} - \frac{\ddot{R}}{R} \right),
\]
with the gas pressure inside the bubble modeled as
\[
P_G = P_{G0} \left( \frac{R_0}{R} \right)^{3k}.
\]

To nondimensionalize the equation, we introduce characteristic scales:
\[
a = R_0, \quad \tau = t_{\text{max}}, \quad P^* = n \frac{\rho a^2}{\tau^2},
\]
and define the nondimensional variables:
\[
\bar{t} = \frac{t}{\tau}, \quad \bar{R} = \frac{R}{a}, \quad \bar{P} = \frac{P}{P^*}.
\]

Differentiating the radius yields
\[
\dot{R} = \frac{a}{\tau} \dot{\bar{R}}, \quad \ddot{R} = \frac{a}{\tau^2} \ddot{\bar{R}}.
\]

Substituting these expressions into Eq.~\eqref{eq:fullK-M}, we transform each term:

\begin{align}
    \left(1 - \frac{\dot{R}}{c} \right) R \ddot{R} 
    &= \frac{a^2}{\tau^2} \left(1 - \frac{a}{\tau c} \dot{\bar{R}} \right) \bar{R} \ddot{\bar{R}}, \\
    \frac{3}{2} \left(1 - \frac{\dot{R}}{3c} \right) \dot{R}^2 
    &= \frac{3}{2} \left(1 - \frac{a}{3\tau c} \dot{\bar{R}} \right) \frac{a^2}{\tau^2} \dot{\bar{R}}^2, \\
    \left(1 + \frac{\dot{R}}{c} \right) \frac{P_L - P_\infty}{\rho} 
    &= \frac{1}{\rho} \left(1 + \frac{a}{\tau c} \dot{\bar{R}} \right) 
    \left( P_{G0} \bar{R}^{-3k} - \frac{2S}{a} \frac{1}{\bar{R}} 
    - \frac{4\mu}{\tau} \frac{\dot{\bar{R}}}{\bar{R}} - P^* \bar{P} \right), \\
    R \frac{\dot{P}_L - \dot{P}_\infty}{\rho c} 
    &= \frac{a}{\rho c} \bigg[ 
        \frac{P_{G0}}{\tau} (-3k) \bar{R}^{-3k} \dot{\bar{R}} 
        + \frac{2S}{\tau a} \frac{\dot{\bar{R}}}{\bar{R}} 
        + \frac{4\mu}{\tau^2} \frac{\dot{\bar{R}}^2}{\bar{R}} 
        - \frac{4\mu}{\tau^2} \ddot{\bar{R}} 
        - \frac{P^*}{\tau} \bar{R} \dot{\bar{P}} 
    \bigg].
\end{align}

Next, we define the following dimensionless number:
\begin{equation}
\label{eq:dimensionless_groups}
\begin{split}
    \text{Mach number:} \quad &M = \frac{a}{\tau c}, \\
    \text{Reynolds number:} \quad &Re = \frac{\rho a^2}{\mu \tau}, \\
    \text{Weber number:} \quad &We = \frac{\rho a^3}{S \tau^2}.
\end{split}
\end{equation}

Dividing the entire equation by \( \frac{a^2}{\tau^2} \) yields the nondimensional form:
\begin{align}
    &\left(1 - M \dot{\bar{R}} \right) \bar{R} \ddot{\bar{R}} 
    + \frac{3}{2} \left(1 - \frac{M}{3} \dot{\bar{R}} \right) \dot{\bar{R}}^2 \nonumber \\
    &= (1 + M \dot{\bar{R}}) \bigg[ 
        \frac{P_{G0} \tau^2}{\rho a^2} \left( \frac{1}{\bar{R}} \right)^{3k}
        - \frac{2}{We} \frac{1}{\bar{R}}
        - \frac{4}{Re} \frac{\dot{\bar{R}}}{\bar{R}} 
        - \frac{P^* \tau^2}{\rho a^2} \bar{P} 
    \bigg] \nonumber \\
    &\quad + M \bigg[
        \frac{P_{G0} \tau^2}{\rho a^2} (-3k) \left( \frac{1}{\bar{R}} \right)^{3k} \dot{\bar{R}} 
        + \frac{2}{We} \frac{\dot{\bar{R}}}{\bar{R}} 
        + \frac{4}{Re} \frac{\dot{\bar{R}}^2}{\bar{R}} 
        - \frac{4}{Re} \ddot{\bar{R}} 
        - \frac{P^* \tau^2}{\rho a^2} \bar{R} \dot{\bar{P}} 
    \bigg].
\end{align}

Finally, with \( n \) as a dimensionless scaling constant, we define 
\[
 P^* = n \frac{\rho a^2}{\tau^2}
\]
we obtain the fully nondimensionalized Keller–Miksis equation:
\begin{align}
    &\left(1 - M \dot{\bar{R}} \right) \bar{R} \ddot{\bar{R}} 
    + \frac{3}{2} \left(1 - \frac{M}{3} \dot{\bar{R}} \right) \dot{\bar{R}}^2 \nonumber \\
    &= (1 + M \dot{\bar{R}}) \bigg[ 
        n\frac{P_{G0}}{P^*} \left( \frac{1}{\bar{R}} \right)^{3k}
        - \frac{2}{We} \frac{1}{\bar{R}}
        - \frac{4}{Re} \frac{\dot{\bar{R}}}{\bar{R}} 
        - n \bar{P} 
    \bigg] \nonumber \\
    &\quad + M \bigg[
        n\frac{P_{G0}}{P^*} (-3k) \left( \frac{1}{\bar{R}} \right)^{3k} \dot{\bar{R}} 
        + \frac{2}{We} \frac{\dot{\bar{R}}}{\bar{R}} 
        + \frac{4}{Re} \frac{\dot{\bar{R}}^2}{\bar{R}} 
        - \frac{4}{Re} \ddot{\bar{R}} 
        - n \bar{R} \dot{\bar{P}} 
    \bigg].
\end{align}

The resulting system for a Runge–Kutta (RK) solver is expressed in first-order form as:
\begin{equation}
\label{eq:state_form}
\begin{aligned}
    \bar{y} &= \begin{bmatrix} \bar{R} \\ \dot{\bar{R}} \end{bmatrix}, \\
    \dot{\bar{y}} &= \begin{bmatrix} 
        \dot{\bar{R}} \\
        \displaystyle\frac{1}{\left(1 - M \dot{\bar{R}} \right) \bar{R} + \frac{4M}{Re}} \Bigg(
        \begin{aligned}[t]
            & (1 + M \dot{\bar{R}}) \bigg[ 
                n\frac{P_{G0}}{P^*} \left( \frac{1}{\bar{R}} \right)^{3k}
                - \frac{2}{We} \frac{1}{\bar{R}} 
                - \frac{4}{Re} \frac{\dot{\bar{R}}}{\bar{R}} 
                - n \bar{P} \bigg] \\
            & + M \bigg[
                -3k\,n\frac{P_{G0}}{P^*} \left( \frac{1}{\bar{R}} \right)^{3k} \dot{\bar{R}} 
                + \frac{2}{We} \frac{\dot{\bar{R}}}{\bar{R}} 
                + \frac{4}{Re} \frac{\dot{\bar{R}}^2}{\bar{R}} 
                - n \bar{R} \dot{\bar{P}} \bigg] \\
            & - \frac{3}{2} \left(1 - \frac{M}{3} \dot{\bar{R}} \right) \dot{\bar{R}}^2 \Bigg)
        \end{aligned}
    \end{bmatrix}.
\end{aligned}
\end{equation}

\bibliographystyle{elsarticle-num} 
\bibliography{reference}

@article{Lu2021,
author = {Lu, Lu and Jin, Pengzhan and Pang, Guofei and Zhang, Zhongqiang and Karniadakis, George Em},
doi = {10.1038/s42256-021-00302-5},
issn = {25225839},
journal = {Nature Machine Intelligence},
number = {3},
pages = {218--229},
title = {{Learning nonlinear operators via DeepONet based on the universal approximation theorem of operators}},
volume = {3},
year = {2021}
}

@article{Mao2021,
archivePrefix = {arXiv},
arxivId = {2011.03349},
author = {Mao, Zhiping and Lu, Lu and Marxen, Olaf and Zaki, Tamer A. and Karniadakis, George Em},
doi = {10.1016/j.jcp.2021.110698},
eprint = {2011.03349},
issn = {10902716},
journal = {Journal of Computational Physics},
pages = {1--24},
title = {{DeepM\&Mnet for hypersonics: Predicting the coupled flow and finite-rate chemistry behind a normal shock using neural-network approximation of operators}},
volume = {447},
year = {2021}
}

@article{Jin2022,
author = {Jin, Pengzhan and Meng, Shuai and Lu, Lu},
journal = {SIAM Journal on Scientific Computing},
number = {6},
pages = {A3490--A3514},
title = {{MIONet: Learning multiple-input operators via tensor product}},
volume = {44},
year = {2022}
}

@article{hu2024tackling,
  title={Tackling the curse of dimensionality with physics-informed neural networks},
  author={Hu, Zheyuan and Shukla, Khemraj and Karniadakis, George Em and Kawaguchi, Kenji},
  journal={Neural Networks},
  volume={176},
  pages={106369},
  year={2024},
  publisher={Elsevier}
}

@article{menon2025anant,
  title={Anant-net: Breaking the curse of dimensionality with scalable and interpretable neural surrogate for high-dimensional pdes},
  author={Menon, Sidharth S and Jagtap, Ameya D},
  journal={arXiv preprint arXiv:2505.03595},
  year={2025}
}

@article{Lin2021_1,
author = {Lin, Chensen and Maxey, Martin and Li, Zhen and Karniadakis, George Em},
doi = {10.1017/jfm.2021.866},
issn = {14697645},
journal = {Journal of Fluid Mechanics},
pages = {1--14},
title = {{A seamless multiscale operator neural network for inferring bubble dynamics}},
volume = {929},
year = {2021}
}

@article{Lin2021_2,
archivePrefix = {arXiv},
arxivId = {2012.12816},
author = {Lin, Chensen and Li, Zhen and Lu, Lu and Cai, Shengze and Maxey, Martin and Karniadakis, George Em},
doi = {10.1063/5.0041203},
eprint = {2012.12816},
issn = {10897690},
journal = {Journal of Chemical Physics},
number = {10},
pmid = {33722055},
title = {{Operator learning for predicting multiscale bubble growth dynamics}},
volume = {154},
year = {2021}
}

@article{Cao2021,
archivePrefix = {arXiv},
arxivId = {1912.01198},
author = {Cao, Yuan and Fang, Zhiying and Wu, Yue and Zhou, Ding Xuan and Gu, Quanquan},
doi = {10.24963/ijcai.2021/304},
eprint = {1912.01198},
isbn = {9780999241196},
issn = {10450823},
journal = {IJCAI International Joint Conference on Artificial Intelligence},
pages = {2205--2211},
title = {{Towards Understanding the Spectral Bias of Deep Learning}},
year = {2021}
}

@article{Rahaman2019,
archivePrefix = {arXiv},
arxivId = {1806.08734},
author = {Rahaman, Nasim and Baratin, Aristide and Arpit, Devansh and Draxlcr, Felix and Lin, Min and Hamprecht, Fred A. and Bengio, Yoshua and Courville, Aaron},
eprint = {1806.08734},
isbn = {9781510886988},
journal = {36th International Conference on Machine Learning, ICML 2019},
number = {1},
pages = {9230--9239},
title = {{On the spectral bias of neural networks}},
volume = {2019-June},
year = {2019}
}

@article{Jagtap2022,
archivePrefix = {arXiv},
arxivId = {2105.09513},
author = {Jagtap, Ameya D. and Shin, Yeonjong and Kawaguchi, Kenji and Karniadakis, George Em},
doi = {10.1016/j.neucom.2021.10.036},
eprint = {2105.09513},
issn = {18728286},
journal = {Neurocomputing},
pages = {165--180},
publisher = {Elsevier B.V.},
title = {{Deep Kronecker neural networks: A general framework for neural networks with adaptive activation functions}},
url = {https://doi.org/10.1016/j.neucom.2021.10.036},
volume = {468},
year = {2022}
}

@article{Gulcehre2016,
archivePrefix = {arXiv},
arxivId = {1603.00391},
author = {Gulcehre, Caglar and Moczulski, Marcin and Denil, Misha and Bengio, Yoshua},
eprint = {1603.00391},
isbn = {9781510829008},
journal = {33rd International Conference on Machine Learning, ICML 2016},
pages = {4457--4466},
title = {{Noisy activation functions}},
volume = {6},
year = {2016}
}

@article{Shridhar2019,
archivePrefix = {arXiv},
arxivId = {1905.10761},
author = {Shridhar, Kumar and Lee, Joonho and Hayashi, Hideaki and Mehta, Purvanshi and Iwana, Brian Kenji and Kang, Seokjun and Uchida, Seiichi and Ahmed, Sheraz and Dengel, Andreas},
eprint = {1905.10761},
journal = {arXiv preprint arXiv:1905.10761},
title = {{ProbAct: A Probabilistic Activation Function for Deep Neural Networks}},
url = {http://arxiv.org/abs/1905.10761},
year = {2019}
}

@article{Jagtap2020,
archivePrefix = {arXiv},
arxivId = {1906.01170},
author = {Jagtap, Ameya D. and Kawaguchi, Kenji and Karniadakis, George Em},
doi = {10.1016/j.jcp.2019.109136},
eprint = {1906.01170},
issn = {10902716},
journal = {Journal of Computational Physics},
pages = {109136},
publisher = {Elsevier Inc.},
title = {{Adaptive activation functions accelerate convergence in deep and physics-informed neural networks}},
url = {https://doi.org/10.1016/j.jcp.2019.109136},
volume = {404},
year = {2020}
}

@article{Karniadakis2021,
author = {Karniadakis, George Em and Kevrekidis, Ioannis G. and Lu, Lu and Perdikaris, Paris and Wang, Sifan and Yang, Liu},
doi = {10.1038/s42254-021-00314-5},
isbn = {0123456789},
issn = {25225820},
journal = {Nature Reviews Physics},
number = {6},
pages = {422--440},
title = {{Physics-informed machine learning}},
volume = {3},
year = {2021}
}

@article{Raissi2019,
author = {Raissi, M. and Perdikaris, P. and Karniadakis, G. E.},
doi = {10.1016/j.jcp.2018.10.045},
issn = {10902716},
journal = {Journal of Computational Physics},
pages = {686--707},
publisher = {Elsevier Inc.},
title = {{Physics-informed neural networks: A deep learning framework for solving forward and inverse problems involving nonlinear partial differential equations}},
url = {https://doi.org/10.1016/j.jcp.2018.10.045},
volume = {378},
year = {2019}
}

@article{Zhu2019,
archivePrefix = {arXiv},
arxivId = {1901.06314},
author = {Zhu, Yinhao and Zabaras, Nicholas and Koutsourelakis, Phaedon Stelios and Perdikaris, Paris},
doi = {10.1016/j.jcp.2019.05.024},
eprint = {1901.06314},
issn = {10902716},
journal = {Journal of Computational Physics},
pages = {56--81},
publisher = {Elsevier Inc.},
title = {{Physics-constrained deep learning for high-dimensional surrogate modeling and uncertainty quantification without labeled data}},
url = {https://doi.org/10.1016/j.jcp.2019.05.024},
volume = {394},
year = {2019}
}

@article{Cheng2022,
author = {Cheng, Lin and Wagner, Gregory J.},
doi = {10.1016/j.cma.2021.114507},
issn = {00457825},
journal = {Computer Methods in Applied Mechanics and Engineering},
month = {feb},
publisher = {Elsevier B.V.},
title = {{A representative volume element network (RVE-net) for accelerating RVE analysis, microscale material identification, and defect characterization}},
volume = {390},
year = {2022}
}

@article{lee2024training,
  title={On the training and generalization of deep operator networks},
  author={Lee, Sanghyun and Shin, Yeonjong},
  journal={SIAM Journal on Scientific Computing},
  volume={46},
  number={4},
  pages={C273--C296},
  year={2024},
  publisher={SIAM}
}

@article{wang2021learning,
  title={Learning the solution operator of parametric partial differential equations with physics-informed DeepONets},
  author={Wang, Sifan and Wang, Hanwen and Perdikaris, Paris},
  journal={Science advances},
  volume={7},
  number={40},
  pages={eabi8605},
  year={2021},
  publisher={American Association for the Advancement of Science}
}

@article{Mao2020,
author = {Mao, Zhiping and Jagtap, Ameya D. and Karniadakis, George Em},
doi = {10.1016/j.cma.2019.112789},
issn = {00457825},
journal = {Computer Methods in Applied Mechanics and Engineering},
pages = {112789},
publisher = {Elsevier B.V.},
title = {{Physics-informed neural networks for high-speed flows}},
url = {https://doi.org/10.1016/j.cma.2019.112789},
volume = {360},
year = {2020}
}

@article{Zhu2021,
archivePrefix = {arXiv},
arxivId = {2008.13547},
author = {Zhu, Qiming and Liu, Zeliang and Yan, Jinhui},
doi = {10.1007/s00466-020-01952-9},
eprint = {2008.13547},
issn = {14320924},
journal = {Computational Mechanics},
month = {feb},
number = {2},
pages = {619--635},
publisher = {Springer Science and Business Media Deutschland GmbH},
title = {{Machine learning for metal additive manufacturing: predicting temperature and melt pool fluid dynamics using physics-informed neural networks}},
volume = {67},
year = {2021}
}

@book{brennen2014cavitation,
  title={Cavitation and bubble dynamics},
  author={Brennen, Christopher E},
  year={2014},
  publisher={Cambridge university press}
}

@incollection{franc2007rayleigh,
  title={The Rayleigh-Plesset equation: a simple and powerful tool to understand various aspects of cavitation},
  author={Franc, Jean-Pierre},
  booktitle={Fluid dynamics of cavitation and cavitating turbopumps},
  pages={1--41},
  year={2007},
  publisher={Springer}
}

@article{Hong2024,
author = {Hong, Yi and Li, Miaomiao and He, Xiaodong and Xing, Jing Tang},
doi = {10.1016/j.oceaneng.2024.118072},
issn = {00298018},
journal = {Ocean Engineering},
number = {April},
pages = {118072},
publisher = {Elsevier Ltd},
title = {{Energy flow investigations of Rayleigh-Plesset equation for cavitation simulations}},
url = {https://doi.org/10.1016/j.oceaneng.2024.118072},
volume = {306},
year = {2024}
}

@article{Denner2023, doi = {10.21105/joss.05435}, url = {https://doi.org/10.21105/joss.05435}, year = {2023}, publisher = {The Open Journal}, volume = {8}, number = {86}, pages = {5435}, author = {Fabian Denner and Sören Schenke}, title = {APECSS: A software library for cavitation bubble dynamics and acoustic emissions}, journal = {Journal of Open Source Software} }

@article{Dormand1986,
author = {Dormand, J. R. and Prince, P. J.},
doi = {10.1016/0377-0427(86)90027-0},
issn = {03770427},
journal = {Journal of Computational and Applied Mathematics},
number = {2},
pages = {203--211},
title = {{A family of embedded Runge-Kutta formulae}},
volume = {15},
year = {1986}
}

@article{keller1980bubble,
  title={Bubble oscillations of large amplitude},
  author={Keller, Joseph B and Miksis, Michael},
  journal={The Journal of the Acoustical Society of America},
  volume={68},
  number={2},
  pages={628--633},
  year={1980},
  publisher={Acoustical Society of America}
}

@article{Peyvan2024,
archivePrefix = {arXiv},
arxivId = {2401.08886},
author = {Peyvan, Ahmad and Oommen, Vivek and Jagtap, Ameya D. and Karniadakis, George Em},
doi = {10.1016/j.cma.2024.116996},
eprint = {2401.08886},
issn = {00457825},
journal = {Computer Methods in Applied Mechanics and Engineering},
number = {April},
pages = {116996},
publisher = {Elsevier B.V.},
title = {{RiemannONets: Interpretable neural operators for Riemann problems}},
url = {https://doi.org/10.1016/j.cma.2024.116996},
volume = {426},
year = {2024}
}

@article{osorio2022forecasting,
  title={Forecasting solar-thermal systems performance under transient operation using a data-driven machine learning approach based on the deep operator network architecture},
  author={Osorio, Julian D and Wang, Zhicheng and Karniadakis, George and Cai, Shengze and Chryssostomidis, Chrys and Panwar, Mayank and Hovsapian, Rob},
  journal={Energy Conversion and Management},
  volume={252},
  pages={115063},
  year={2022},
  publisher={Elsevier}
}

@article{pua2009ultrasound,
  title={Ultrasound-mediated drug delivery},
  author={Pua, Eric C and Zhong, Pei},
  journal={IEEE engineering in medicine and biology magazine},
  volume={28},
  number={1},
  pages={64--75},
  year={2009},
  publisher={IEEE}
}

@article{cako2022cavitation,
  title={Cavitation based cleaner technologies for biodiesel production and processing of hydrocarbon streams: A perspective on key fundamentals, missing process data and economic feasibility--A review},
  author={Cako, Elvana and Wang, Zhaohui and Castro-Mu{\~n}oz, Roberto and Rayaroth, Manoj P and Boczkaj, Grzegorz},
  journal={Ultrasonics sonochemistry},
  volume={88},
  pages={106081},
  year={2022},
  publisher={Elsevier}
}

@article{rayleigh1917viii,
  title={VIII. On the pressure developed in a liquid during the collapse of a spherical cavity},
  author={Rayleigh, Lord},
  journal={The London, Edinburgh, and Dublin Philosophical Magazine and Journal of Science},
  volume={34},
  number={200},
  pages={94--98},
  year={1917},
  publisher={Taylor \& Francis}
}

@article{Lee2024,
archivePrefix = {arXiv},
arxivId = {2309.01020},
author = {Lee, Sanghyun and Shin, Yeonjong},
doi = {10.1137/23M1598751},
eprint = {2309.01020},
issn = {10957197},
journal = {SIAM Journal on Scientific Computing},
number = {4},
pages = {C273--C296},
title = {{on the Training and Generalization of Deep Operator Networks}},
volume = {46},
year = {2024}
}

@article{gnanaskandan2019modeling,
  title={Modeling of microbubble-enhanced high-intensity focused ultrasound},
  author={Gnanaskandan, Aswin and Hsiao, Chao-Tsung and Chahine, Georges},
  journal={Ultrasound in medicine \& biology},
  volume={45},
  number={7},
  pages={1743--1761},
  year={2019},
  publisher={Elsevier}
}

@article{gnanaskandan2016large,
  title={Large eddy simulation of the transition from sheet to cloud cavitation over a wedge},
  author={Gnanaskandan, Aswin and Mahesh, Krishnan},
  journal={International Journal of Multiphase Flow},
  volume={83},
  pages={86--102},
  year={2016},
  publisher={Elsevier}
}

@book{gilmore1952growth,
  title={The growth or collapse of a spherical bubble in a viscous compressible liquid},
  author={Gilmore, Forrest Richard},
  volume={26},
  year={1952},
  publisher={California Institute of Technology Pasadena, CA}
}

@article{prosperetti1986bubble,
  title={Bubble dynamics in a compressible liquid. Part 1. First-order theory},
  author={Prosperetti, A and Lezzi, A},
  journal={Journal of Fluid Mechanics},
  volume={168},
  pages={457--478},
  year={1986},
  publisher={Cambridge University Press}
}

@inproceedings{zhang2025bubble,
  title={Physics informed operator learning for predicting bubble dynamics},
  author={Zhang, Yunhao and Cheng, Lin and Gnanaskandan, Aswin},
  booktitle={Fluids Engineering Division Summer Meeting},
  volume={88995},
  pages={V001T01A004},
  year={2025},
  organization={American Society of Mechanical Engineers}
}

@article{jagtap2022physics,
  title={Physics-informed neural networks for inverse problems in supersonic flows},
  author={Jagtap, Ameya D and Mao, Zhiping and Adams, Nikolaus and Karniadakis, George Em},
  journal={Journal of Computational Physics},
  volume={466},
  pages={111402},
  year={2022},
  publisher={Elsevier}
}

@article{jagtap2020conservative,
  title={Conservative physics-informed neural networks on discrete domains for conservation laws: Applications to forward and inverse problems},
  author={Jagtap, Ameya D and Kharazmi, Ehsan and Karniadakis, George Em},
  journal={Computer Methods in Applied Mechanics and Engineering},
  volume={365},
  pages={113028},
  year={2020},
  publisher={Elsevier}
}

@article{jagtap2022deep,
  title={Deep learning of inverse water waves problems using multi-fidelity data: Application to Serre--Green--Naghdi equations},
  author={Jagtap, Ameya D and Mitsotakis, Dimitrios and Karniadakis, George Em},
  journal={Ocean Engineering},
  volume={248},
  pages={110775},
  year={2022},
  publisher={Elsevier}
}

@article{hu2023augmented,
  title={Augmented Physics-Informed Neural Networks (APINNs): A gating network-based soft domain decomposition methodology},
  author={Hu, Zheyuan and Jagtap, Ameya D and Karniadakis, George Em and Kawaguchi, Kenji},
  journal={Engineering Applications of Artificial Intelligence},
  volume={126},
  pages={107183},
  year={2023},
  publisher={Elsevier}
}

@article{penwarden2023unified,
  title={A unified scalable framework for causal sweeping strategies for physics-informed neural networks (PINNs) and their temporal decompositions},
  author={Penwarden, Michael and Jagtap, Ameya D and Zhe, Shandian and Karniadakis, George Em and Kirby, Robert M},
  journal={Journal of Computational Physics},
  volume={493},
  pages={112464},
  year={2023},
  publisher={Elsevier}
}

@article{shukla2021physics,
  title={A physics-informed neural network for quantifying the microstructural properties of polycrystalline nickel using ultrasound data: A promising approach for solving inverse problems},
  author={Shukla, Khemraj and Jagtap, Ameya D and Blackshire, James L and Sparkman, Daniel and Karniadakis, George Em},
  journal={IEEE Signal Processing Magazine},
  volume={39},
  number={1},
  pages={68--77},
  year={2021},
  publisher={IEEE}
}

@article{abbasi2025history,
  title={History-matching of imbibition flow in fractured porous media using physics-informed neural networks (pinns)},
  author={Abbasi, Jassem and Moseley, Ben and Kurotori, Takeshi and Jagtap, Ameya D and Kovscek, Anthony R and Hiorth, Aksel and Andersen, P{\aa}l {\O}steb{\o}},
  journal={Computer Methods in Applied Mechanics and Engineering},
  volume={437},
  pages={117784},
  year={2025},
  publisher={Elsevier}
}

@article{abbasi2025challenges,
  title={Challenges and advancements in modeling shock fronts with physics-informed neural networks: A review and benchmarking study},
  author={Abbasi, Jassem and Jagtap, Ameya D and Moseley, Ben and Hiorth, Aksel and Andersen, P{\aa}l {\O}steb{\o}},
  journal={arXiv preprint arXiv:2503.17379},
  year={2025}
}

@article{jagtap2020extended,
  title={Extended physics-informed neural networks (XPINNs): A generalized space-time domain decomposition based deep learning framework for nonlinear partial differential equations},
  author={Jagtap, Ameya D and Karniadakis, George Em},
  journal={Communications in Computational Physics},
  volume={28},
  number={5},
  year={2020},
  publisher={Brown Univ., Providence, RI (United States)}
}

@article{goswami2024learning,
  title={Learning stiff chemical kinetics using extended deep neural operators},
  author={Goswami, Somdatta and Jagtap, Ameya D and Babaee, Hessam and Susi, Bryan T and Karniadakis, George Em},
  journal={Computer Methods in Applied Mechanics and Engineering},
  volume={419},
  pages={116674},
  year={2024},
  publisher={Elsevier}
}

@article{jagtap2020locally,
  title={Locally adaptive activation functions with slope recovery for deep and physics-informed neural networks},
  author={Jagtap, Ameya D and Kawaguchi, Kenji and Em Karniadakis, George},
  journal={Proceedings of the Royal Society A},
  volume={476},
  number={2239},
  pages={20200334},
  year={2020},
  publisher={The Royal Society}
}

@article{jagtap2023important,
  title={How important are activation functions in regression and classification? A survey, performance comparison, and future directions},
  author={Jagtap, Ameya D and Karniadakis, George Em},
  journal={Journal of Machine Learning for Modeling and Computing},
  volume={4},
  number={1},
  year={2023},
  publisher={Begel House Inc.}
}

@article{de2024error,
  title={Error estimates for physics-informed neural networks approximating the Navier--Stokes equations},
  author={De Ryck, Tim and Jagtap, Ameya D and Mishra, Siddhartha},
  journal={IMA Journal of Numerical Analysis},
  volume={44},
  number={1},
  pages={83--119},
  year={2024},
  publisher={Oxford University Press}
}

@article{hu2021extended,
  title={When do extended physics-informed neural networks (XPINNs) improve generalization?},
  author={Hu, Zheyuan and Jagtap, Ameya D and Karniadakis, George Em and Kawaguchi, Kenji},
  journal={arXiv preprint arXiv:2109.09444},
  year={2021}
}

@article{mhaskar2025approximation,
  title={An Approximation Theory Perspective on Machine Learning},
  author={Mhaskar, Hrushikesh N and Tsoukanis, Efstratios and Jagtap, Ameya D},
  journal={arXiv preprint arXiv:2506.02168},
  year={2025}
}

@article{shin2020convergence,
  title={On the convergence of physics informed neural networks for linear second-order elliptic and parabolic type PDEs},
  author={Shin, Yeonjong and Darbon, Jerome and Karniadakis, George Em},
  journal={arXiv preprint arXiv:2004.01806},
  year={2020}
}

@article{li2020fourier,
  title={Fourier neural operator for parametric partial differential equations},
  author={Li, Zongyi and Kovachki, Nikola and Azizzadenesheli, Kamyar and Liu, Burigede and Bhattacharya, Kaushik and Stuart, Andrew and Anandkumar, Anima},
  journal={arXiv preprint arXiv:2010.08895},
  year={2020}
}

@article{li2022transformer,
  title={Transformer for partial differential equations' operator learning},
  author={Li, Zijie and Meidani, Kazem and Farimani, Amir Barati},
  journal={arXiv preprint arXiv:2205.13671},
  year={2022}
}

@article{tripura2023wavelet,
  title={Wavelet neural operator for solving parametric partial differential equations in computational mechanics problems},
  author={Tripura, Tapas and Chakraborty, Souvik},
  journal={Computer Methods in Applied Mechanics and Engineering},
  volume={404},
  pages={115783},
  year={2023},
  publisher={Elsevier}
}

@inproceedings{raonic2023convolutional,
  title={Convolutional neural operators},
  author={Raonic, Bogdan and Molinaro, Roberto and Rohner, Tobias and Mishra, Siddhartha and de Bezenac, Emmanuel},
  booktitle={ICLR 2023 workshop on physics for machine learning},
  year={2023}
}

@article{liu2024kan,
  title={Kan: Kolmogorov-arnold networks},
  author={Liu, Ziming and Wang, Yixuan and Vaidya, Sachin and Ruehle, Fabian and Halverson, James and Solja{\v{c}}i{\'c}, Marin and Hou, Thomas Y and Tegmark, Max},
  journal={arXiv preprint arXiv:2404.19756},
  year={2024}
}

@article{shukla2024comprehensive,
  title={A comprehensive and FAIR comparison between MLP and KAN representations for differential equations and operator networks},
  author={Shukla, Khemraj and Toscano, Juan Diego and Wang, Zhicheng and Zou, Zongren and Karniadakis, George Em},
  journal={arXiv preprint arXiv:2406.02917},
  year={2024}
}

@article{Sidharth2024ChebyshevPK,
  title={Chebyshev Polynomial-Based Kolmogorov-Arnold Networks: An Efficient Architecture for Nonlinear Function Approximation},
  author={SS Sidharth and R Gokul},
  journal={ArXiv},
  year={2024},
  volume={abs/2405.07200},
  url={https://api.semanticscholar.org/CorpusID:269757002}
}

@article{10763509,
      author = {Rigas, Spyros and Papachristou, Michalis and Papadopoulos, Theofilos and Anagnostopoulos, Fotios and Alexandridis, Georgios},
      journal = {IEEE Access}, 
      title = {Adaptive Training of Grid-Dependent Physics-Informed Kolmogorov-Arnold Networks}, 
      year = {2024},
      volume = {12},
      pages = {176982-176998},
      doi = {10.1109/ACCESS.2024.3504962}
}

@inproceedings{rahaman2019spectral,
  title={On the spectral bias of neural networks},
  author={Rahaman, Nasim and Baratin, Aristide and Arpit, Devansh and Draxler, Felix and Lin, Min and Hamprecht, Fred and Bengio, Yoshua and Courville, Aaron},
  booktitle={International conference on machine learning},
  pages={5301--5310},
  year={2019},
  organization={PMLR}
}

@article{abueidda2025deepokan,
  title={Deepokan: Deep operator network based on kolmogorov arnold networks for mechanics problems},
  author={Abueidda, Diab W and Pantidis, Panos and Mobasher, Mostafa E},
  journal={Computer Methods in Applied Mechanics and Engineering},
  volume={436},
  pages={117699},
  year={2025},
  publisher={Elsevier}
}
\end{document}